\def\year{2020}\relax
\documentclass[letterpaper]{article} 

\usepackage{cite}
\usepackage{subcaption}

\def\BibTeX{{\rm B\kern-.05em{\sc i\kern-.025em b}\kern-.08em
    T\kern-.1667em\lower.7ex\hbox{E}\kern-.125emX}}

\usepackage{aaai20}  
\usepackage{times}  
\usepackage{helvet} 
\usepackage{courier}  
\usepackage[hyphens]{url}  
\usepackage{graphicx} 
\usepackage{xcolor}
\usepackage{comment}

\newcommand{\citeauthoryearp}[1]{\citeauthor{#1} (\citeyear{#1})}

\usepackage{multirow}
\begin{document}

\title{Exploring Level Blending across Platformers via Paths and Affordances}
\author{Anurag Sarkar\textsuperscript{1}, Adam Summerville\textsuperscript{2}, Sam Snodgrass\textsuperscript{3}, Gerard Bentley\textsuperscript{4} and Joseph Osborn\textsuperscript{4} \\
\textsuperscript{1}Northeastern University\\
\textsuperscript{2}California State Polytechnic University\\
\textsuperscript{3}modl.ai\\
\textsuperscript{4}Pomona College\\
sarkar.an@northeastern.edu,
asummerville@cpp.edu,
sam@modl.ai,
gbkh2015@mymail.pomona.edu,
joseph.osborn@pomona.edu\\}

\maketitle

\begin{abstract}
\label{sec:abstract}
Techniques for procedural content generation via machine learning (PCGML) have been shown to be useful for generating novel game content. While used primarily for producing new content in the style of the game domain used for training, recent works have increasingly started to explore methods for discovering and generating content in novel domains via techniques such as level blending and domain transfer. In this paper, we build on these works and introduce a new PCGML approach for producing novel game content spanning multiple domains. We use a new affordance and path vocabulary to encode data from six different platformer games and train variational autoencoders on this data, enabling us to capture the latent level space spanning all the domains and generate new content with varying proportions of the different domains.
\end{abstract}

\section{Introduction}\label{sec:intro}
Procedural content generation via machine learning (PCGML)~\cite{summerville2017procedural} refers to techniques for PCG using models trained on existing game data. This enables the generation of game content that is novel but still captures the characteristics and patterns of the data used for training. While a lot of prior PCGML work has looked at generating content for a single domain such as \textit{Super Mario Bros.}~\cite{summerville2016mariostring}, \textit{The Legend of Zelda}~\cite{summerville2015samplinghyrule} and \textit{Doom}~\cite{giacomello2018doom}, recent works have started to focus on more creative PCGML techniques~\cite{guzdial2018combinatorial} that attempt to learn models capable of producing content outside of the domain used for training. That is, rather than attempting to produce novel variations of content within an existing domain, these techniques blend existing domains into entirely new ones and generate content for these new domains. This yields models that can generalize better across different domains---a key challenge of PCGML.  Moreover, such cross-domain models could prompt the discovery of novel game designs previously hidden in the combined design space of the input domains. Techniques for such PCGML approaches have involved domain transfer~\cite{snodgrass2016approach}, blending separate models to produce generators~\cite{sarkar2018blending}, and constructing game graphs using learned models of levels and game rules~\cite{guzdial2018automated}.  Our paper most directly builds on  the approach of training models on multiple domains to learn a new blended domain space~\cite{sarkar2019blending}.

We extend the prior method in a number of ways. We increase the input domain from two to six games thus greatly expanding the possibility space of the learned, blended domain in terms of new content. We also introduce a new, unified affordance vocabulary consistent across all domains which expands significantly on such vocabularies used in prior PCGML work. Rather than ignoring game mechanics, we annotate the input domains with paths generated by an A* agent informed by the jump arcs of each domain. Finally, like prior blending work, we use variational autoencoders (VAEs) to learn our models but train two varieties---one composed of linear layers and another using GRU layers---and compare their performance. To the best of our knowledge, this is the first application of a GRU-VAE for PCGML.

\section{Related Work}\label{sec:rw}
Procedural content generation via machine learning (PCGML)~\cite{summerville2017procedural} describes a collection of PCG techniques that leverage existing data to build a model from which new content can be sampled. There has been extensive work on developing and adapting various models (e.g., LSTMs~\cite{summerville2016mariostring}, GANs~\cite{volz2018evolving}, Bayesian Networks~\cite{guzdial2016game}, etc.) to be used in specific domains (e.g., \textit{Super Mario Bros.}~\cite{snodgrass2017learning}, \textit{The Legend of Zelda}~\cite{summerville2015samplinghyrule}, \textit{Kid Icarus}~\cite{snodgrass2017learning}, etc.). The drawback of these PCGML approaches is that they require training data from the domain they are meant to create content for, limiting their use in new domains. 

Several researchers have considered the limitation of requiring training data in the target domain. \citeauthoryearp{snodgrass2016approach} presented a domain transfer approach that tried to find mappings between domains to enable data in one domain to supplement the training data in another. This still requires some data from a target domain and cannot generate content for new domains. We instead extend prior work that blends domains together to produce new domains. \citeauthoryearp{sarkar2018blending} used LSTMs trained on a combined dataset as well as weighting separately trained LSTMs to generate blended levels with segments from both \textit{Super Mario Bros.} and \textit{Kid Icarus}. \citeauthoryearp{sarkar2019blending} further applied VAEs for level blending, training models on a combined dataset of segments from \textit{Mario} and \textit{Kid Icarus} to admit interpolation between these domains. We extend the above blending and transfer approaches by: incorporating several additional domains to blend; defining a unified level representation for the domains; annotating level data with path information (so models can train on gameplay as well as structural information); and evaluating a GRU-VAE approach to level blending.

Other approaches have also been considered for blending games. \citeauthoryearp{gow2015towards} presented a framework for blending VGDL~\cite{schaul2014extensible} descriptions of games, demonstrating their approach by creating \textit{Frolda}---a game created by blending \textit{Zelda} and \textit{Frogger}. However, their approach was manual and limited to the VGDL framework. \citeauthoryearp{guzdial2018automated} presented \textit{conceptual expansion}---an ML-based approach which creates game concept graphs that can be blended and recombined to produce new games. Recently, \citeauthoryearp{snodgrass2020multi} combined binary space partitioning and VAEs to generate blended platformer levels using a sketch-based level representation.

Like some past blending works, we use variational autoencoders (VAEs). VAEs~\cite{kingma2013autoencoding}, as well as vanilla autoencoders~\cite{hinton2006reducing}, have been used in several prior PCGML works. \citeauthoryearp{jain2016autoencoders} used autoencoders for generating and repairing Mario levels while~\citeauthoryearp{guzdial2018explainable} used autoencoders to generate Mario level structures conforming to specific design patterns. \citeauthoryearp{thakkar2019autoencoder} applied both vanilla and variational autoencoders for generating \textit{Lode Runner} levels. Aside from level generation,~\citeauthoryearp{alvernaz2017autoencoder} used an autoencoder to learn a low-dimensional representation of the \textit{VizDoom} environment and used it to evolve controllers while~\citeauthoryearp{soares2019deep} used VAEs for classifying procedurally generated NPC behavior.

\section{Level Representation}\label{sec:rep}
We evaluate our approaches using six classic NES platforming games, namely, \textit{Super Mario Bros.}, \textit{Super Mario Bros. II: The Lost Levels}, \textit{Ninja Gaiden}, \textit{Metroid}, \textit{Mega Man}, and \textit{Castlevania}. Levels in these games vary greatly in their specific tile-based representations commonly used by PCGML techniques~\cite{summerville2017procedural} and the VGLC~\cite{summerville2016vglc} (e.g., different tile types defined for enemy types, power-ups, obstacles, etc.). For our models to reason across domains, we need to unify the representations. We accomplish this by leveraging the tile affordance work of \citeauthoryearp{bentley2019videogame} to derive a common language. 

We describe tiles in terms of the following affordances: \textit{solid} (the player can stand on this tile); \textit{climbable} (the player can use this tile to climb); \textit{passable} (the player can pass through this tile); \textit{powerup} (this tile strengthens the player in some way); \textit{hazard} (this tile harms the player); \textit{moving} (this tile changes location); \textit{portal} (this tile transports the player somewhere); \textit{collectable} (the player can pick up this tile); \textit{breakable} (the player can destroy this); and \textit{null} (this tile indicates a position outside of the actual level geometry).  This differs from the Bentley and Osborn affordances and is specialized for side-scrolling games.  Using the above affordances, we defined a uniform set of tile types to represent all of the domains in our experiments:
\begin{itemize}
    \item[] \texttt{X}: \textit{solid}, (e.g., ground or platforms)
    \item[] \texttt{S}: \textit{solid, breakable}, (e.g., breakable bricks in \textit{SMB})
    \item[] \texttt{\#}: \textit{solid, moving}, (e.g., moving platforms)
    \item[] \texttt{|}: \textit{solid, passable, climbable}, (e.g., ladders)
    \item[] \texttt{v}: \textit{hazard}, (e.g., spikes)
    \item[] \texttt{\^}: \textit{solid, hazard}, (e.g., lava or solid spikes)
    \item[] \texttt{e}: \textit{moving, hazard}, (e.g., enemies)
    \item[] \texttt{E}: \textit{solid, moving, passable, hazard}, (e.g., enemies the player could pass through or jump on)
    \item[] \texttt{o}: \textit{collectable}, (e.g., coins)
    \item[] \texttt{*}: \textit{collectable, powerup}, (e.g., weapon refills in \textit{MM})
    \item[] \texttt{Q}: \textit{solid, collectable}, (e.g., coin blocks in \textit{SMB})
    \item[] \texttt{!}: \textit{solid, powerup}, (e.g., mushroom blocks in \textit{SMB})
    \item[] \texttt{\$}: \textit{portal}, (e.g., doors in \textit{Metroid})
    \item[] \texttt{@}: \textit{solid, null, hazard}
\end{itemize}

\noindent We further annotate our levels with $A^*$ agent-generated paths, where the agent is tuned to each domain prior to annotation according to the possible jump arcs in that domain. We annotate our levels with paths because 1) incorporating pathing information has been shown to be beneficial to PCGML techniques when learning the level geometry~\cite{summerville2016mariostring,snodgrass2017procedural}, and 2) it is our longer term goal to learn and blend not only the level geometry, but the jump physics as well. 

Lastly, the levels in each domain have vastly different sizes and dimensions. We address this by breaking each level into $15\times32$ sized segments, padding them vertically when needed. For this work, we used the horizontal portions of these levels to enable more holistic blending across domains and leave considerations of vertical level segments in \textit{Ninja Gaiden}, \textit{Metroid} and \textit{Mega Man} for future work. Thus, to generate our training data, we slid a $15\times32$ window horizontally across all levels. Since some of the games have discrete rooms (\textit{Castlevania}, \textit{Mega Man}, \textit{Ninja Gaiden}, and \textit{Metroid}), if a door is found in the segment and it does not lie on the edge of the segment, the segment is discarded. Duplicate segments are also discarded (these often appear in \textit{Metroid}). We produced $775$ segments for \textit{Castlevania (CV)}, $1907$ segments for \textit{Mario (SMB)} (henceforth, we use \textit{Mario} and \textit{SMB} to refer to the combined domain of both \textit{Super Mario Bros.} and \textit{Super Mario Bros II: The Lost Levels}), $924$ segments for \textit{Mega Man (MM)}, $1833$ segments for \textit{Metroid (Met)} and $504$ segments for \textit{Ninja Gaiden (NG)}. We oversampled all domains other than \textit{SMB} to obtain approximately equal number of segments from all games in order to prevent the learned blended models from skewing towards any specific game(s).

\begin{figure*}	
	\centering
	\begin{subfigure}{.19\textwidth}
		\centering
		\includegraphics[width=0.15\pdfpagewidth]{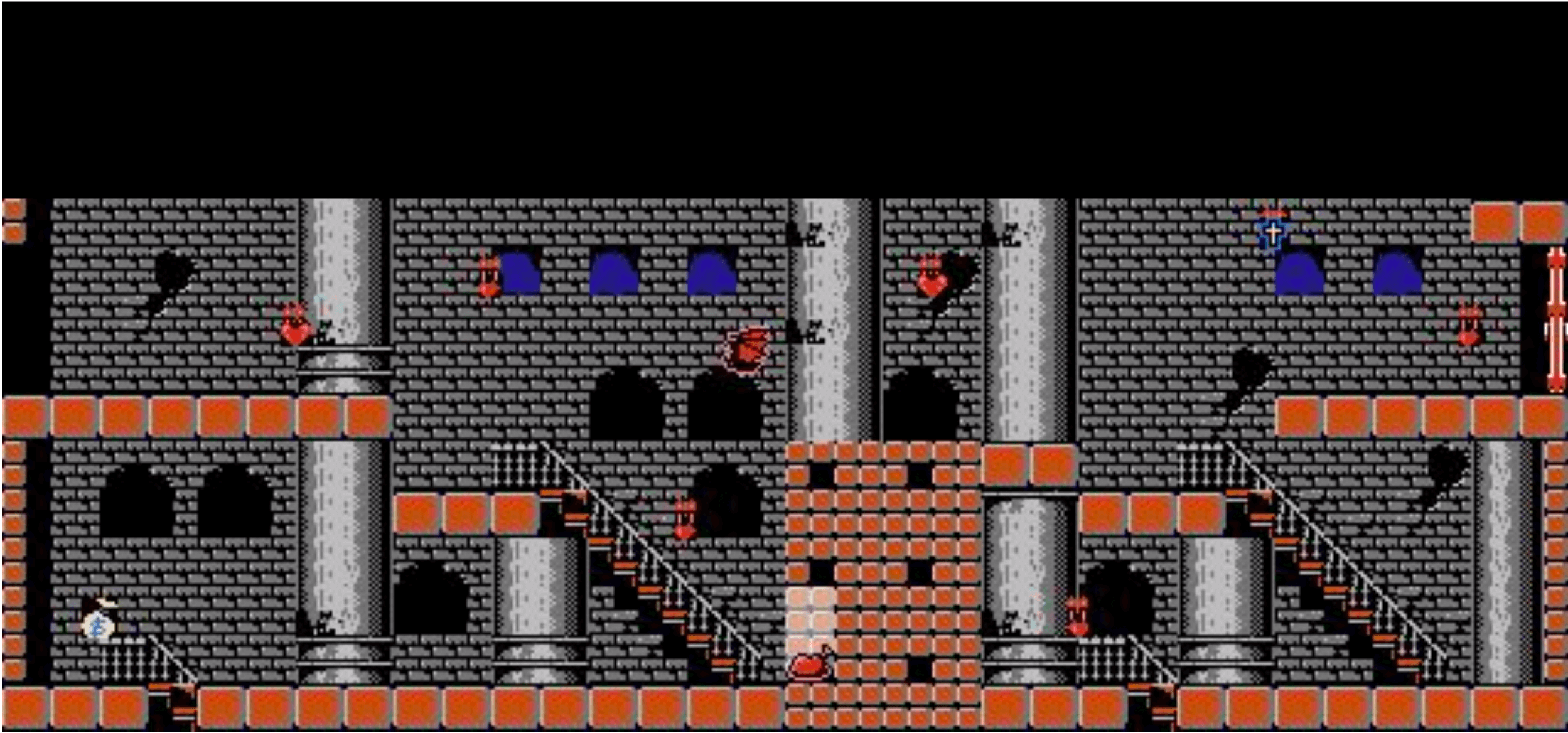}
		\caption{\textit{CV} level window}
	\end{subfigure}
	\begin{subfigure}{.19\textwidth}
		\centering
		\includegraphics[width=0.15\pdfpagewidth]{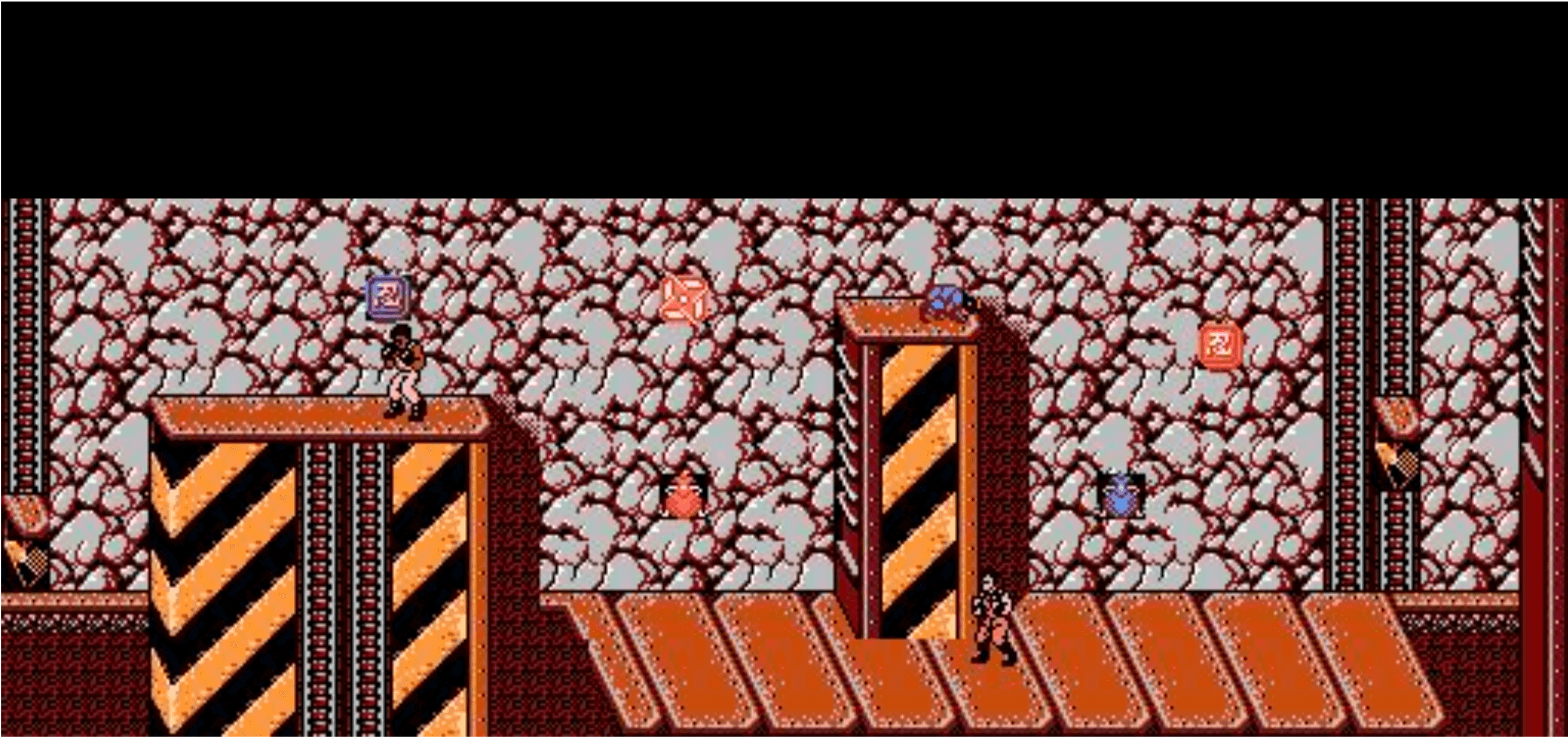}
		\caption{\textit{NG} level window}
	\end{subfigure}
	\begin{subfigure}{.19\textwidth}
		\centering
		\includegraphics[width=0.15\pdfpagewidth]{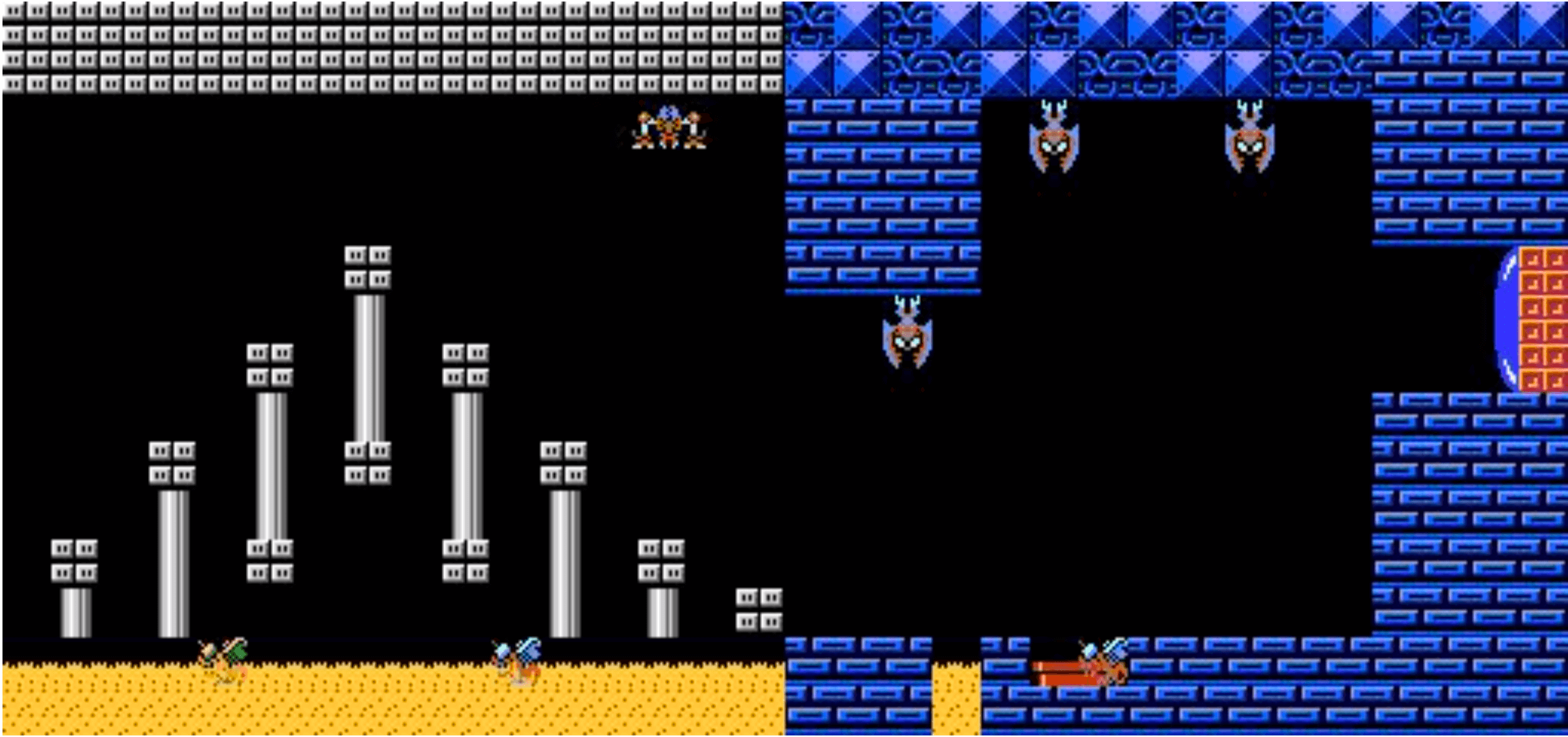}
		\caption{\textit{Met} level window}		
	\end{subfigure}
	\begin{subfigure}{.19\textwidth}
		\centering
		\includegraphics[width=0.15\pdfpagewidth]{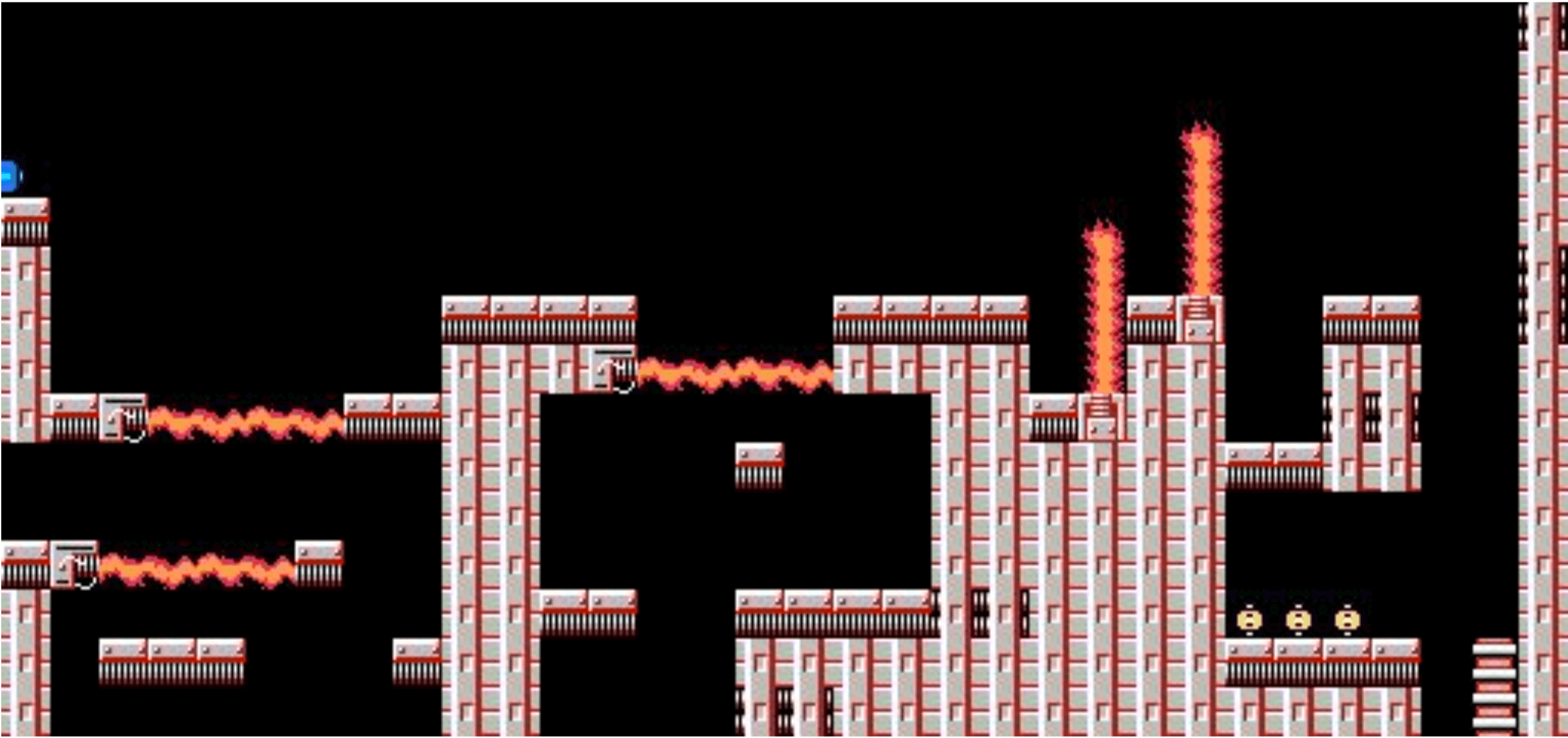}
		\caption{\textit{MM} level window}
	\end{subfigure}
	\begin{subfigure}{.19\textwidth}
		\centering
		\includegraphics[width=0.15\pdfpagewidth]{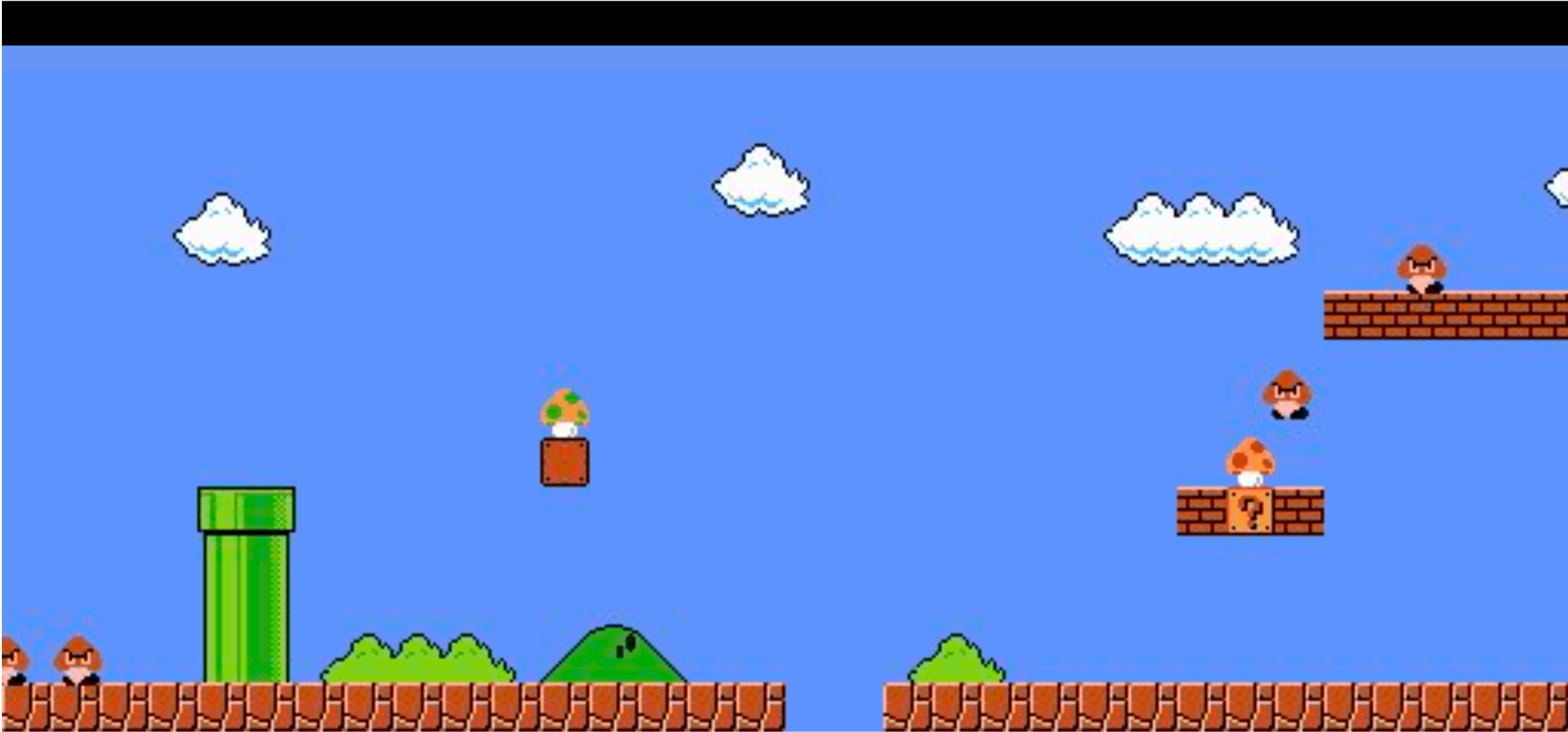}
		\caption{\textit{SMB} level window}
	\end{subfigure}
	
	\begin{subfigure}{.19\textwidth}
		\centering
		\includegraphics[width=0.15\pdfpagewidth]{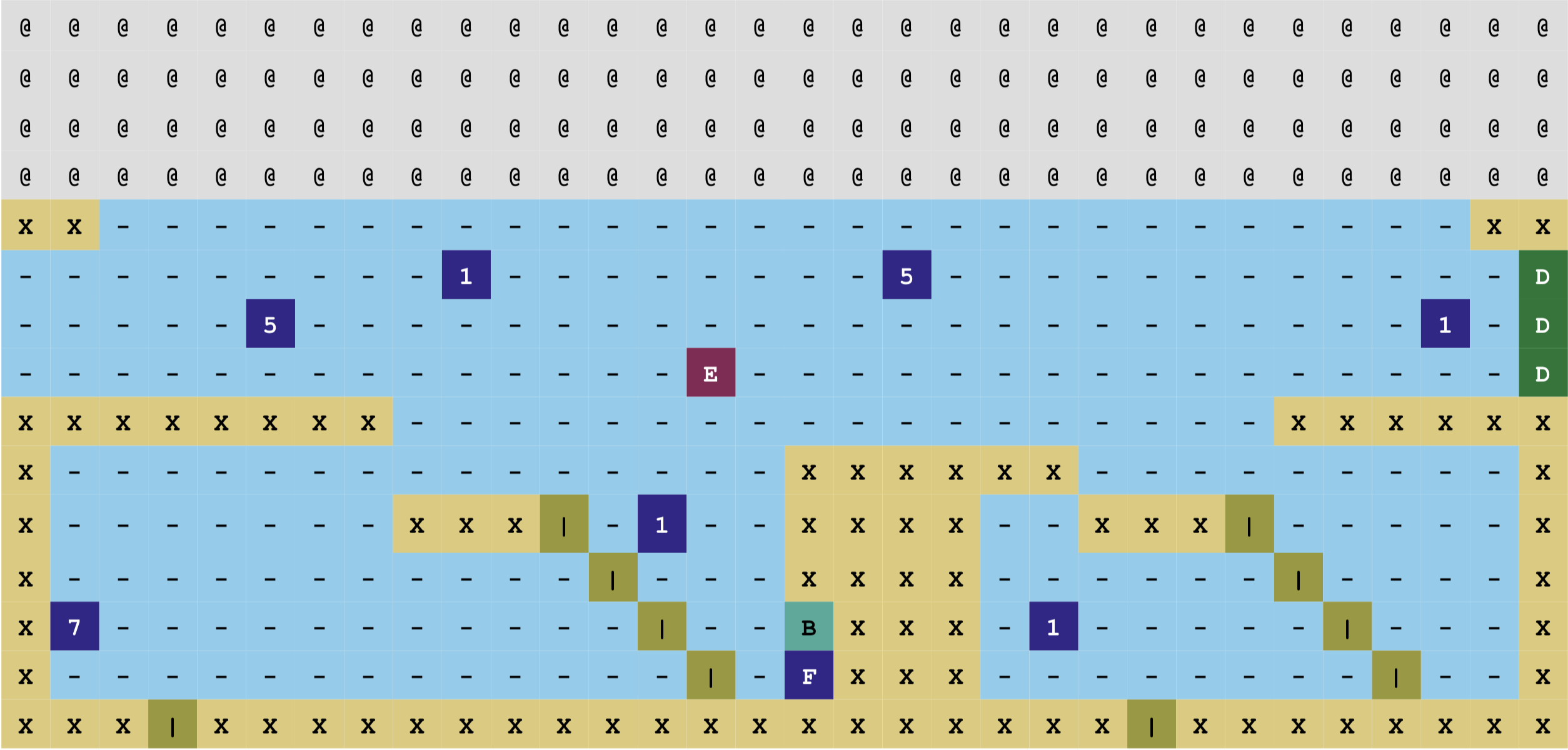}
		\caption{\textit{CV} basic tileset}
	\end{subfigure}
	\begin{subfigure}{.19\textwidth}
		\centering
		\includegraphics[width=0.15\pdfpagewidth]{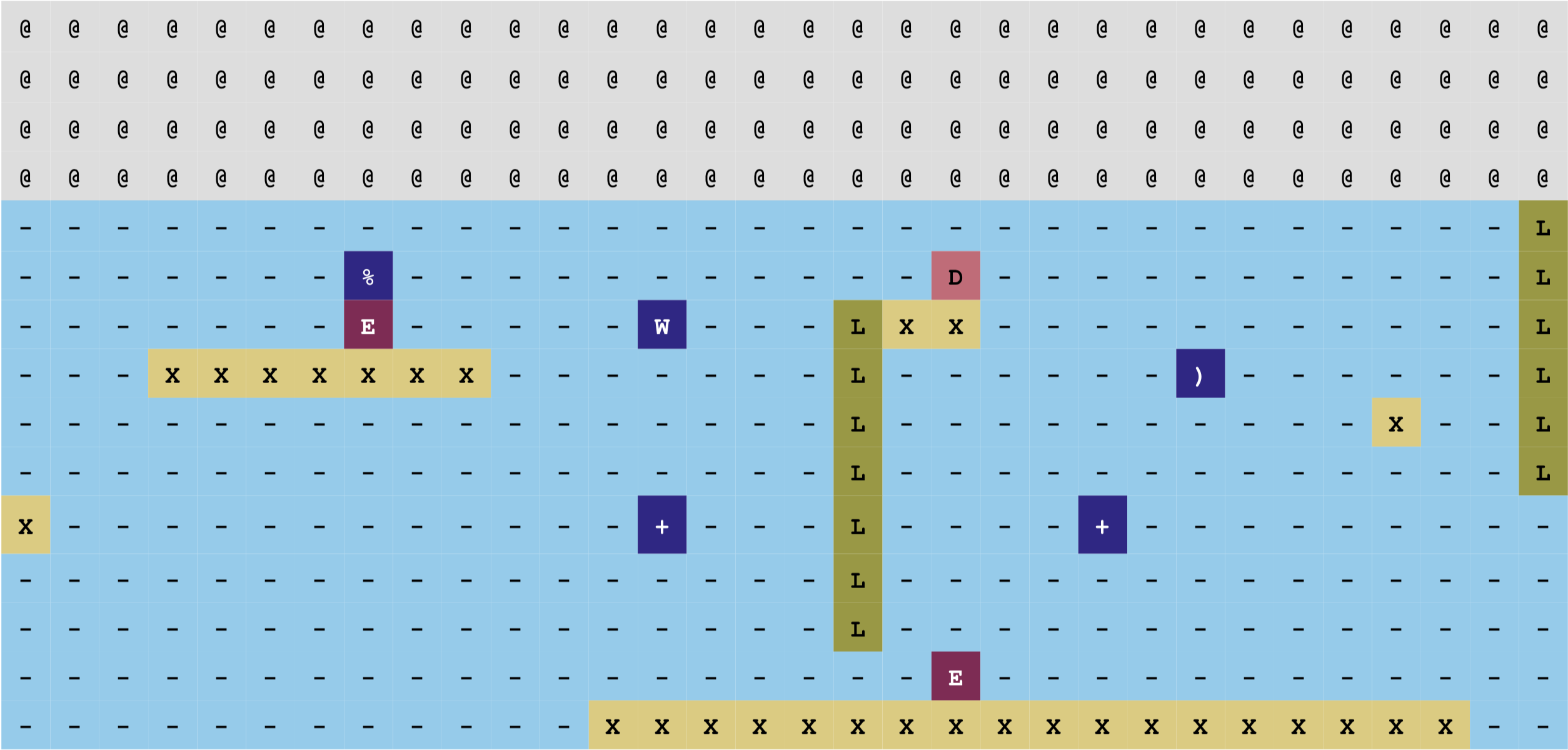}
		\caption{\textit{NG} basic tileset}
	\end{subfigure}
	\begin{subfigure}{.19\textwidth}
		\centering
		\includegraphics[width=0.15\pdfpagewidth]{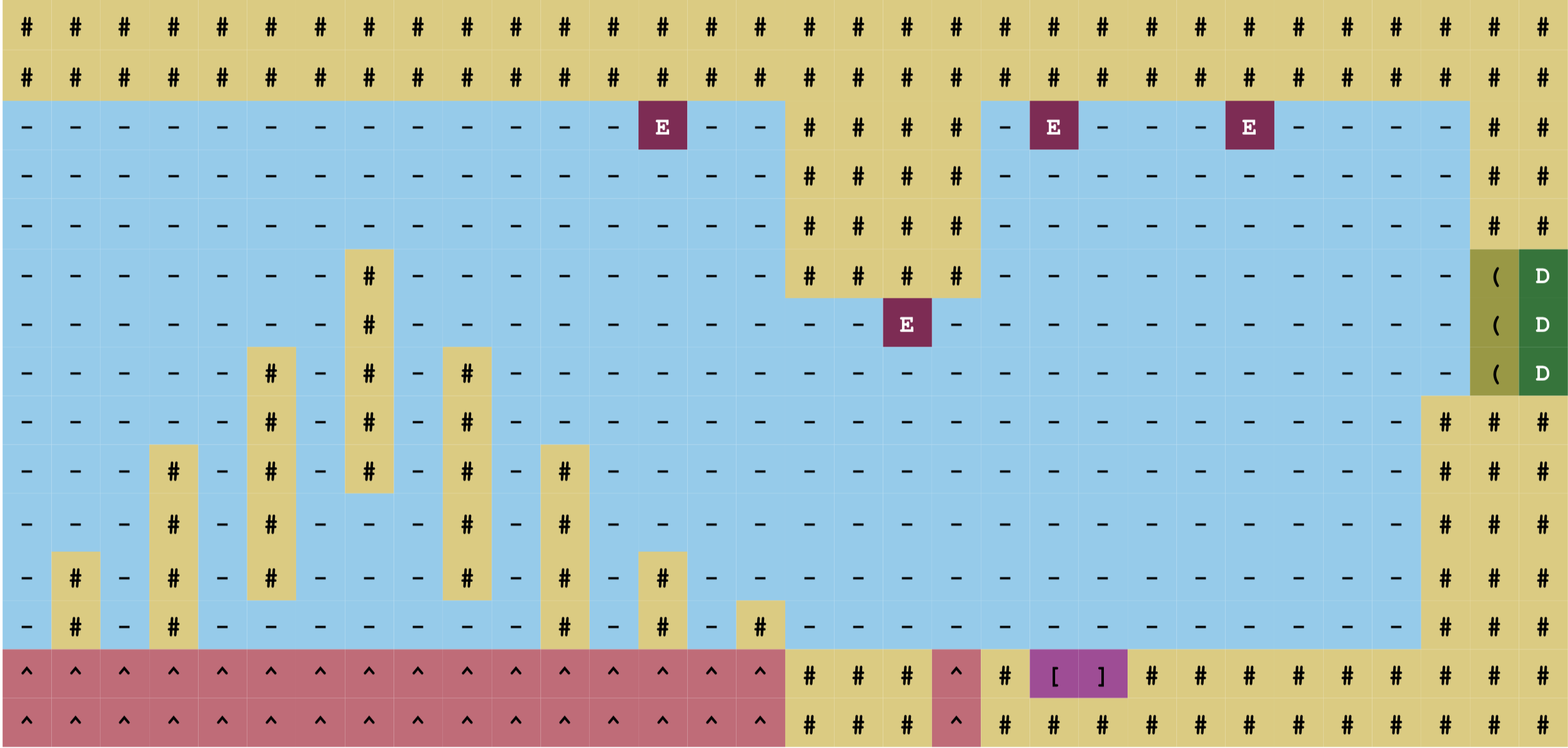}
		\caption{\textit{Met} basic tileset}		
	\end{subfigure}
	\begin{subfigure}{.19\textwidth}
		\centering
		\includegraphics[width=0.15\pdfpagewidth]{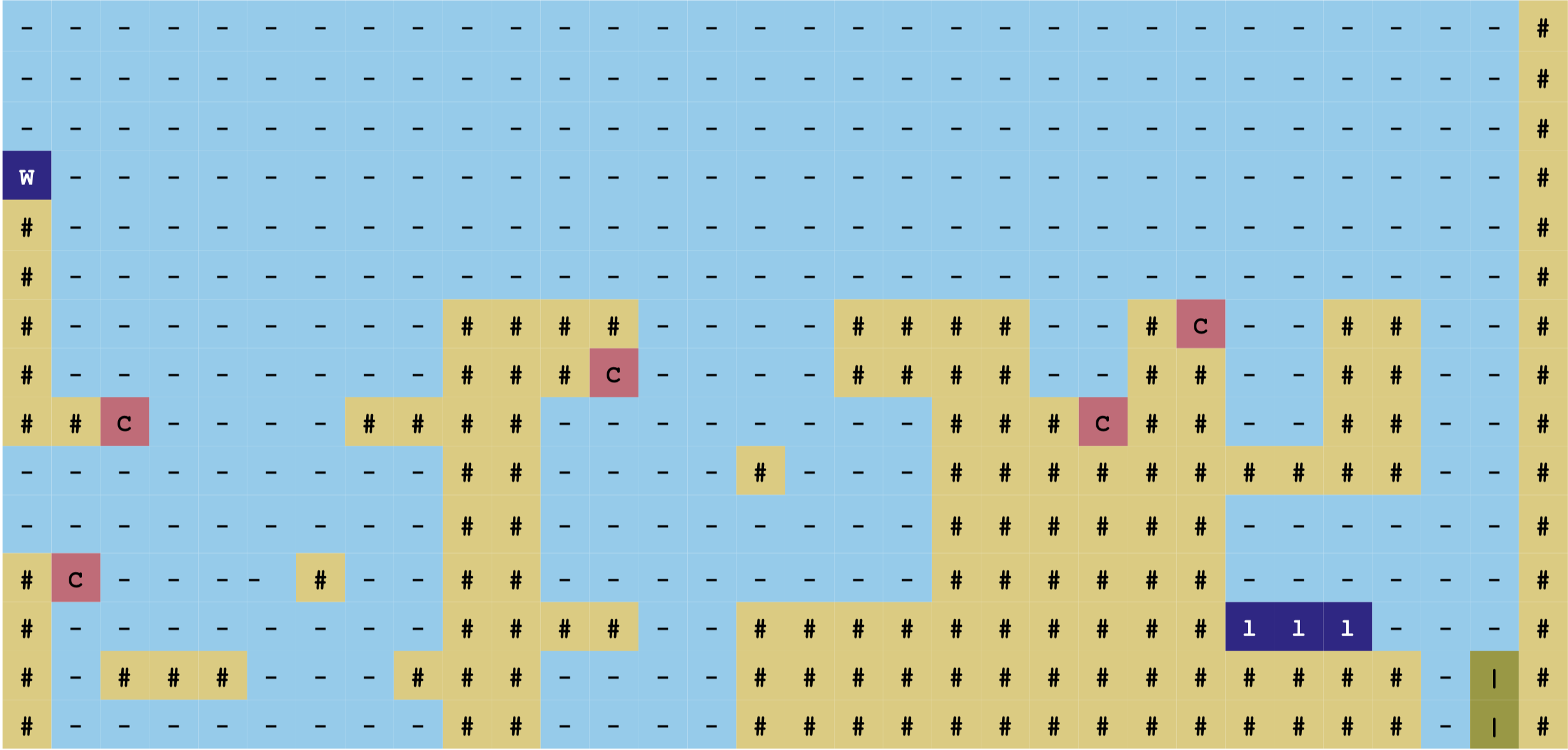}
		\caption{\textit{MM} basic tileset}
	\end{subfigure}
	\begin{subfigure}{.19\textwidth}
		\centering
		\includegraphics[width=0.15\pdfpagewidth]{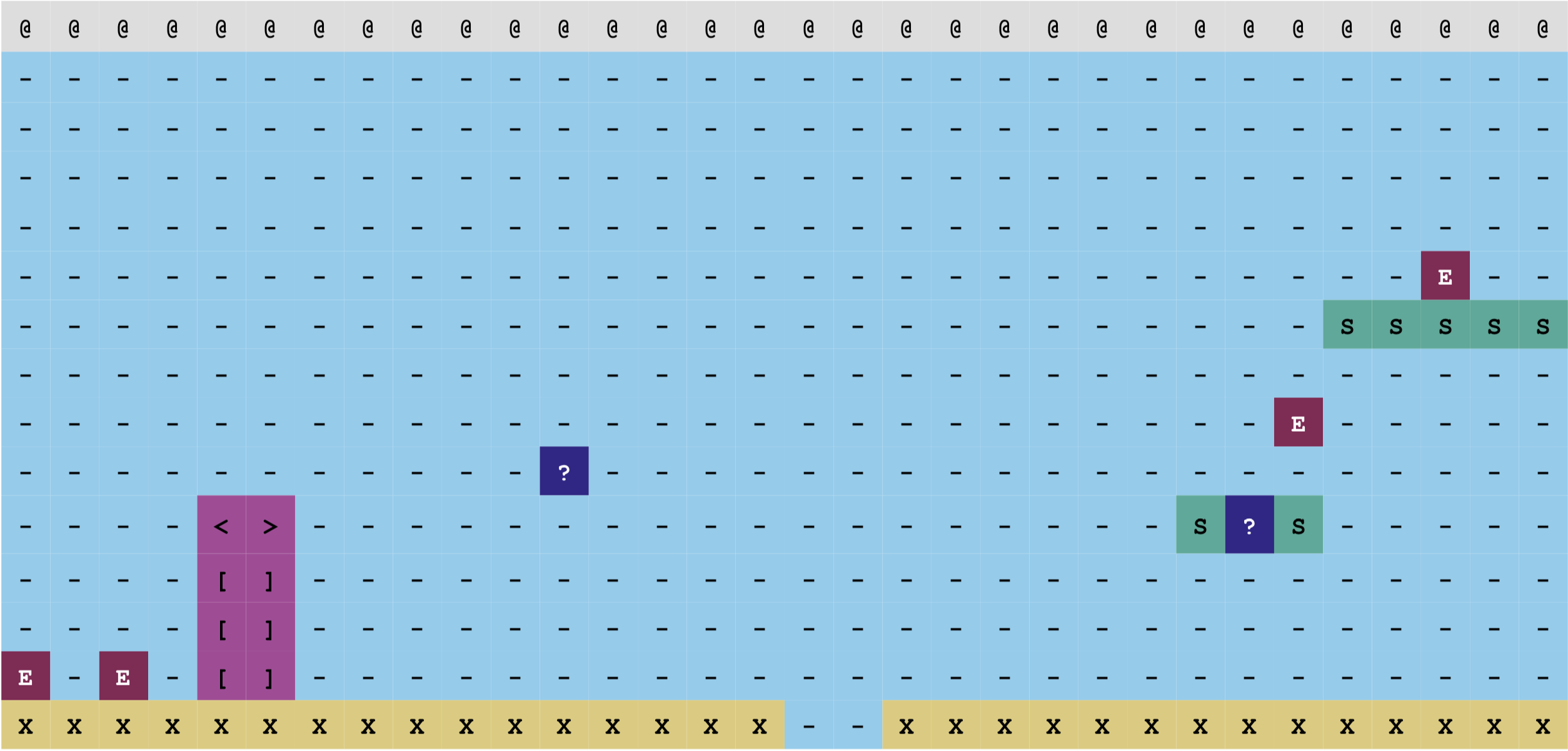}
		\caption{\textit{SMB} basic tileset}
	\end{subfigure}
	
	\begin{subfigure}{.19\textwidth}
		\centering
		\includegraphics[width=0.15\pdfpagewidth]{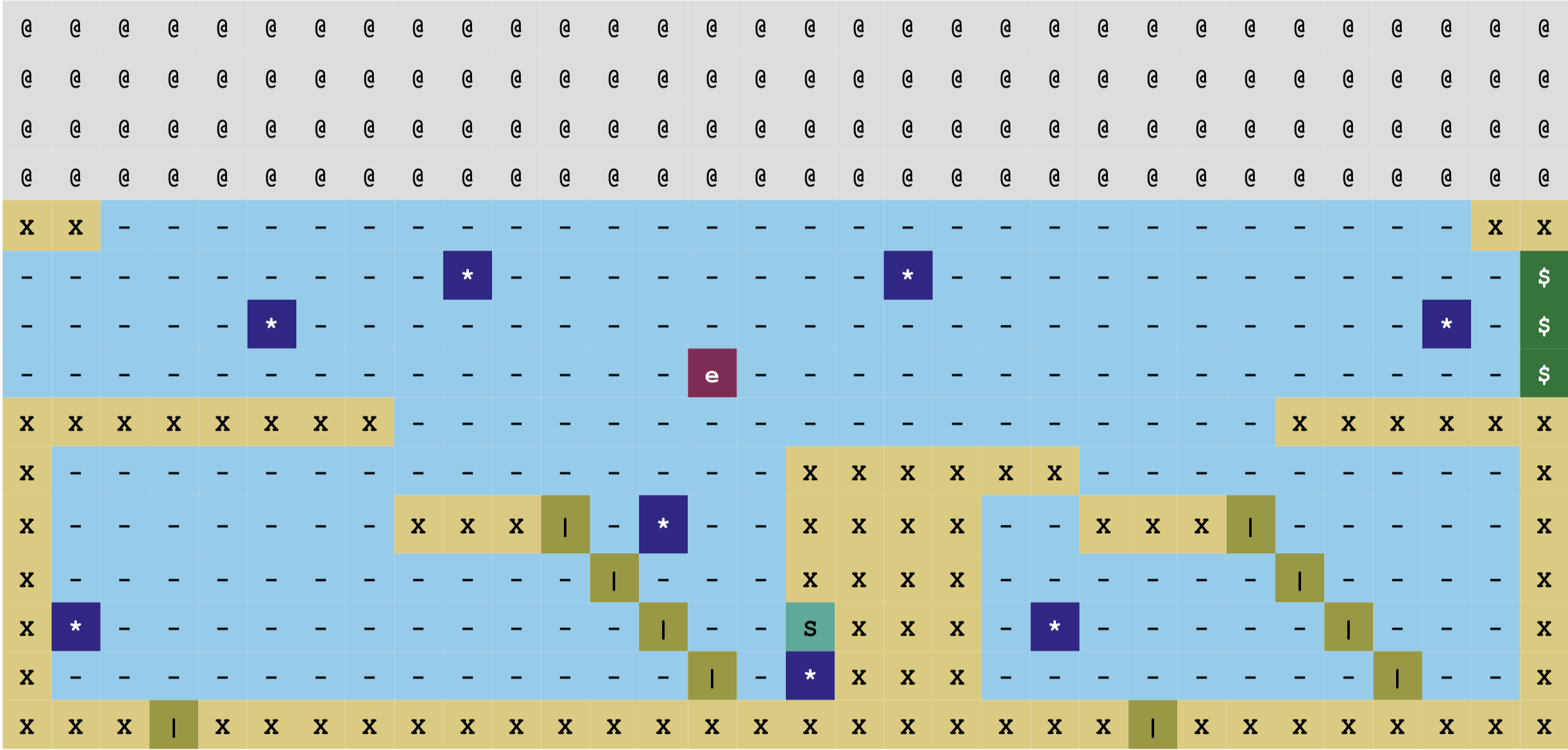}
		\caption{\textit{CV} uniform tileset}
	\end{subfigure}
	\begin{subfigure}{.19\textwidth}
		\centering
		\includegraphics[width=0.15\pdfpagewidth]{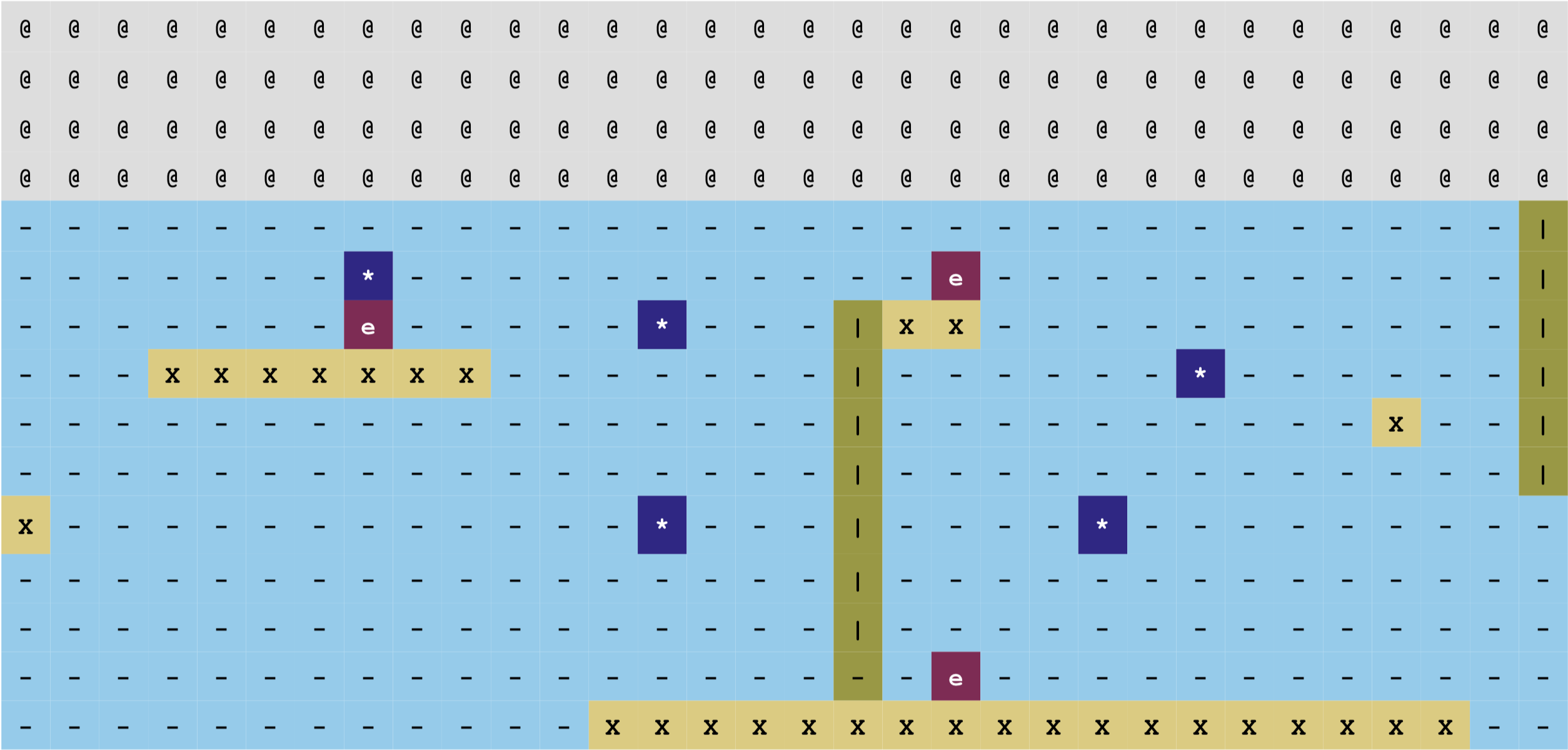}
		\caption{\textit{NG} uniform tileset}
	\end{subfigure}
	\begin{subfigure}{.19\textwidth}
		\centering
		\includegraphics[width=0.15\pdfpagewidth]{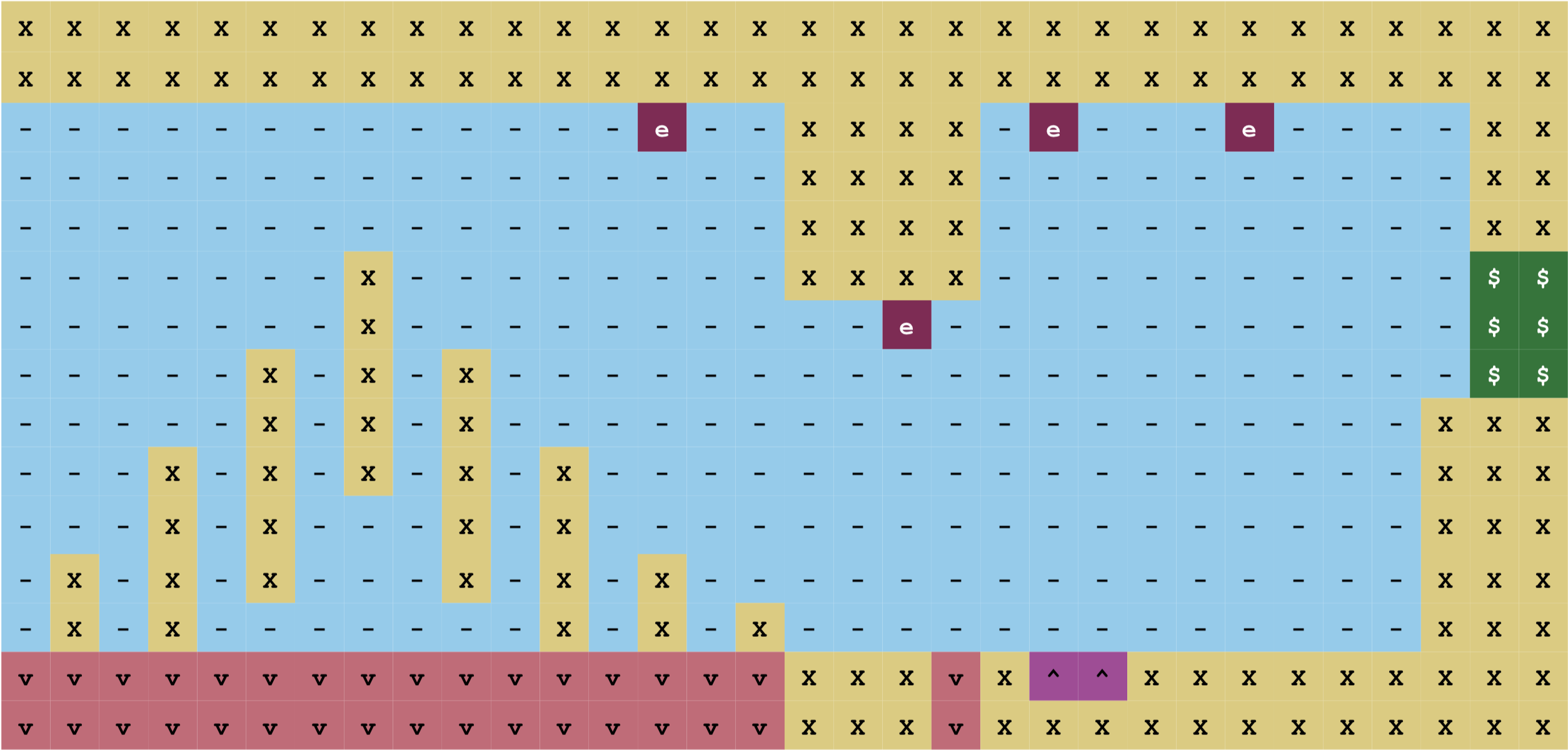}
		\caption{\textit{Met} uniform tileset}		
	\end{subfigure}
	\begin{subfigure}{.19\textwidth}
		\centering
		\includegraphics[width=0.15\pdfpagewidth]{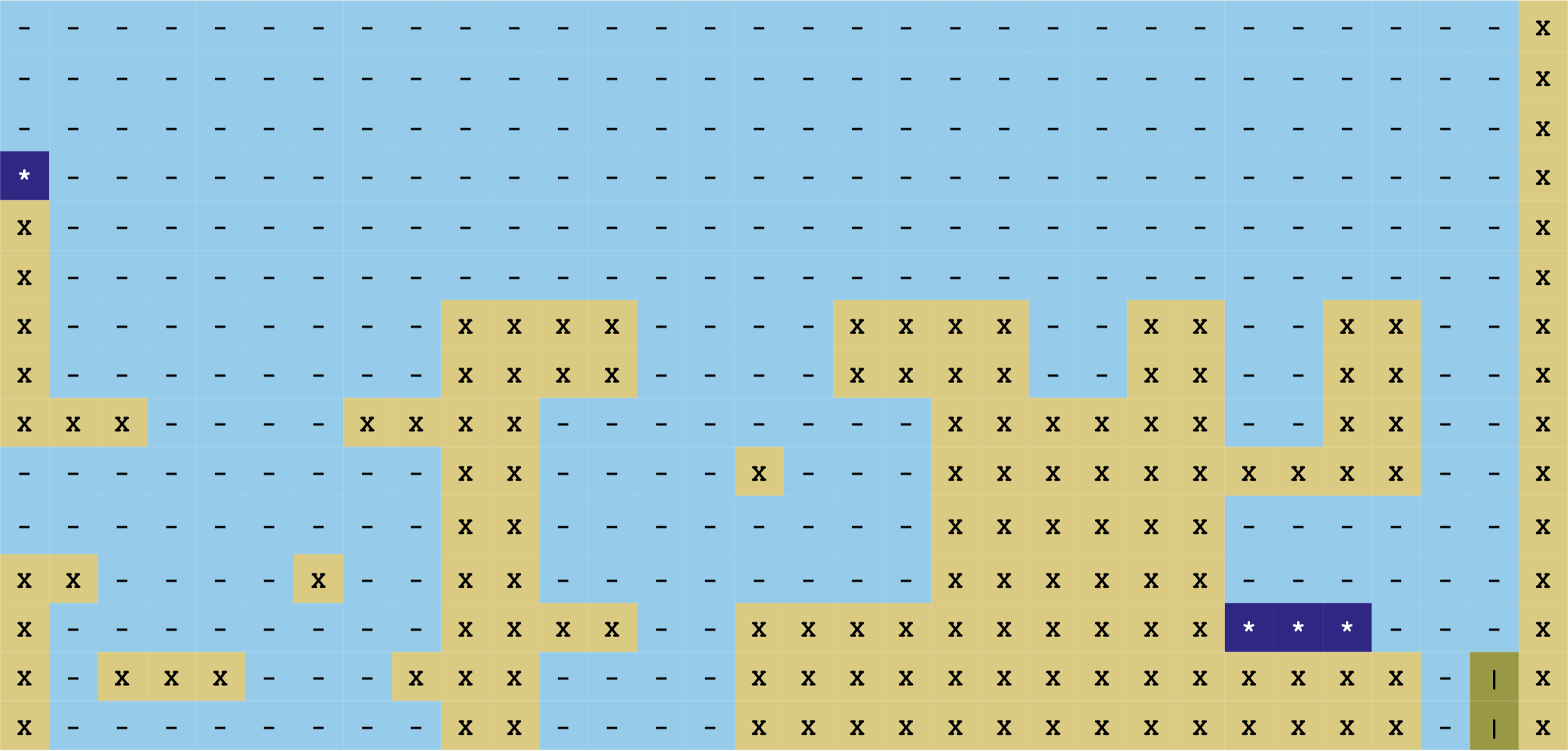}
		\caption{\textit{MM} uniform tileset}
	\end{subfigure}
	\begin{subfigure}{.19\textwidth}
		\centering
		\includegraphics[width=0.15\pdfpagewidth]{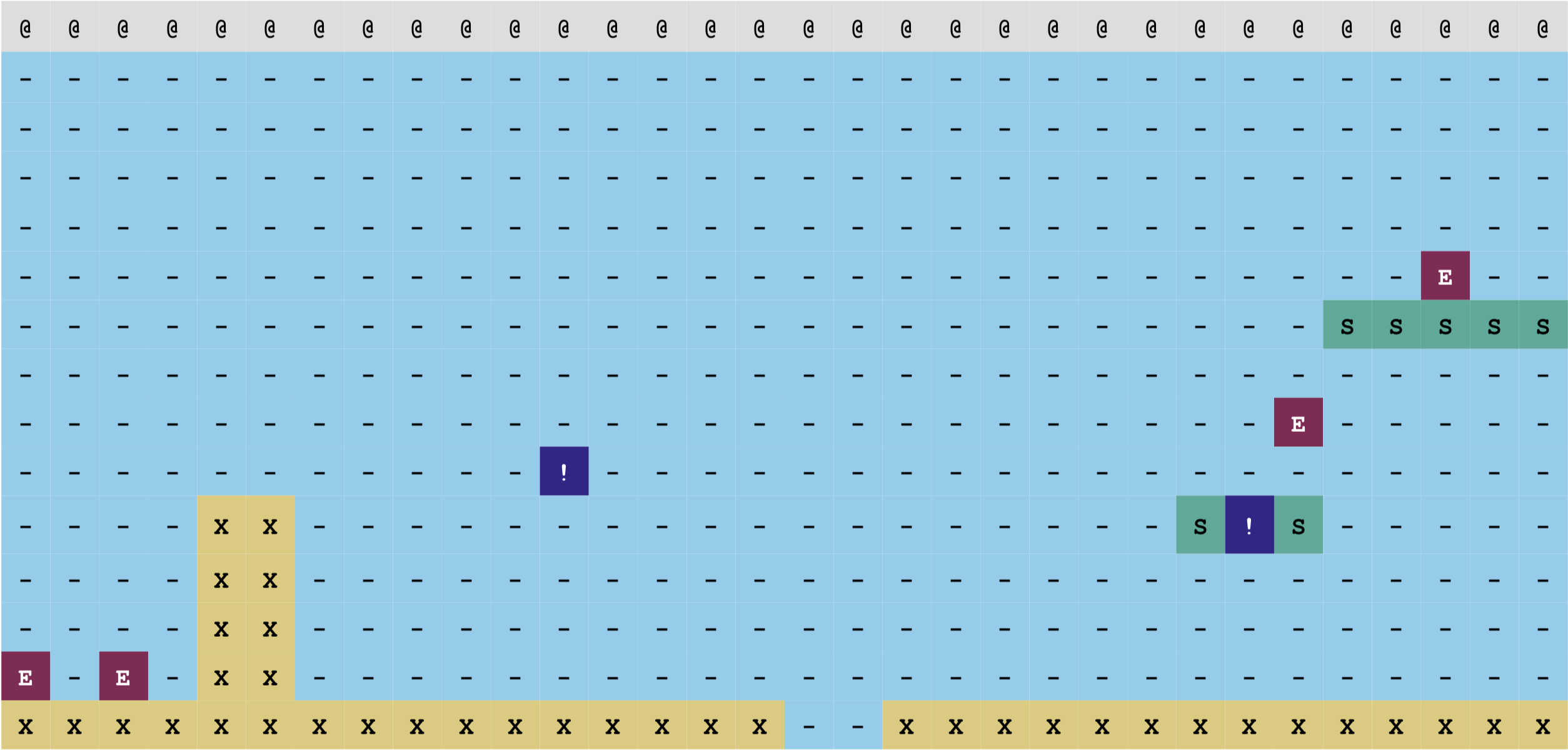}
		\caption{\textit{SMB} uniform tileset}
	\end{subfigure}
	
	\caption{This figure shows a window from a level from each of our domains (top row), that window represented using that domain's defined tileset (center row), and that window represented using our uniform tileset (bottom row).}\label{fig:Rep}
\end{figure*}


\section{Modeling and Generation}\label{sec:gen}

We used two different variants of VAEs for building our models. Both were implemented using PyTorch~\cite{paszke2017automatic} and are described in the following sections.

\subsection{Linear VAE}
For the linear model, the encoder and decoder each consisted of 4 fully-connected linear layers using ReLU activation. Encoder layers had dimensionalities (11520x1024), (1024x512), (512x256) and (256xlatent size) with decoder layers having these in reverse order. We trained four versions of this model using latent dimensions of 32, 64, 128 and 256. All models were trained for 5000 epochs using the Adam optimizer and a learning rate of 0.001.

\subsection{GRU-VAE}

The GRU-VAE is a sequence-to-sequence recurrent architecture that has encoder and decoder Gated Recurrent Units (GRU)~\cite{gru} with a variational latent sampling.  All GRU-VAEs were trained with the same numbers and sizes of recurrent layers, with the only differences being the latent dimensions of 32, 64, 128 and 256---similar to the Linear VAE.  As in the earlier work of~\citeauthoryearp{summerville2016mariostring}, a top-to-bottom approach is used to turn the 2D levels into a 1D sequence. The encoder had 3 hidden layers of size 1024, and the decoder had 2 hidden layers of size 256---both had a dropout rate of 50\%.  To aid in the convergence of the model, the variational loss was annealed with a linear rate from 0 to 0.05 times the variational loss over 5 epochs before the rest of the training continued at that rate---for a total of 50 epochs using the Adam optimizer and a learning rate of $1e-5$. Note that the GRU-VAE is non-deterministic in its decoding.  At decoding time, the decoder is initialized with the latent embedding and then proceeds to decode in an auto-regressive manner with sampling.  For each generation, 10 segments are sampled and the one with the lowest perplexity (highest likelihood) is kept.                                               

\section{Evaluation}\label{sec:eval}

\subsection{Tile-based Metrics}
For each model, we generated 1000 segments and evaluated them against the training data with the following metrics:

\begin{itemize}
    \item \textit{Density}: proportion of a segment not occupied by background or path tiles
    \item \textit{Non-Linearity}: how well the topology of a segment fits to a line; measured by fitting a line to the topmost point of columns in a segment using linear regression
    \item \textit{Leniency}: the proportion of a segment not occupied by hazard tiles; this acts as a very simple proxy for difficulty
    \item \textit{Interestingness}: the proportion of a segment occupied by powerups, portals and collectables
    \item \textit{Path-Proportion}: the proportion of a segment occupied by the optimal path through it
\end{itemize}

\noindent For our evaluations, we computed the \textit{E-distance}~\cite{szekely2013energy} which measures the distance between two distributions. Summerville~\shortcite{summerville2018expanding} suggests E-distance as a metric for evaluating generative models. Lower values of E-distance imply higher similarity between distributions. We calculated E-distance with the above metrics for the generated segments against the training segments, per domain. The 5 metrics were combined into a 5-dimensional feature vector for each distribution and the E-distance was computed using these combined features.

\newcommand{\XTABLEtilemetrics}{\begin{table*}[t!]
\centering
\begin{tabular}{|c|c|c|c|c|c|}
\hline
Domain & Density & Nonlinearity & Leniency & Interestingness & Path-Prop \\
\hline
Castlevania & $18.44 \pm 14.04$    & $8.27 \pm 6.77$    & $99.14 \pm 3.69$ & $1.13 \pm 0.7$ & $7.58 \pm 1.88$  \\ %
Mega Man & $32.98 \pm 17.51$    & $6.16 \pm 4.97$    & $98.86 \pm 30.12$ & $0.09 \pm 0.13$ & $8.66 \pm 1.84$  \\
Metroid & $37.29 \pm 17.78$    & $0.74 \pm 3.48$    & $91.76 \pm 5.8$ & $0.47 \pm 1.23$ & $7.35 \pm 2.04$  \\ %
Mario   & $12.27 \pm 8.85$    & $10.42 \pm 6.05$    & $99.47 \pm 0.46$ & $0.66 \pm 0.96$ & $8.12 \pm 1.03$  \\ %
Ninja Gaiden & $13.95 \pm 6.71$    & $9.68 \pm 6.86$    & $99.47 \pm 0.43$ & $0.66 \pm 0.34$ & $8.12 \pm 1.27$  \\ %
\hline
\hline
Combined & $24.16 \pm 17.99$    & $6.43 \pm 6.76$    & $96.92 \pm 4.89$ & $0.57 \pm 0.96$ & $7.86 \pm 1.72$  \\
\hline
\hline
LIN-32 & $22.57 \pm 15.97$    & $9.01 \pm 8.13$    & $98.1 \pm 3.73$ & $0.25 \pm 0.56$ & $6.63 \pm 2$  \\ %
LIN-64 & $21.67 \pm 15.94$    & $8.57 \pm 8.03$ & $98.2 \pm 3.68$ & $0.28 \pm 0.62$ & $6.54 \pm 1.93$  \\ %
LIN-128 & $20.89 \pm 16.13$    & $9.36 \pm 8.31$    & $98.55 \pm 3.24$ & $0.23 \pm 0.55$ & $6.27 \pm 2.09$  \\ %
LIN-256 & $20.69 \pm 14.86$    & $8.82 \pm 7.98$    & $98.1 \pm 3.76$ & $0.29 \pm 0.59$ & $6.41 \pm 2.27$  \\ %
\hline
\hline
GRU-32 & $31.99 \pm 22$    & $5.81 \pm 6.4$    & $99.67 \pm 0.76$ & $0.19 \pm 0.35$ & $7.38 \pm 1.6$  \\ 
GRU-64 & $23.47 \pm 13.02$    & $8.55 \pm 7.17$    & $99.48 \pm 0.57$ & $0.25 \pm 0.33$ & $7.5 \pm 1.5$  \\
GRU-128 & $23.49 \pm 19.85$    & $5.55 \pm 5.65$    & $98.26 \pm 3.2$ & $0.56 \pm 0.66$ & $7.55 \pm 1.17$  \\ 
GRU-256 & $23.62 \pm 18.07$    & $5.49 \pm 5.9$    & $98.14 \pm 3.62$ & $0.51 \pm 0.65$ & $7.69 \pm 1.24$  \\ 
\hline
\end{tabular}
\caption{Tile metric evaluation}
\label{XTABLEtilemetrics}
\end{table*}
}

\newcommand{\XTABLEed}{\begin{table}[t!]
\centering
\begin{tabular}{|c|c|c|c|c|c|}
\hline
Model & CV & MM & Met & SMB & NG \\
\hline
LIN-32 & $1.26$    & $4.7$    & $\textbf{12.53}$ & $4.45$ & $3.44$  \\ %
LIN-64 & $\textbf{0.82}$    & $5.63$    & $13.67$ & $3.67$ & $2.68$  \\ %
LIN-128 & $0.92$    & $\textit{6.47}$    & $\textit{15.13}$ & $\textbf{3.13}$ & $\textbf{2.47}$  \\ %
LIN-256 & $\textbf{0.82}$    & $5.9$    & $14.07$ & $3.49$ & $2.55$  \\ %
\hline
GRU-64 & $\textit{1.89}$    & $\textbf{3.23}$    & $12.64$ & $\textit{6.33}$ & $\textit{4.9}$  \\ 
\hline
\end{tabular}
\caption{E-Distances (entire model vs each separate game)}
\label{XTABLEed}
\end{table}
}

\newcommand{\XTABLEedall}{\begin{table}[t!]
\centering
\begin{tabular}{|c|c|}
\hline
Model & E-distance\\ 
\hline
LIN-32 & $\mathit{ 0.58}$\\
LIN-64 & $0.6$\\
LIN-128 & $0.99$\\
LIN-256 &$0.88 $\\
\hline
GRU-32 & $2.28$\\
GRU-64 & $1.25$\\
GRU-128 & $0.46$\\
GRU-256 &$\mathbf{0.32}$\\
\hline
\end{tabular}
\caption{E-Distances between each model and all games together. $p = .01$ with 100 resamples for all models}
\label{XTABLEedall}
\end{table}
}

\newcommand{\XTABLEedgame}{\begin{table}[t!]
\centering
\begin{tabular}{|c|c|c|c|c|c|}
\hline
Model & CV & MM & Met & SMB & NG \\
\hline

LIN-32 & 1.66    & \textit{3.53}    & 12 & 4.76 & 3.88  \\ 
LIN-64 & 0.9    & 5.52    & \textit{10.6} & 3.1 & 2.54  \\
LIN-128 & \textit{0.86}    & 6.58    & 13.72 & 3.25 & \textit{2.29}  \\
LIN-256 & 0.97    & 5.92    & 14.14 & \textit{3.00} & 2.82  \\
\hline
GRU-32 & 1.22    & 4.16    &\textbf{ 0.4} & {0.38} & \textbf{1.22}  \\
GRU-64 & 1.15    &\textbf{ 2.4 }   & 0.89 & \textbf{0.34} & 1.9  \\ 
GRU-128 & 0.76    & 5.08    & 2.45 & 2.07 & 1.36  \\ 
GRU-256 &\textbf{ 0.66 }   & 4.81    & 0.52 & 3.13 & 1.78  \\ 
\hline
\end{tabular}
\caption{Game-specific E-distances. $ p = .01$ for all models using 100 resamples. \textit{Italics} denote the best for the specific NN architecture, and \textbf{Bolds} denote the best across all architectures.} 
\label{XTABLEedgame}
\end{table}
}

\newcommand{\XTABLEedcombined}{
\begin{table}[t!]
\centering
\small
\begin{tabular}{|c||c||c|c|c|c|c|}
\hline
Model & ALL & CV & MM & Met & SMB & NG \\
\hline

LIN32 & \textit{0.58} & 1.66    & \textit{3.53}    & 12 & 4.76 & 3.88  \\ 
-64 & 0.6 & 0.9    & 5.52    & \textit{10.6} & 3.1 & 2.54  \\
-128 & 0.99 & \textit{0.86}    & 6.58    & 13.72 & 3.25 & \textit{2.29}  \\
-256 & 0.88 & 0.97    & 5.92    & 14.14 & \textit{3.00} & 2.82  \\
\hline
\hline
GRU32 & 2.28 & 1.22    & 4.16    &\textbf{ 0.4} & {0.38} & \textbf{1.22}  \\
-64 & 1.25 & 1.15    &\textbf{ 2.4 }   & 0.89 & \textbf{0.34} & 1.9  \\ 
-128 & 0.46 & 0.76    & 5.08    & 2.45 & 2.07 & 1.36  \\ 
-256 & \textbf{0.32} &\textbf{ 0.66 }   & 4.81    & 0.52 & 3.13 & 1.78  \\ 
\hline
\end{tabular}
\caption{E-distances between each model and all games together as well as each game taken separately. For both cases, $ p < .01$ for all models, using 100 resamples. For separate games, \textit{Italics} denote the best for the specific NN architecture, and \textbf{Bolds} denote the best across all architectures.} 
\label{XTABLEedcombined}
\end{table}
}

\newcommand{\XTABLEedself}{\begin{table}[t!]
\centering
\begin{tabular}{|c|c|}
\hline
Model & E-distance\\
\hline
Castlevania & $0.07$\\
Mario & $0.02$\\
Mega Man & $0.04$\\
Metroid & $0.14$\\
Ninja Gaiden & $0.07$\\
\hline
\end{tabular}
\caption{E-Distances between different random samplings of each game}
\label{XTABLEedall}
\end{table}
}

\newcommand{\XTABLEfrechet}{\begin{table*}[t!]
\centering
\small
\begin{tabular}{c||ccccc|c}

& \multicolumn{5}{c|}{Domain} & \multirow{2}{2cm}{Agent Failure Rate}\\
Model   & CV             & MM             & Met            & SMB            & NG  & \\ \hline\hline
LIN-32  &$4.12\pm(2.66)$ &$4.83\pm(2.38)$ &$\mathit{4.59\pm(2.15)}$ &$5.24\pm(2.75)$ &$4.56\pm(2.67)$ &$11.46\%$\\
LIN-64  &$4.12\pm(2.49)$ &$4.85\pm(2.49)$ &$4.69\pm(2.16)$ &$5.23\pm(2.75)$ &$\mathit{4.41\pm(2.52)}$ &$11.72\%$\\
LIN-128 &$\mathit{3.87\pm(2.59)}$ &$\mathit{4.70\pm(2.49)}$ &$4.60\pm(2.38)$ &$5.12\pm(2.91)$ &$4.44\pm(2.61)$ &$\mathit{11.25\%}$ \\
LIN-256 &$4.06\pm(2.69)$ &$4.89\pm(2.54)$ &$4.74\pm(2.38)$ &$\mathit{5.01\pm(2.87)}$ &$4.54\pm(2.65)$ &$11.14\%$ \\ \hline\hline
GRU-32 &$\mathbf{1.81\pm(1.99)}$ &$2.18\pm(2.00)$ &$\mathbf{1.98\pm(1.74)}$ &$2.43\pm(2.14)$ &$\mathbf{1.81\pm(1.85)}$ &$\mathbf{4.52\%}$ \\
GRU-64 &$2.30\pm(2.34)$ &$2.65\pm(2.53)$ &$2.27\pm(2.17)$ &$3.25\pm(2.76)$ &$2.19\pm(2.28)$ &$5.12\%$ \\
GRU-128 &$1.94\pm(2.15)$ &$\mathbf{1.97\pm(2.25)}$ &$2.13\pm(2.13)$ &$\mathbf{2.02\pm(2.16)}$ &$1.98\pm(2.14)$ &$4.81\%$ \\
GRU-256 &$2.10\pm(2.22)$ &$2.24\pm(2.48)$ &$2.38\pm(2.25)$ &$2.26\pm(2.30)$ &$2.18\pm(2.27)$ &$4.55\%$ \\
\end{tabular}
\caption{Fr{\'e}chet distances for each model on each domain. Columns include blended levels using that domain. Values in cells are the average Fr{\'e}chet distance between the generated paths in the blended levels and the agent paths for the blended domains.}
\label{XTABLEfrechet}
\end{table*}
}

\newcommand{\XTABLEfrechetBest}{\begin{table*}[t!]
\centering
\small
\begin{tabular}{c||ccccc}
GRU-32 & CV & MM & Met & SMB & NG \\ \hline \hline
CV  & $1.76\pm(2.38)$ &$1.95\pm(2.12)$ & $1.63\pm(1.70)$ & $2.18\pm(2.19)$ & $1.51\pm(1.73)$\\
MM  & $1.95\pm(2.12)$ &$2.45\pm(1.83)$ & $2.02\pm(1.67)$ & $2.92\pm(2.36)$ & $1.81\pm(1.63)$\\
Met  & $1.63\pm(1.70)$ &$2.02\pm(1.67)$ & $2.21\pm(1.30)$ & $2.41\pm(1.70)$ & $1.74\pm(1.88)$\\
SMB  & $2.18\pm(2.19)$ &$2.92\pm(2.36)$ & $2.41\pm(1.70)$ & $2.69\pm(2.24)$ & $2.10\pm(20.12)$\\
NG  & $1.51\pm(1.73)$ &$1.81\pm(1.63)$ & $1.74\pm(1.88)$ & $2.10\pm(20.12)$ & $1.85\pm(1.51)$\\ \hline\hline
LIN-128 & CV & MM & Met & SMB & NG \\ \hline \hline
CV  & $2.79\pm(2.22)$ & $4.10\pm(2.53)$ & $3.92\pm(2.43)$ & $4.14\pm(2.84)$ & $3.57\pm(2.53)$\\
MM  & $4.10\pm(2.53)$ & $3.64\pm(1.98)$ & $4.56\pm(2.07)$ & $5.81\pm(2.80)$ & $4.57\pm(2.25)$\\
Met & $3.92\pm(2.43)$ & $4.56\pm(2.07)$ & $3.70\pm(1.55)$ & $5.30\pm(2.61)$ & $4.71\pm(2.35)$\\
SMB & $4.14\pm(2.84)$ & $5.81\pm(2.80)$ & $5.30\pm(2.61)$ & $5.62\pm(3.08)$ & $5.10\pm(3.01)$\\
NG  & $3.57\pm(2.53)$ & $4.57\pm(2.25)$ & $4.71\pm(2.35)$ & $5.10\pm(3.01)$ & $3.31\pm(2.34)$\\

\end{tabular}
\caption{Fr{\'e}chet distances for pairs of blended domains for the best performing Linear (LIN-128) and GRU model (GRU-32).}
\label{XTABLEfrechetBest}
\end{table*}
}

\XTABLEedcombined

\XTABLEfrechet
\XTABLEfrechetBest

\begin{figure*}[t]
\centering
\includegraphics[width=0.25\pdfpagewidth]{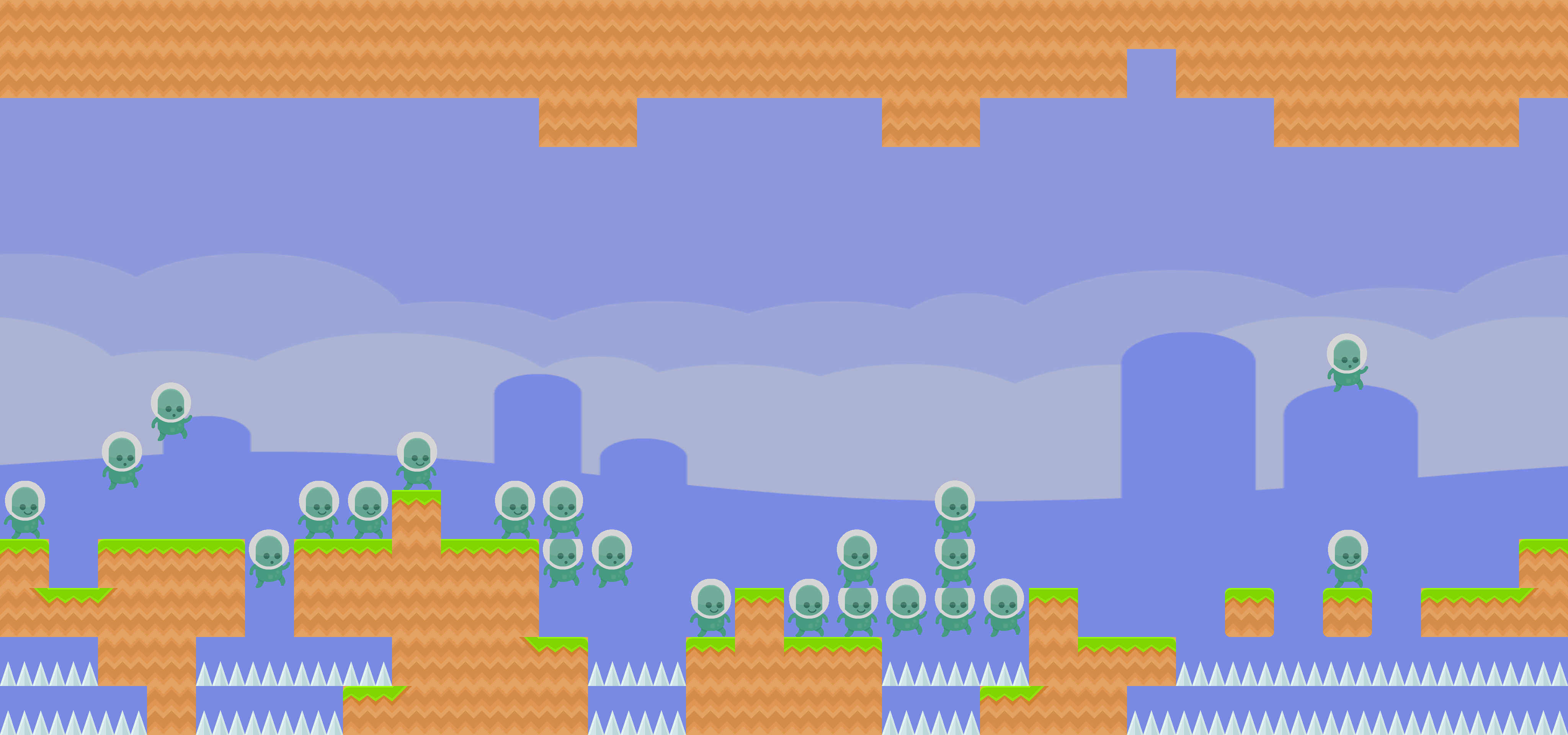}
~
\includegraphics[width=0.25\pdfpagewidth]{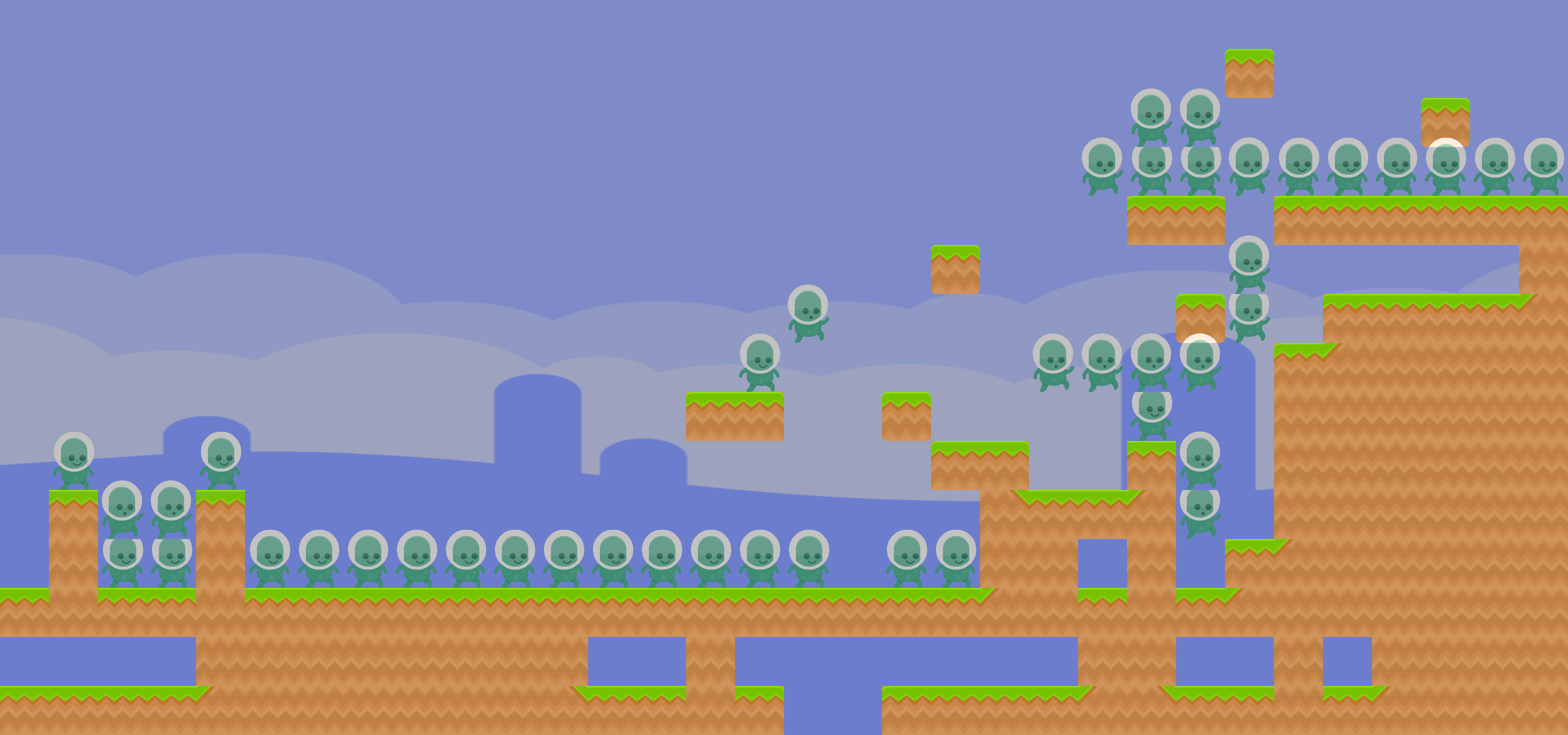}
~
\includegraphics[width=0.25\pdfpagewidth]{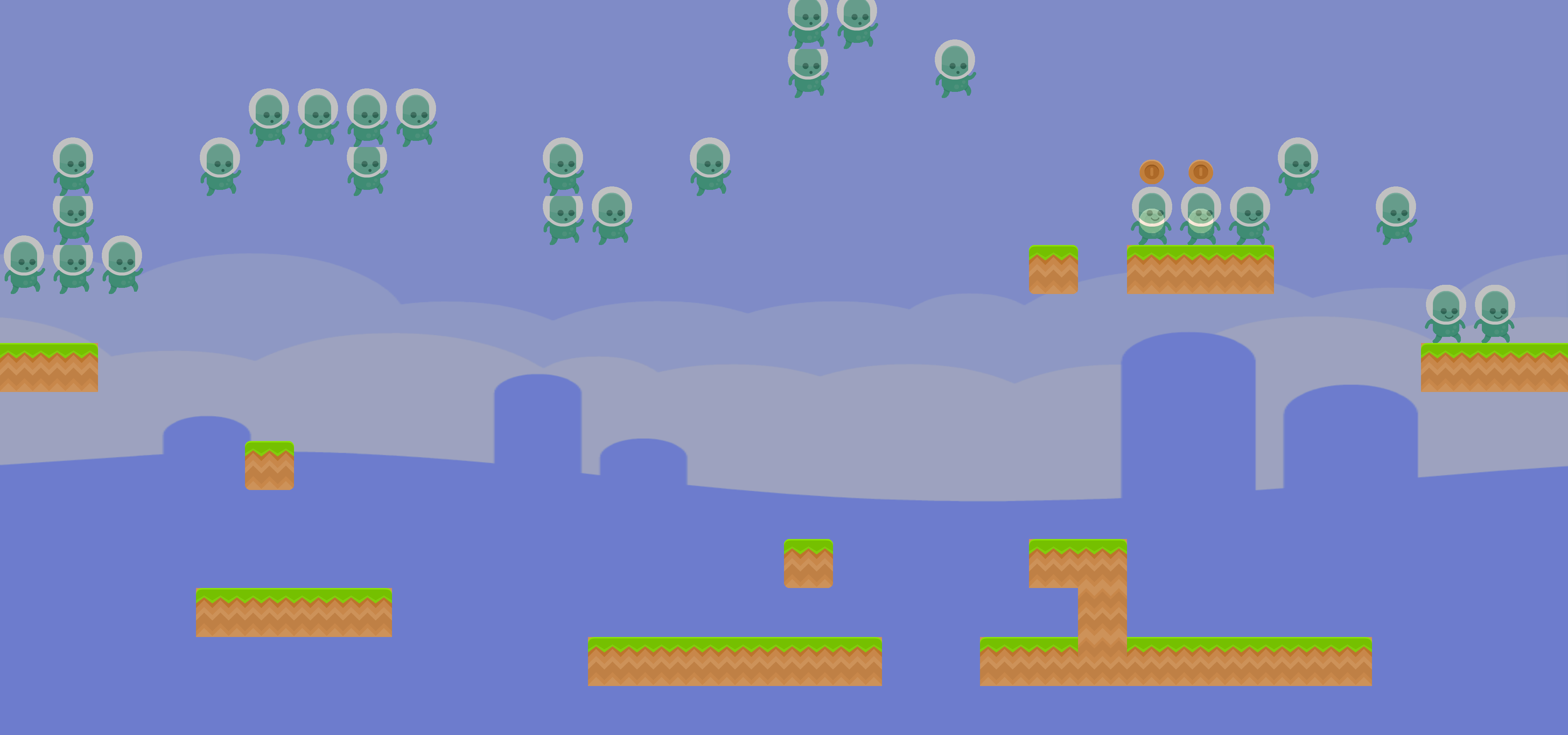}

\caption{Linear Samples}
\label{fig:lin_sample}
\end{figure*}

\begin{figure*}[t]
\centering
\includegraphics[width=0.25\pdfpagewidth]{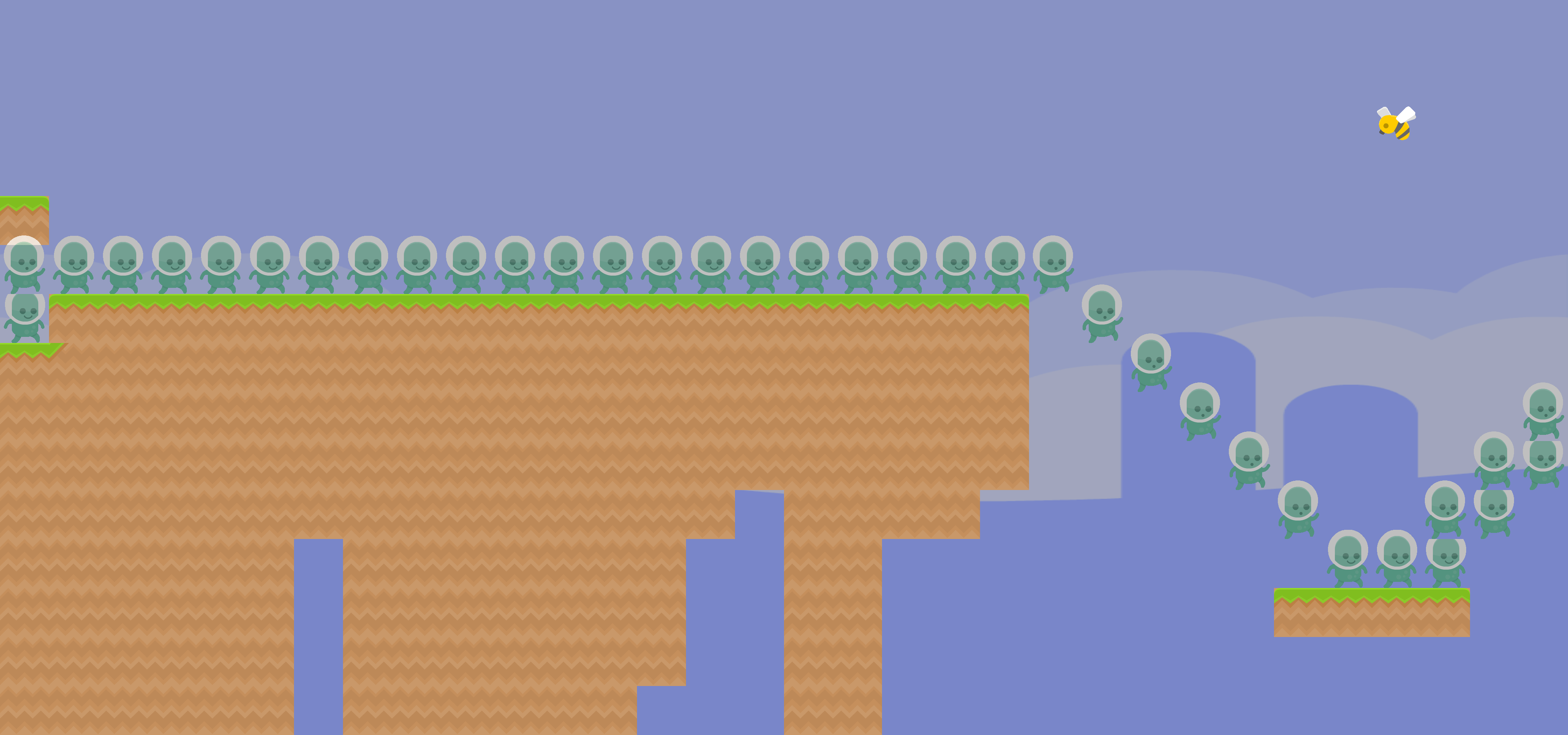}
~
\includegraphics[width=0.25\pdfpagewidth]{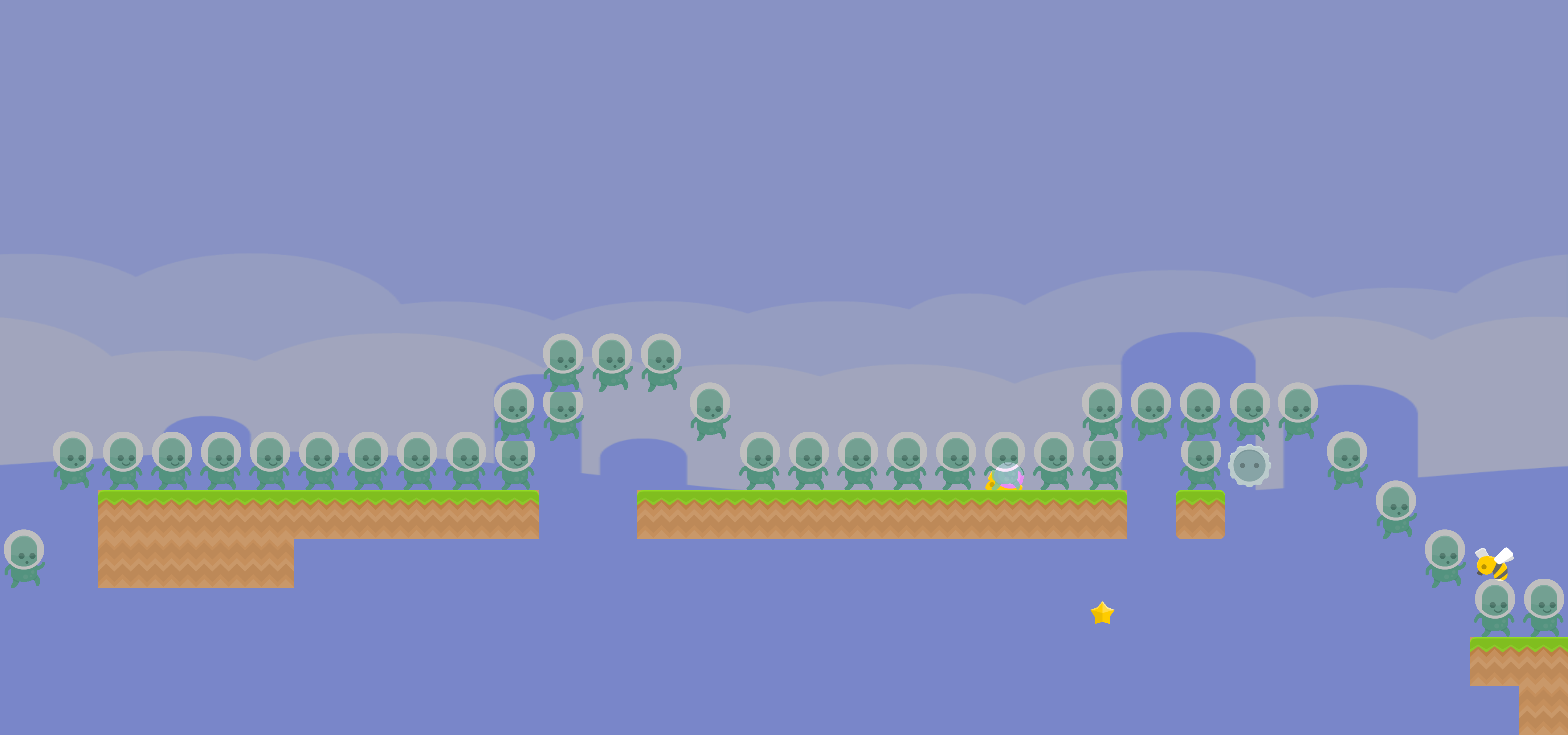}
~
\includegraphics[width=0.25\pdfpagewidth]{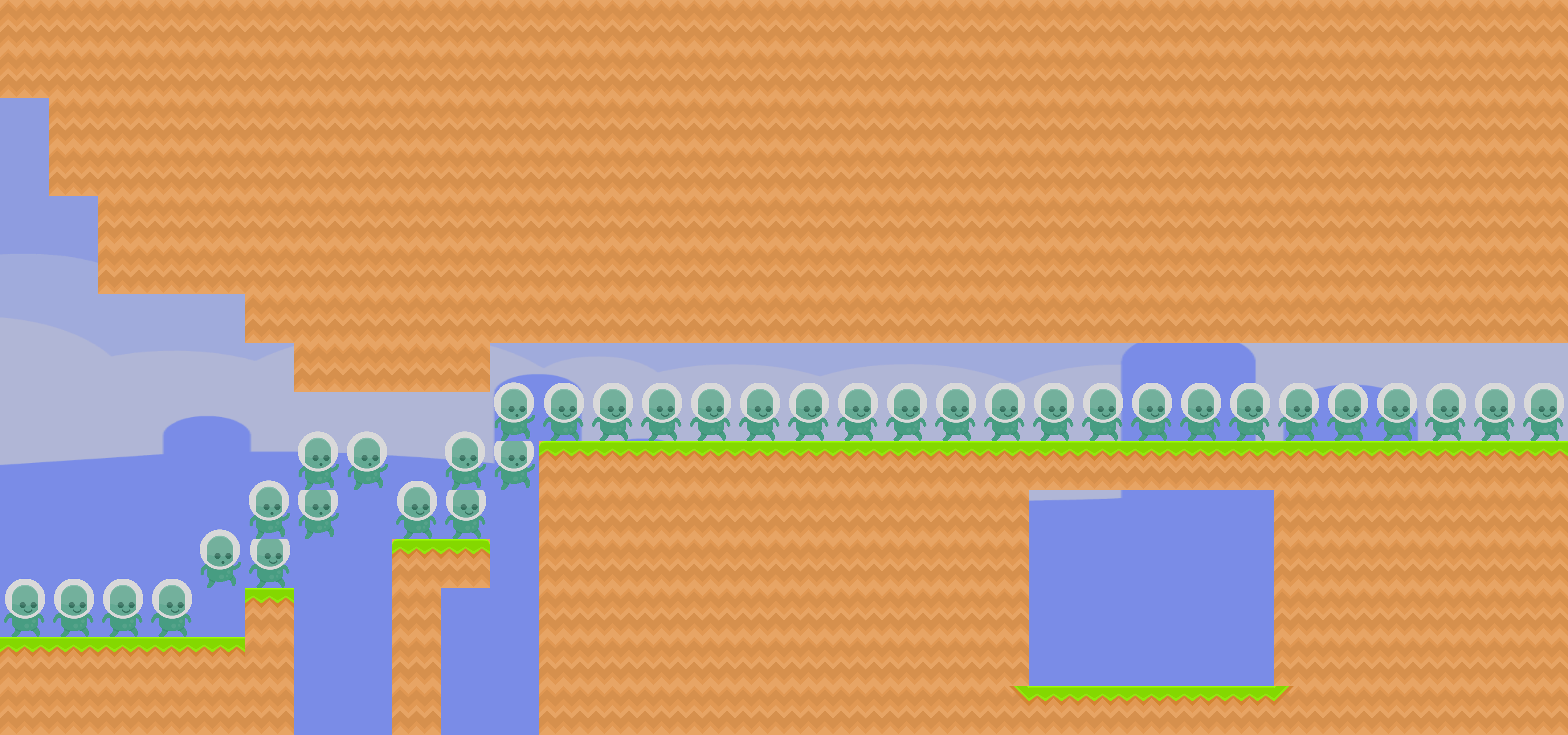}
\caption{GRU Samples}
\label{fig:gru_sample}
\end{figure*}

\subsection{Agent-based Evaluation}
We tested our models' ability to blend domains by interpolating between level segments. For this, we randomly selected $10$ level segments from each domain, passed each segment through the encoder of the VAE to get the latent vector representation of that segment, and then interpolated between such latent vectors in increments of $25\%$ (e.g., $25\%$ domain A, level $n$ and $75\%$ domain B, level $m$). We then generated level segments by forwarding interpolated latent vectors through the VAE decoder. Recall that the models generate sections annotated with beliefs about the path through the section. To evaluate an interpolated section, we compared the generated path to the paths found by A* agents for each of the blended domains by computing the discrete Fr{\'e}chet distance~\cite{eiter1994computing} between the generated paths and the agents' paths. The Fr{\'e}chet distance has been used previously to measure the similarity of agent paths~\cite{snodgrass2017procedural}; it can be thought of as the length of rope needed to connect two people walking on separate paths over the entirety of the paths. Agents that cannot traverse the entire level segment are said to have failed and their attempts are not included in this metric.


The Fr{\'e}chet distance between the generated and agent paths in interpolated segments gives insight into how well such segments capture both pathing and structural information in the component domains. Intuitively, if a segment is meant to be $75\%$ \textit{MM} and $25\%$ \textit{NG}, we expect the \textit{MM} A* agent to find a more similar path to the generated path. Additionally, we can examine which domains our model struggles with by comparing the distances across domains e.g., determining if \textit{SMB} A* paths are typically more similar to the generated paths than \textit{Met} A* paths.

\section{Results}\label{sec:res}
The E-distances between the samples from each of the generators and the original games taken together and separately are shown in Table \ref{XTABLEedcombined}. The distributions for the generated segments were all statistically different from the original games ($p < 0.01$) meaning that no generator successfully replicated the expressive range of the original levels; however, we see that when compared across all games, the linear models generally have lower distances, although the two largest GRU models surpass all linear models. We also see how each generator does when primed to generate segments that are targeted to be like a specific game, done by sampling the latent space using the means and variances of the latent encodings of the segments for that game. In this case, the GRU models are generally closer to the original games than the linear models, although no one model dominates. For e.g., GRU-32 and GRU-64 both outperform all linear models for \textit{Met}, \textit{SMB} and \textit{NG} with GRU-64 also doing better than all linear models for \textit{MM} but both these GRU models do worse than 3 out of the 4 linear models for \textit{CV}. Similarly, we do not see any particular advantage for latent size when compared using games separately, unlike when compared against all taken together. The GRU-32 and GRU-64 models are tied for the most games that they are closest to, but GRU-256 still beats both for \textit{CV}.

A key contribution of this work is the inclusion of agent paths across multiple games, with the goal being the development of novel physics models for blended content. Toward that end, we assess how well agents designed for the input games perform on levels created by these models. Table \ref{XTABLEfrechet} shows the Fr{\'e}chet distance between the generated path from game-specific samples and agents capable of playing that game. The agent failure rate is the percentage of generated levels that the A* agent failed to complete. The GRU models do better than the linear models in generating paths that are more similar to those in the target domains---with a commensurate decrease in agent failure rate. This is expected as a key advantage of recurrent models is the memory during decoding which allows them to remember where they have generated paths so far, thus better generating connected, sensible paths. The best performing model is GRU-32 with the lowest failure rate and the lowest Fr{\'e}chet distances for three of the five games. The best performing linear model is LIN-128 with the lowest failure rate amongst the linear models and the lowest scores for two of five games. 

Fr{\'e}chet distances for blended levels generated by GRU-32 and LIN-128 are shown in Table \ref{XTABLEfrechetBest}. Distances for the GRU model are lower than the linear model, across the board; however, we see some commonalities. The distances for games blended with \textit{SMB} are higher than other games across both models. Additionally, the blends for both \textit{CV} and \textit{NG} produce lower distances across models---both have more predictable jump physics (players do not control jump height) so their physics models may be simpler to learn. 

\begin{figure*}[h!]
\centering
\begin{subfigure}{.18\textwidth}
\centering
\includegraphics[width=0.12\pdfpagewidth]{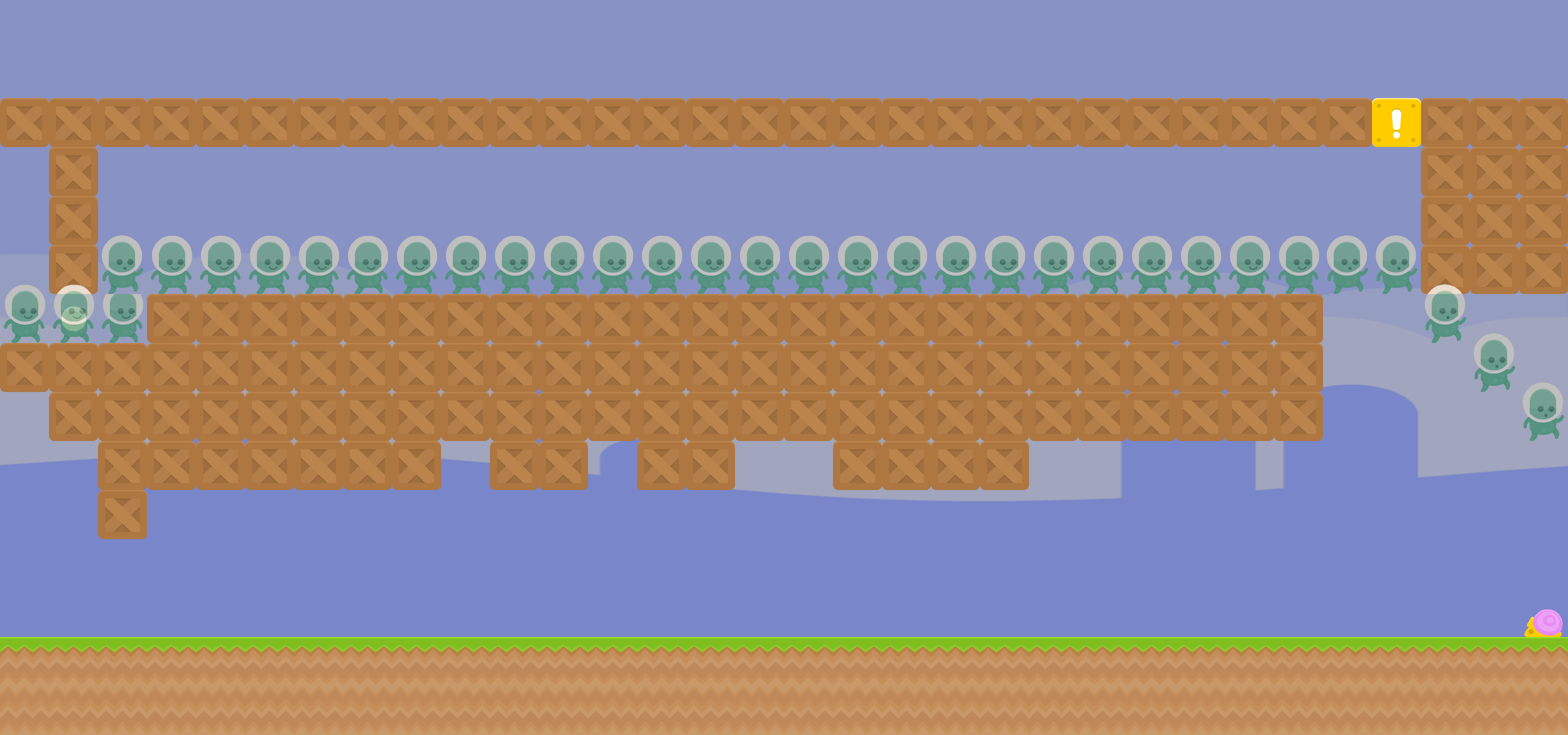}

\includegraphics[width=0.12\pdfpagewidth]{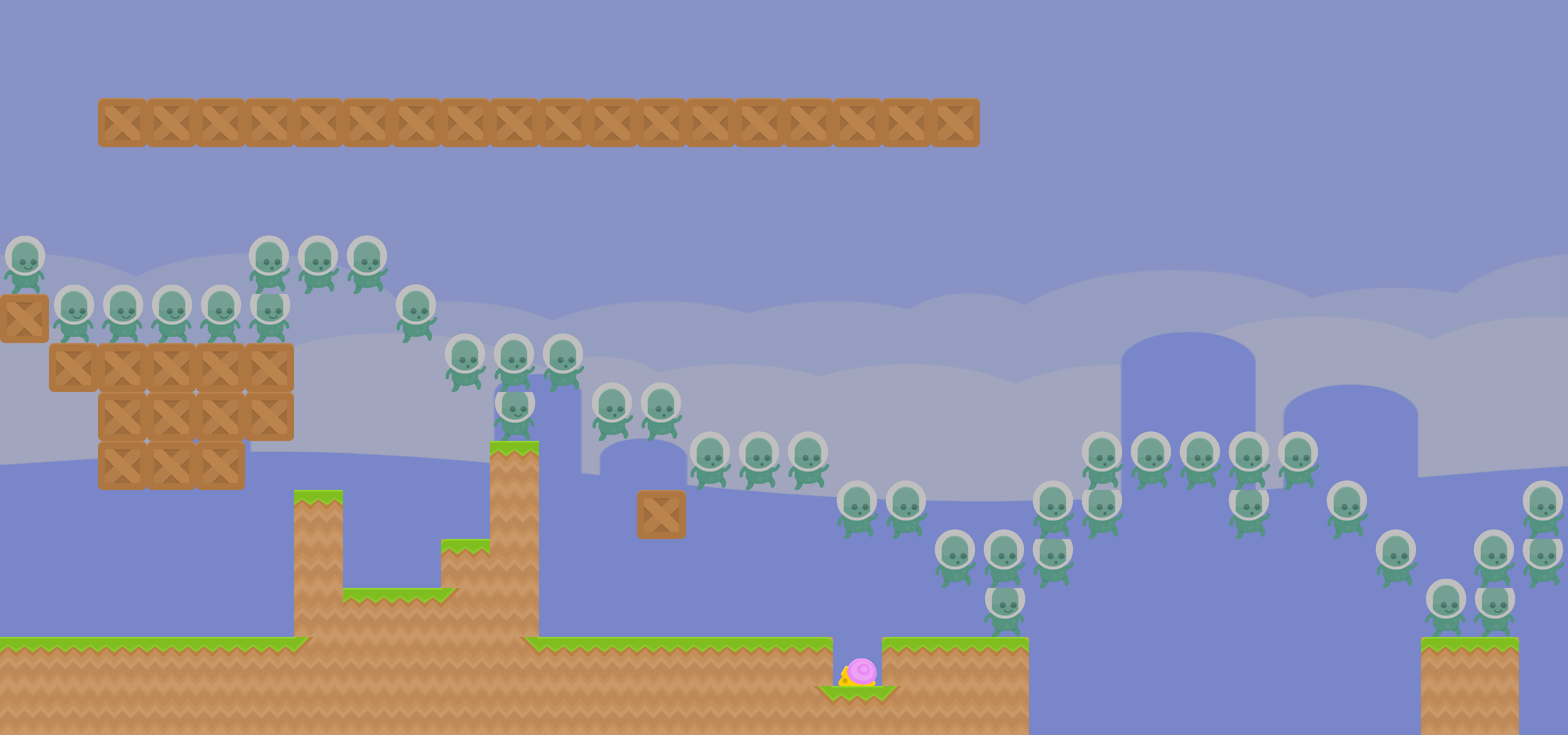}

\includegraphics[width=0.12\pdfpagewidth]{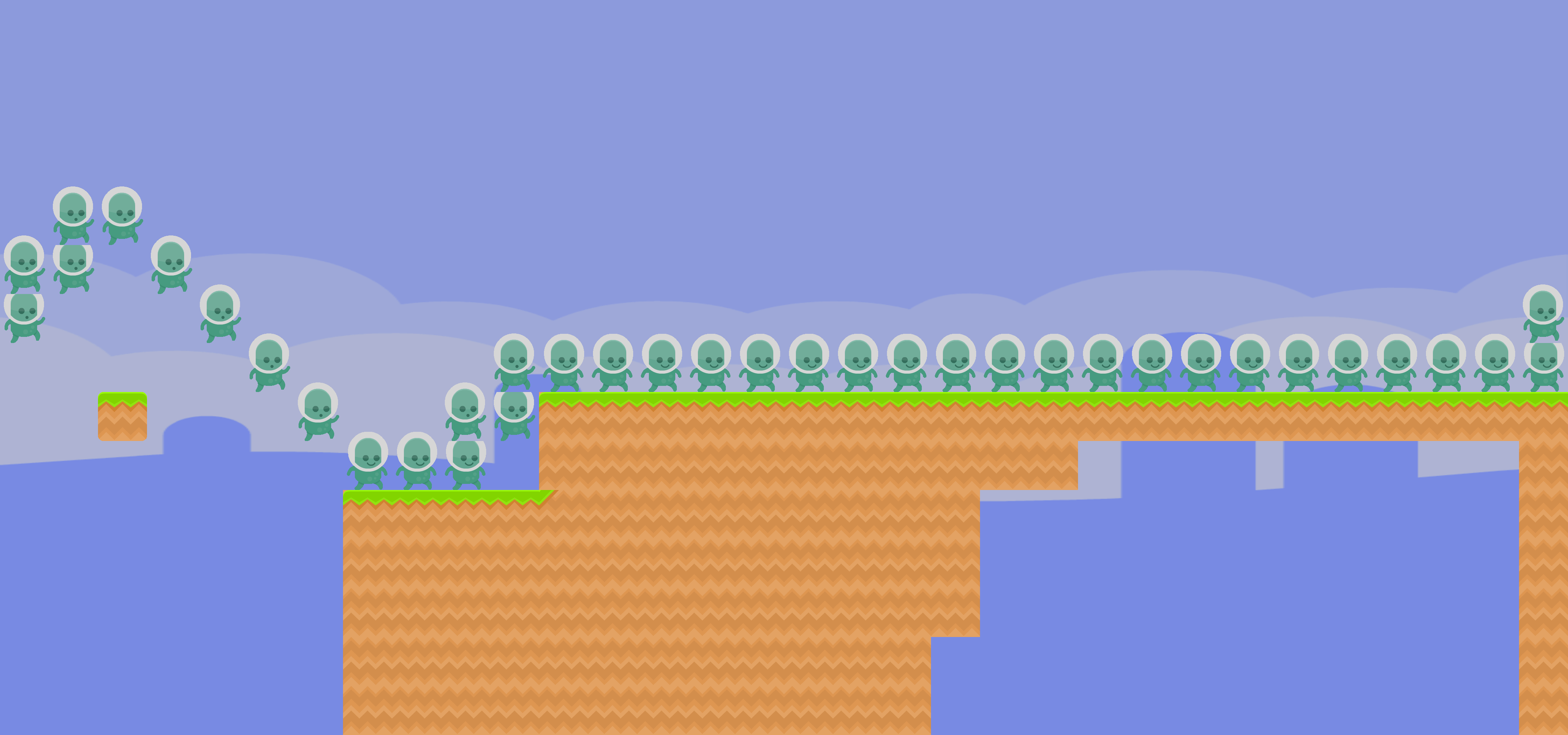}

\includegraphics[width=0.12\pdfpagewidth]{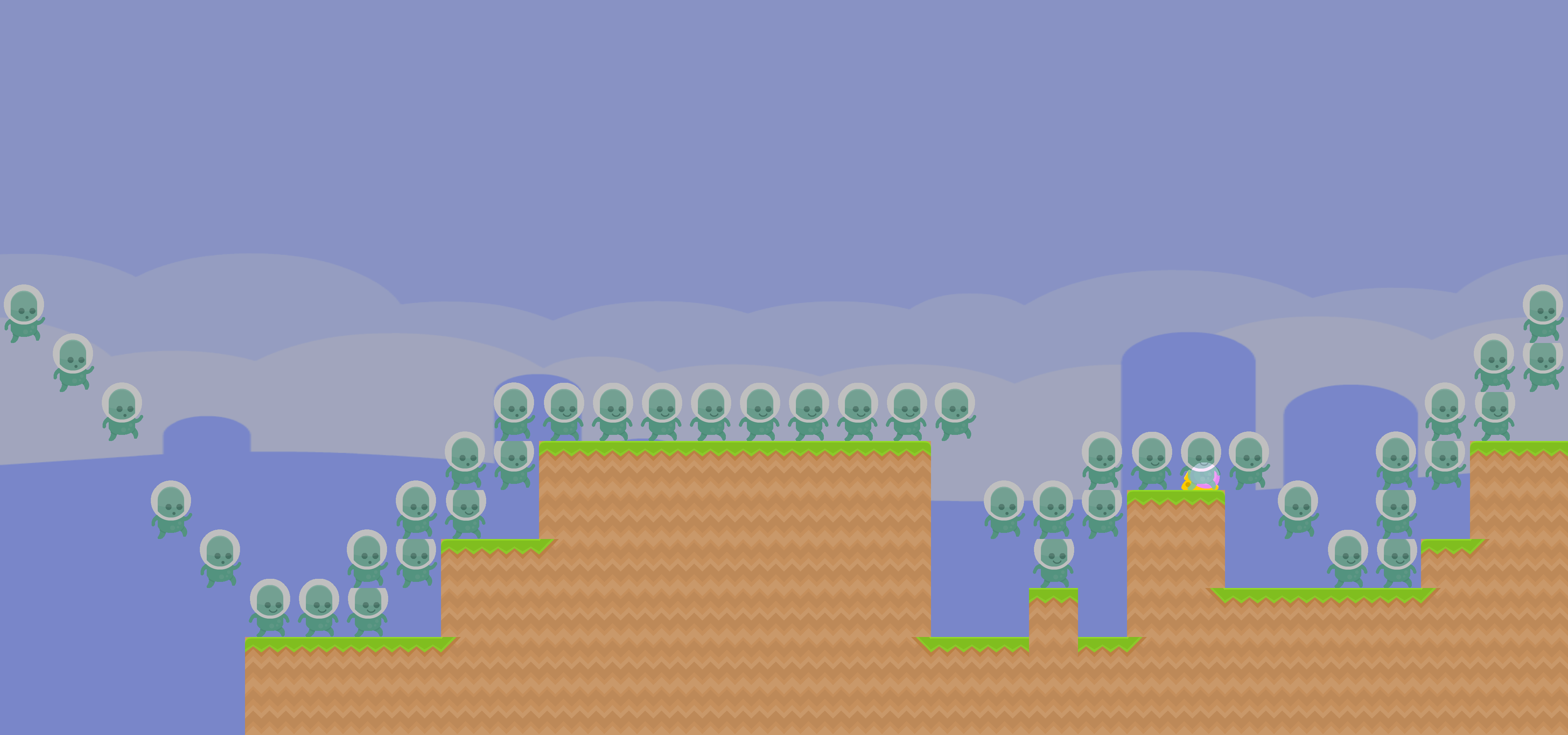}

\includegraphics[width=0.12\pdfpagewidth]{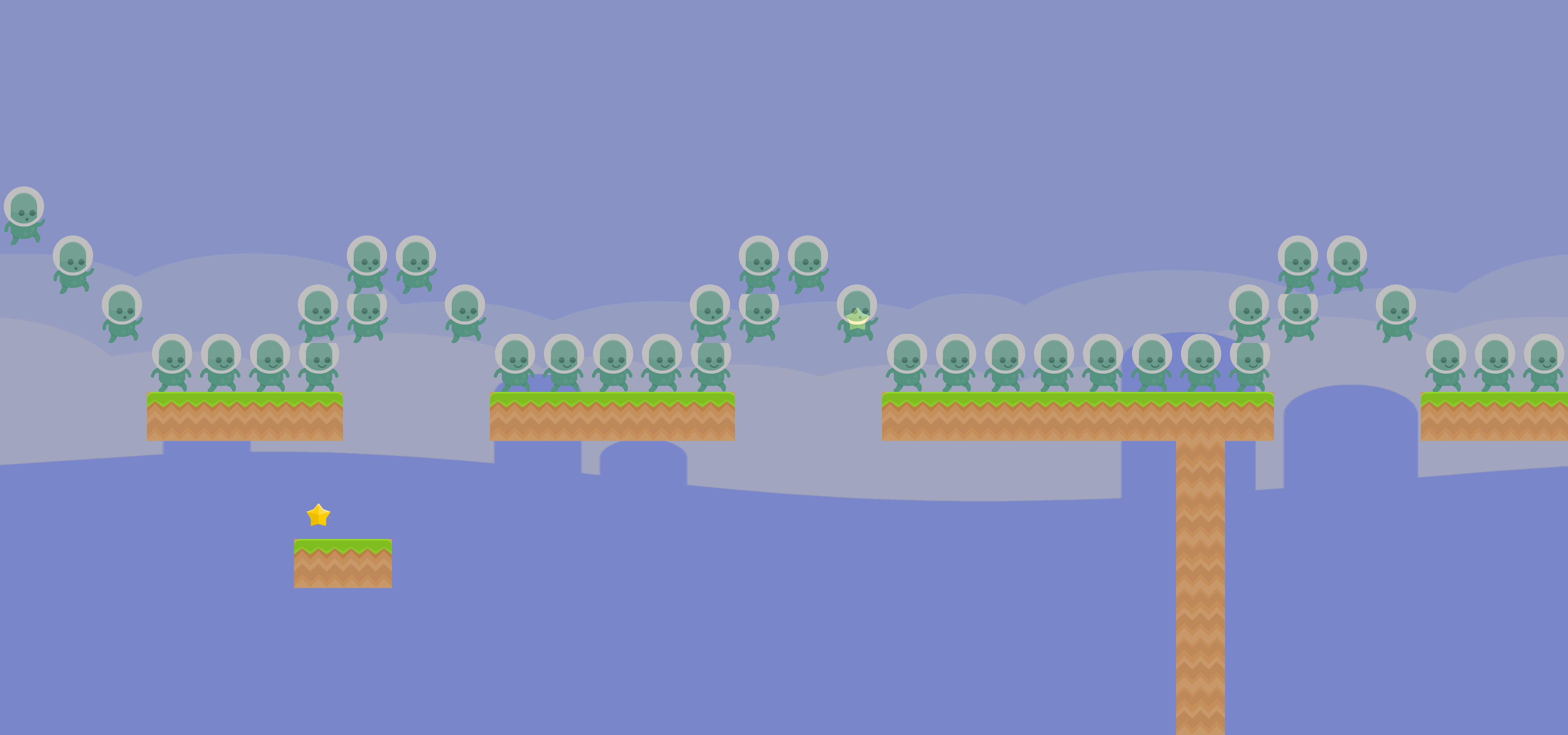}
\caption{GRU  SMB $\downarrow$ MM}
\label{fig:gru_interp_Mario_Megaman}
\end{subfigure}%
\begin{subfigure}{.18\textwidth}
\centering

\includegraphics[width=0.12\pdfpagewidth]{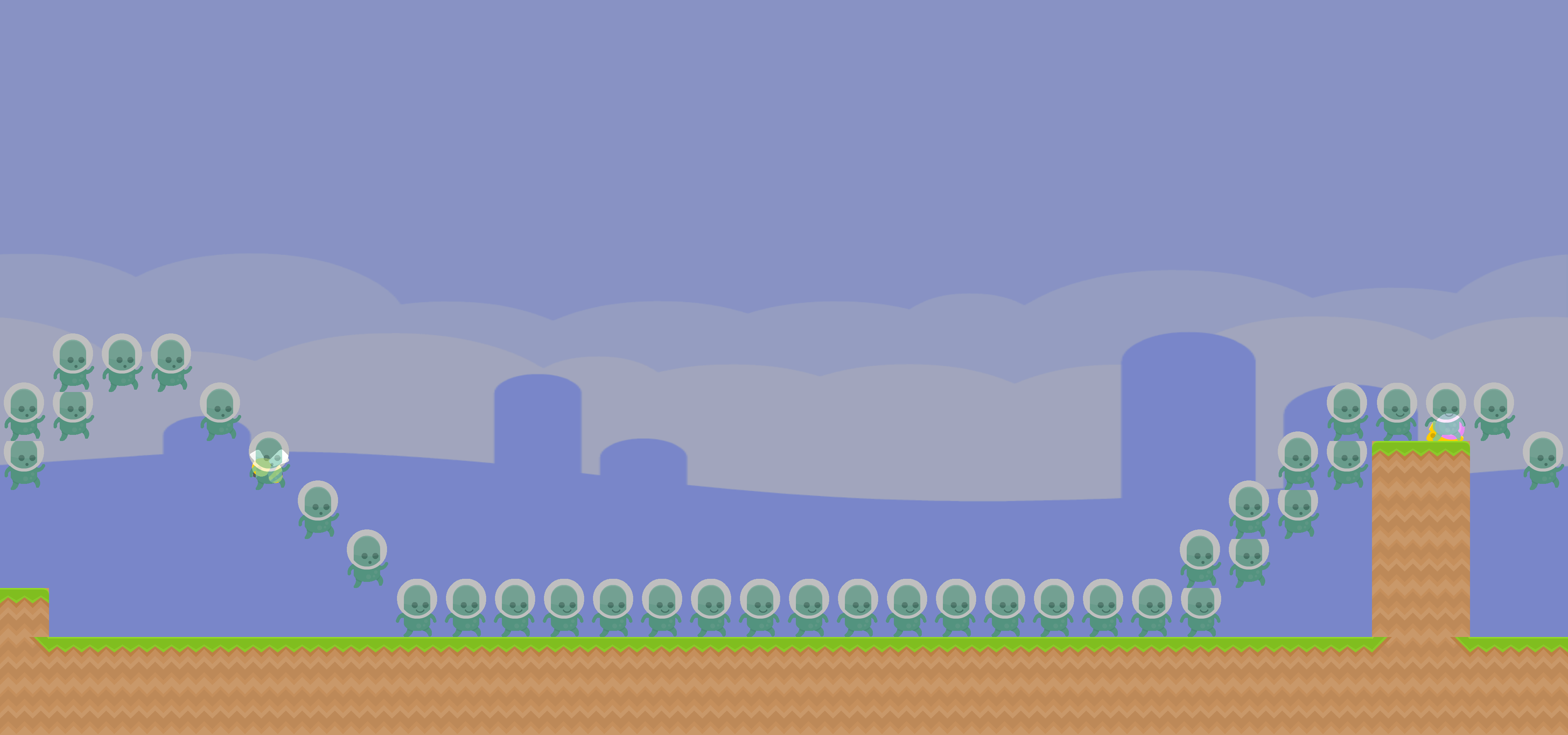}

\includegraphics[width=0.12\pdfpagewidth]{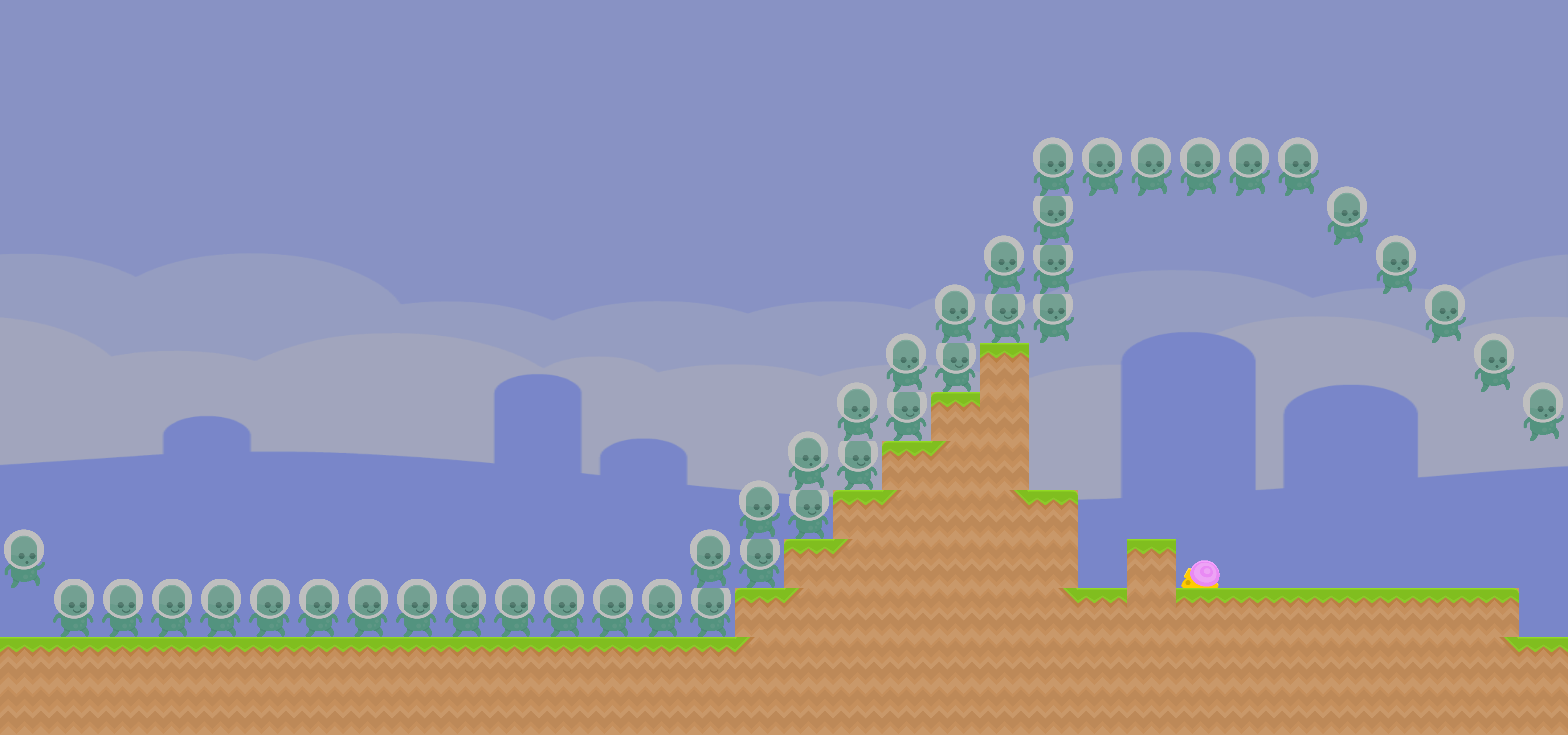}

\includegraphics[width=0.12\pdfpagewidth]{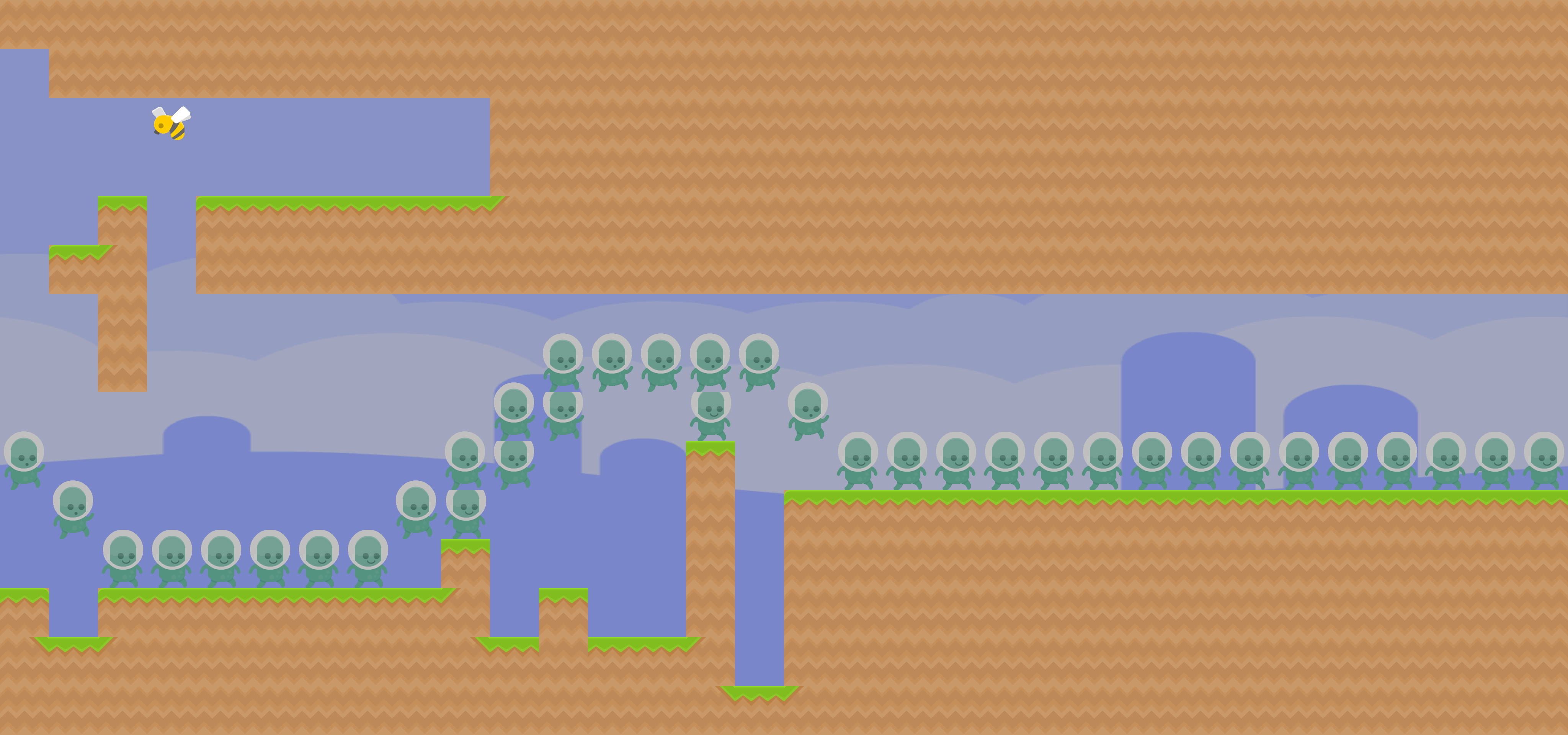}

\includegraphics[width=0.12\pdfpagewidth]{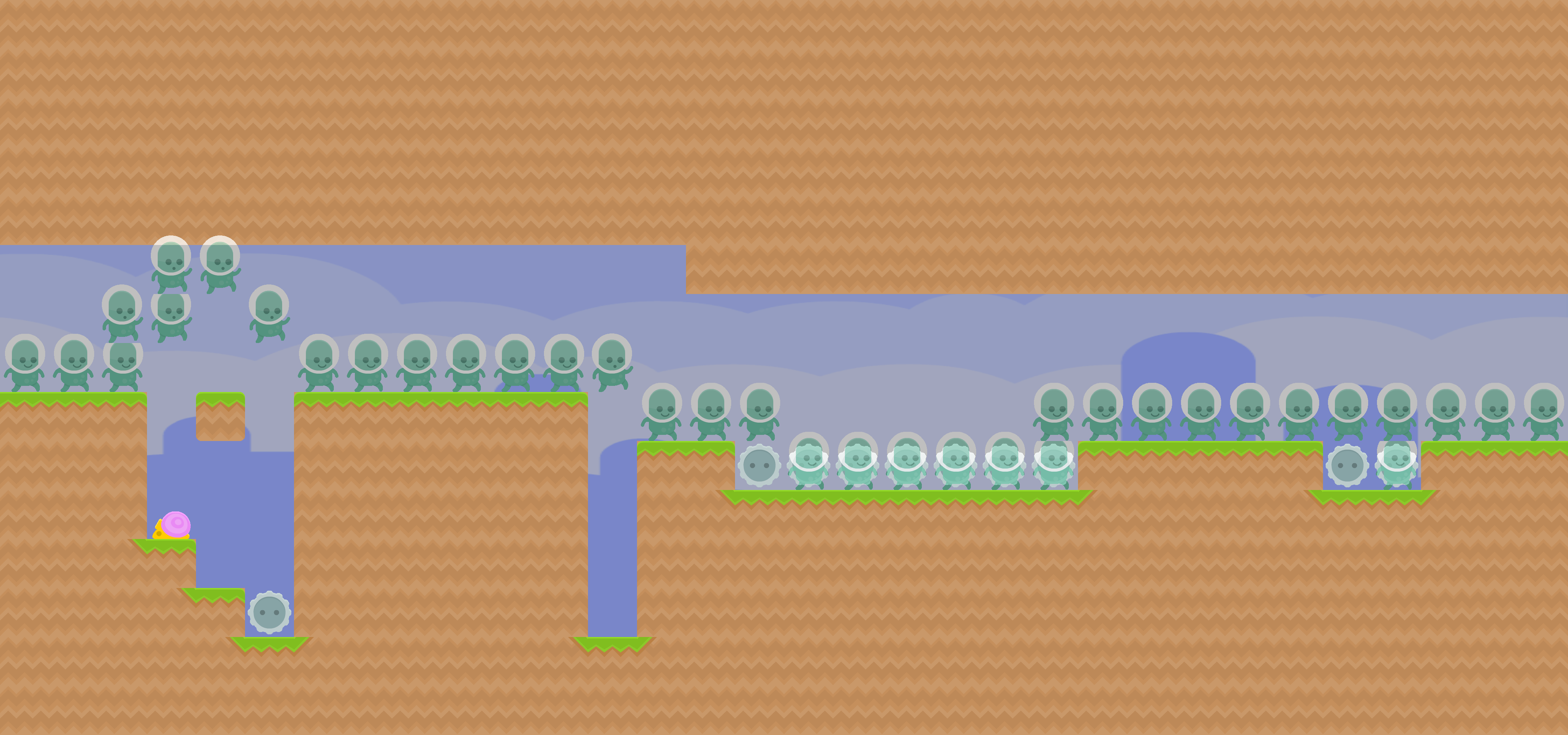}

\includegraphics[width=0.12\pdfpagewidth]{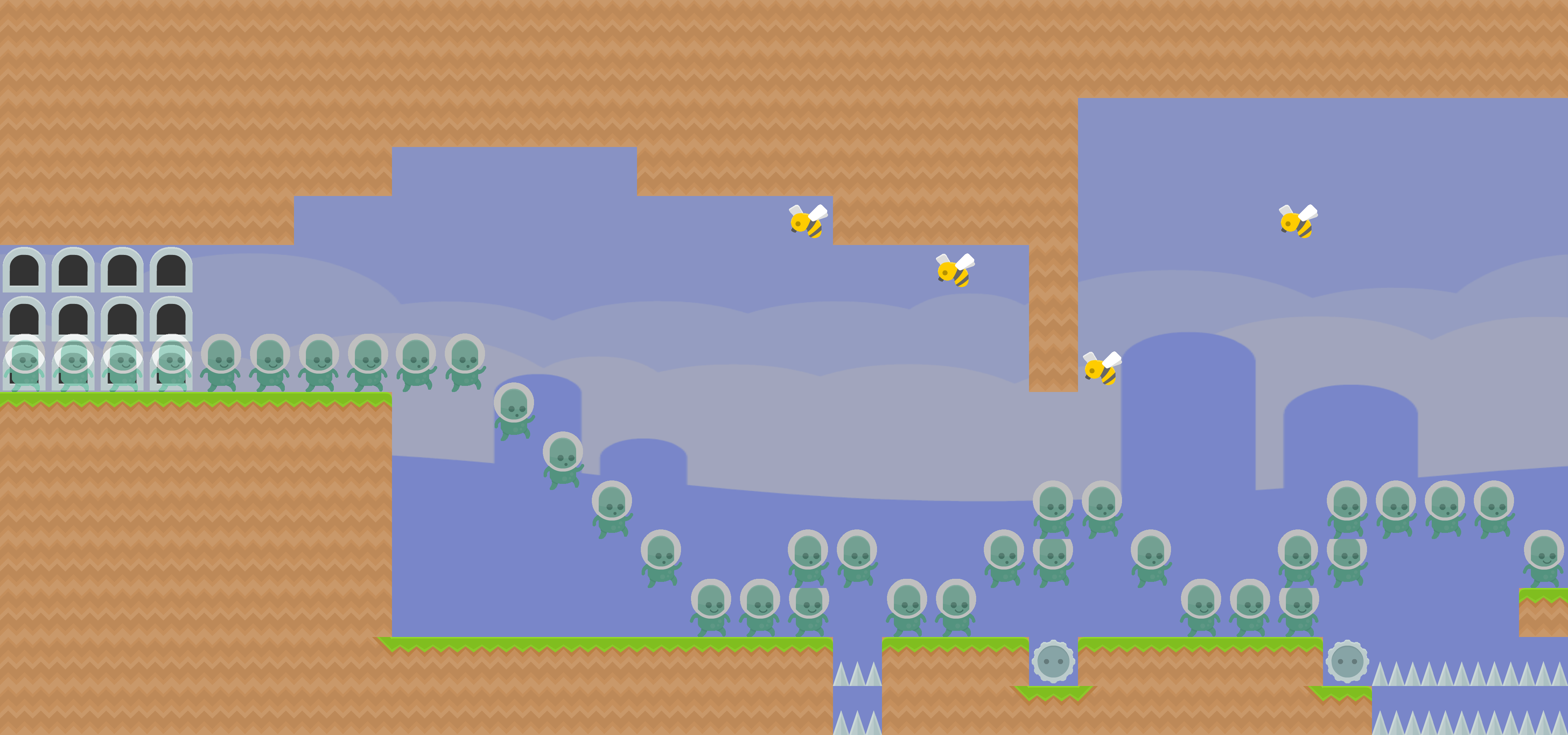}
\caption{GRU  SMB $\downarrow$ Met}
\label{fig:gru_interp_Mario_Metroid}
\end{subfigure}%
\begin{subfigure}{.18\textwidth}
\centering
\includegraphics[width=0.12\pdfpagewidth]{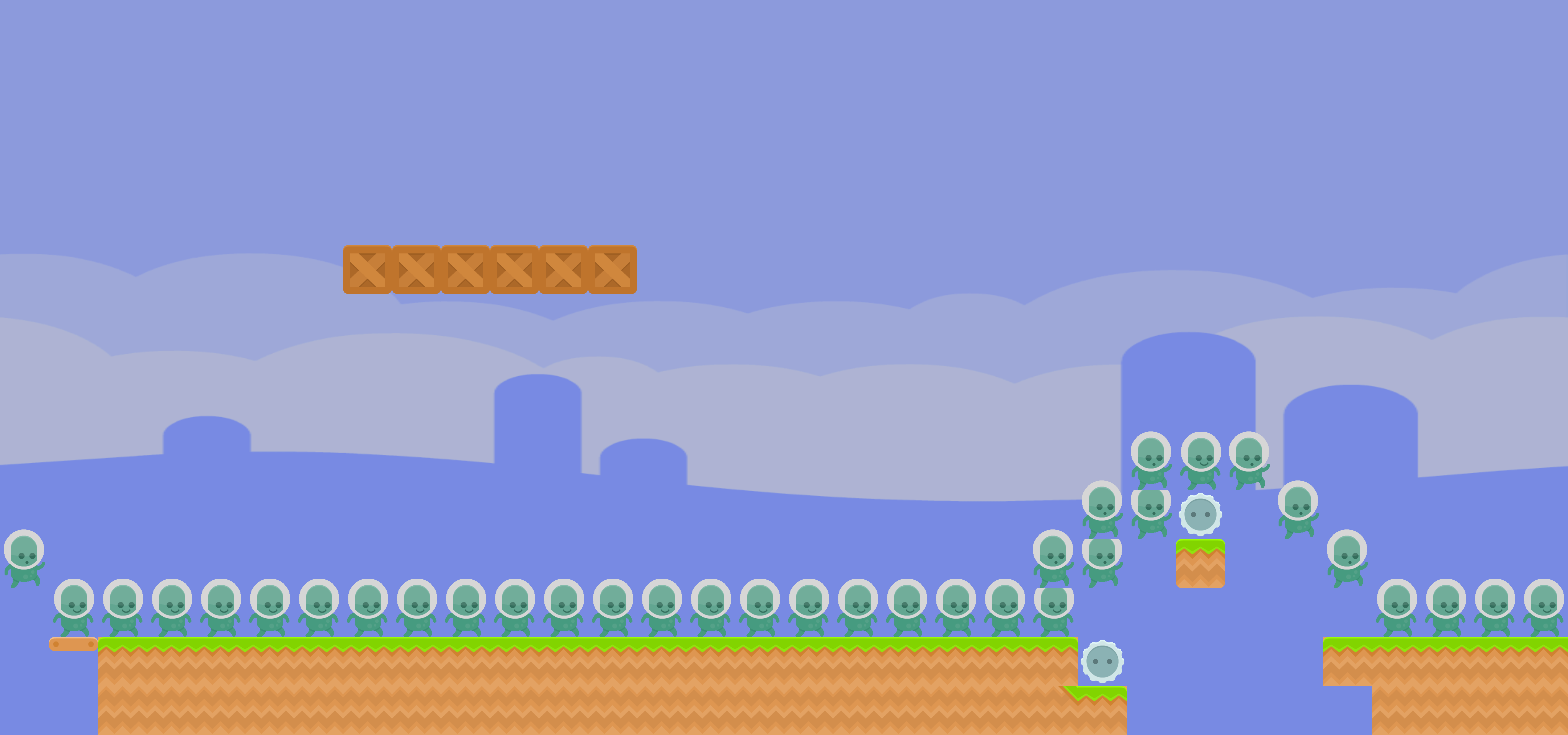}

\includegraphics[width=0.12\pdfpagewidth]{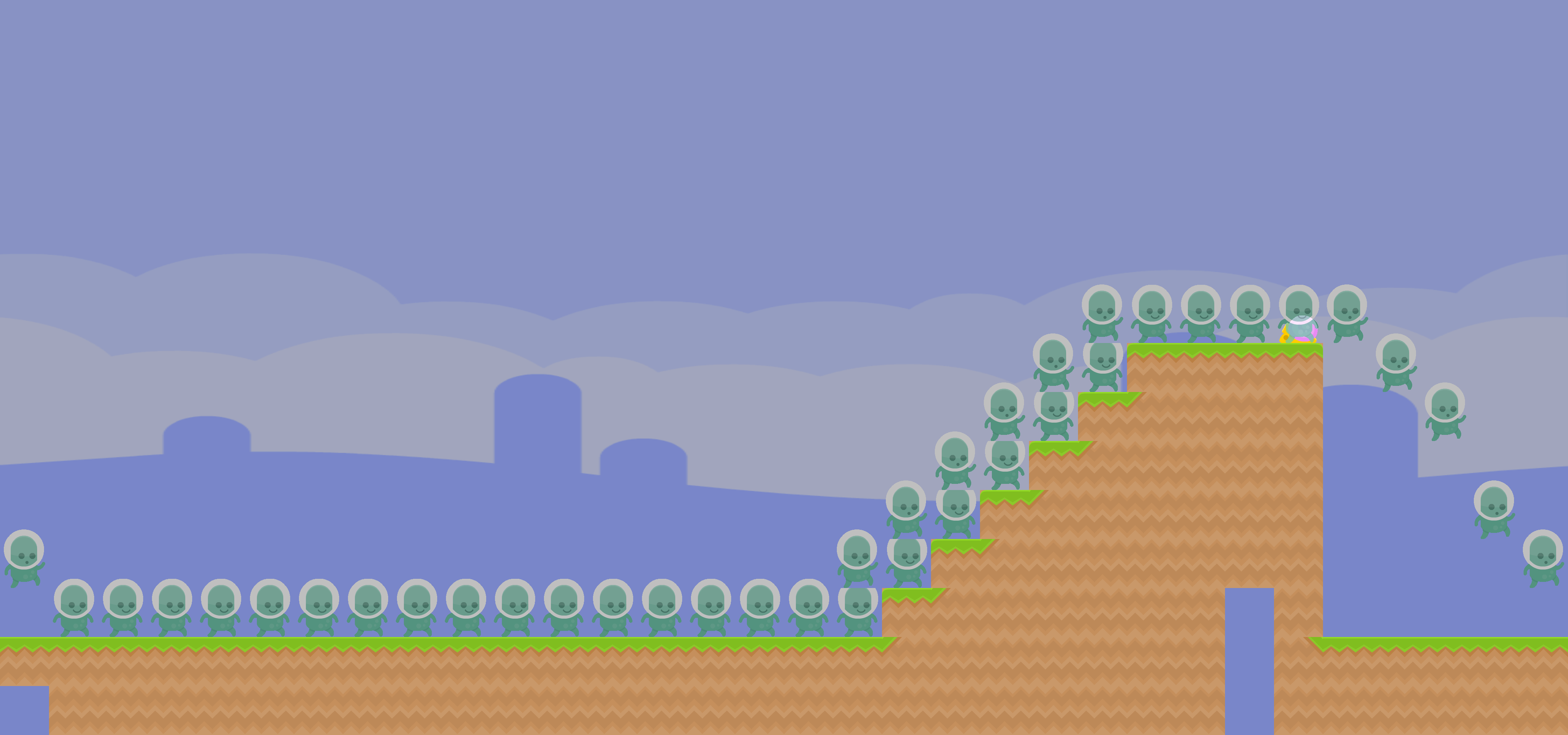}

\includegraphics[width=0.12\pdfpagewidth]{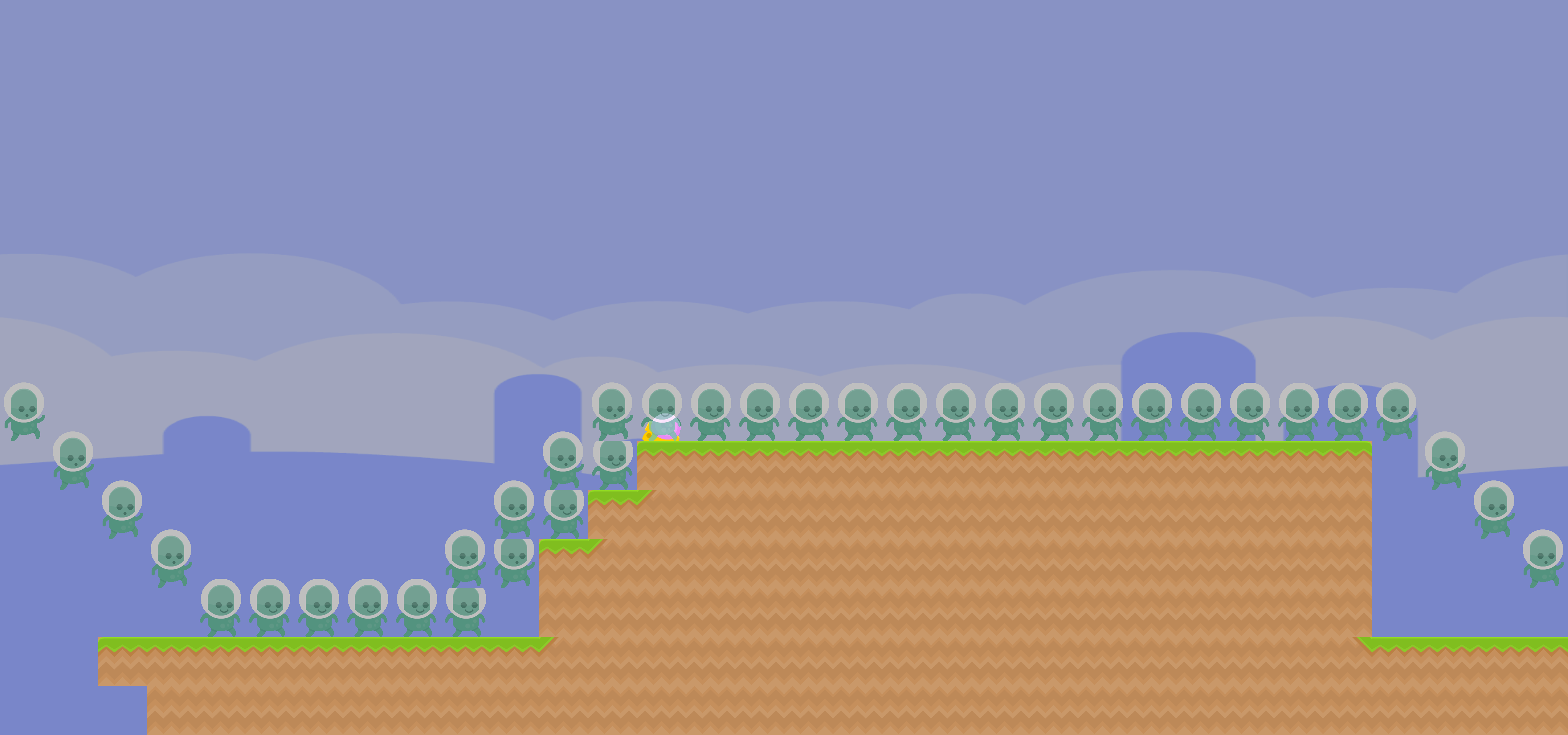}

\includegraphics[width=0.12\pdfpagewidth]{GRU/Mario_983-txt-NinjaGaiden_3379-txt-75_small}

\includegraphics[width=0.12\pdfpagewidth]{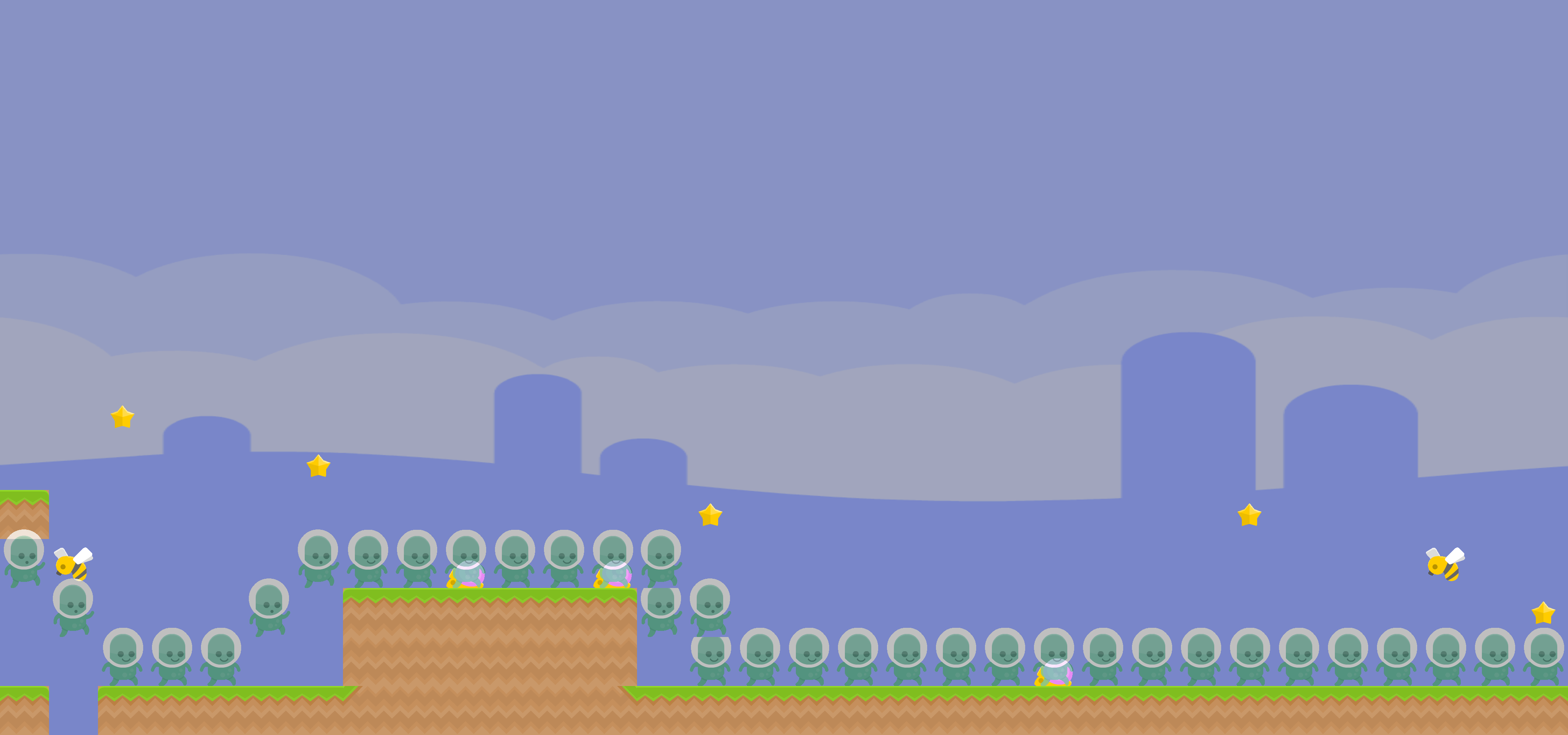}
\caption{GRU  SMB $\downarrow$ NG}
\label{fig:gru_interp_Mario_NG}
\end{subfigure}%
\begin{subfigure}{.18\textwidth}
\centering
\includegraphics[width=0.12\pdfpagewidth]{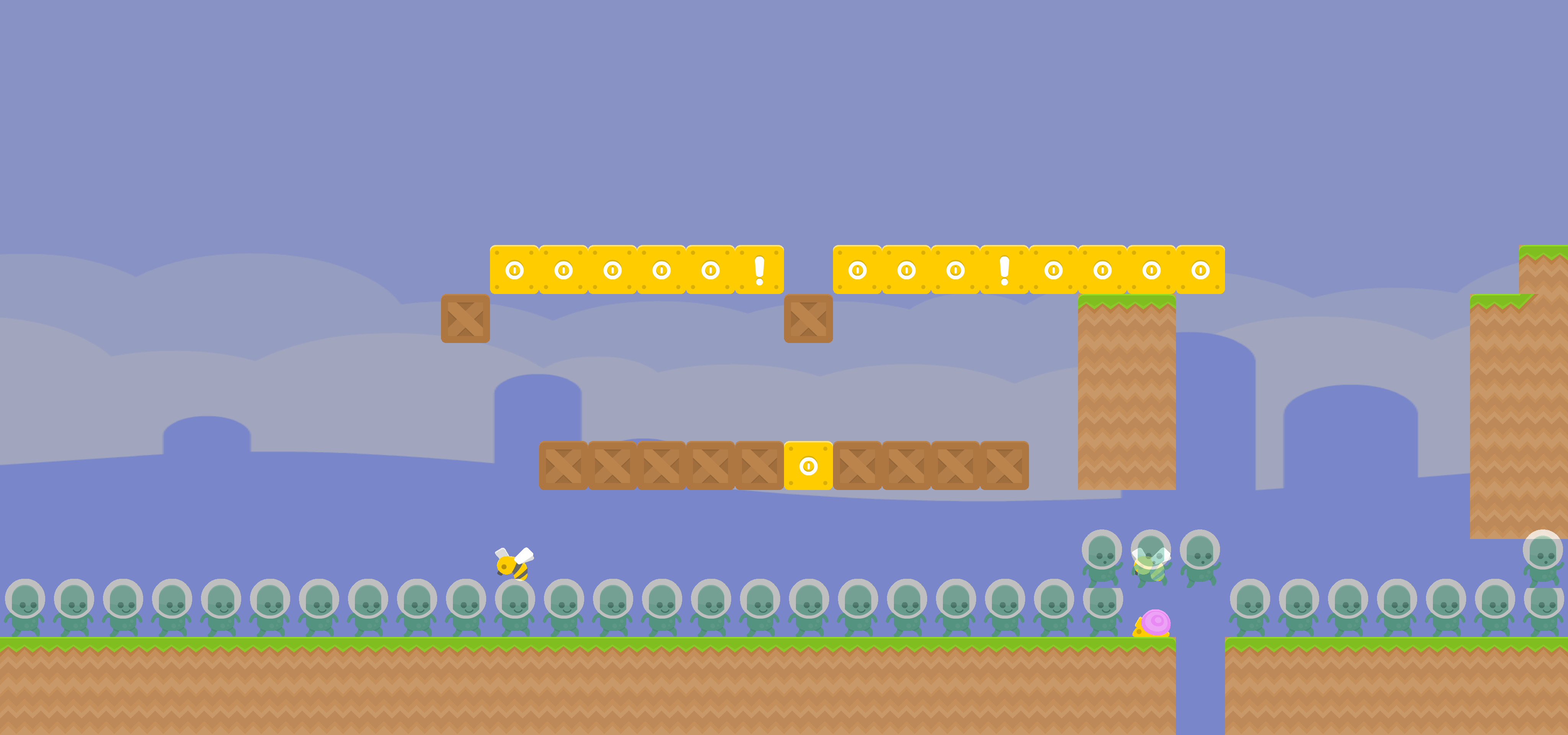}

\includegraphics[width=0.12\pdfpagewidth]{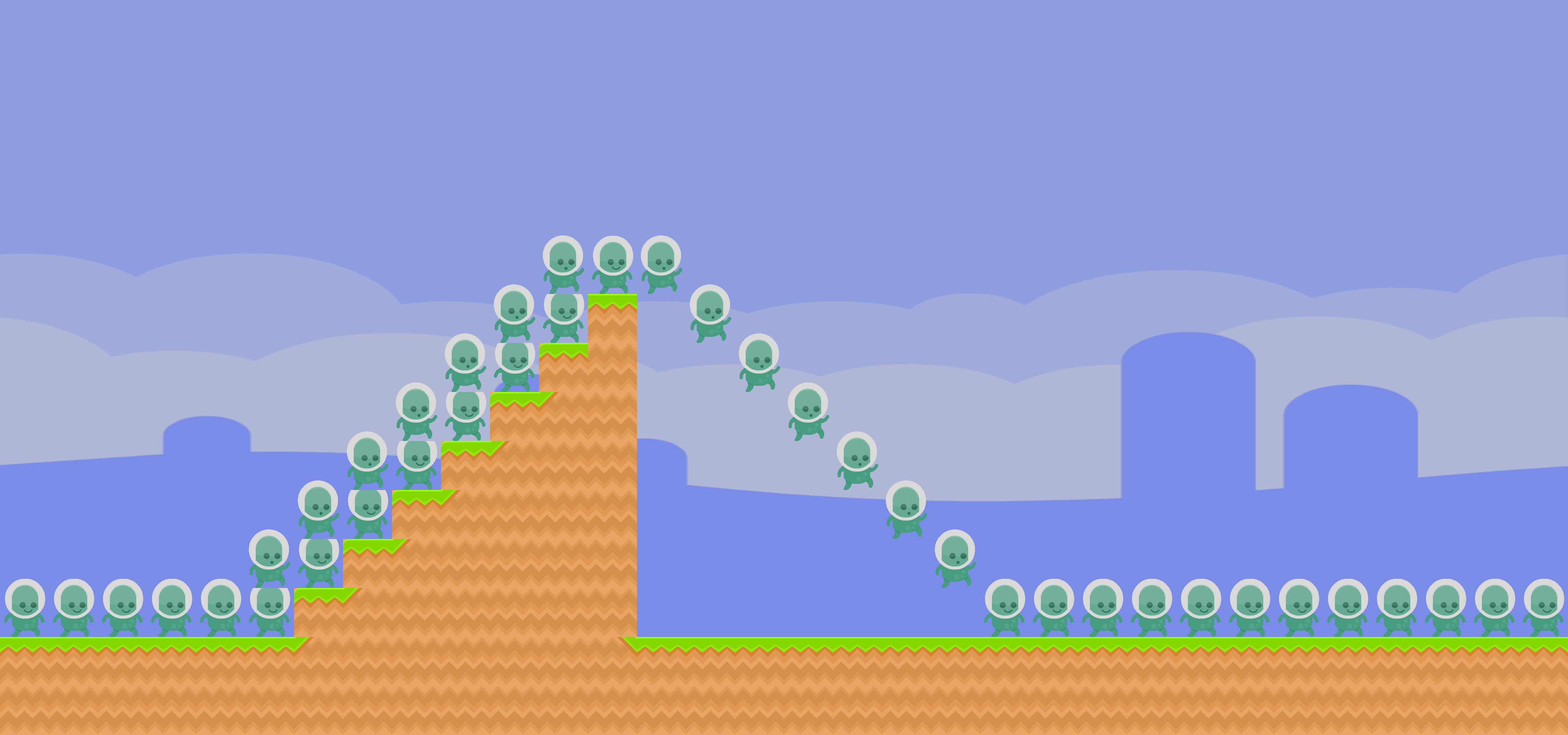}

\includegraphics[width=0.12\pdfpagewidth]{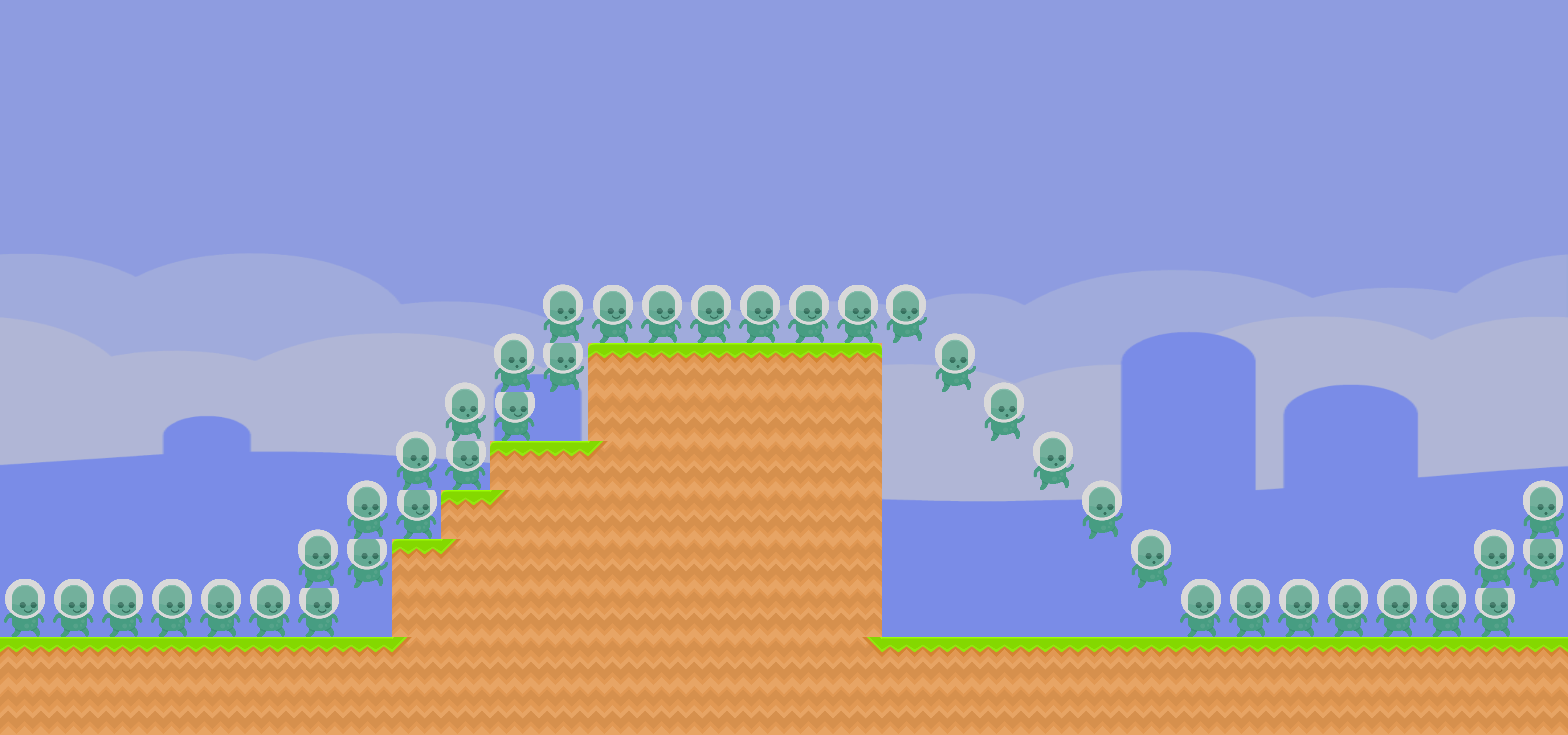}

\includegraphics[width=0.12\pdfpagewidth]{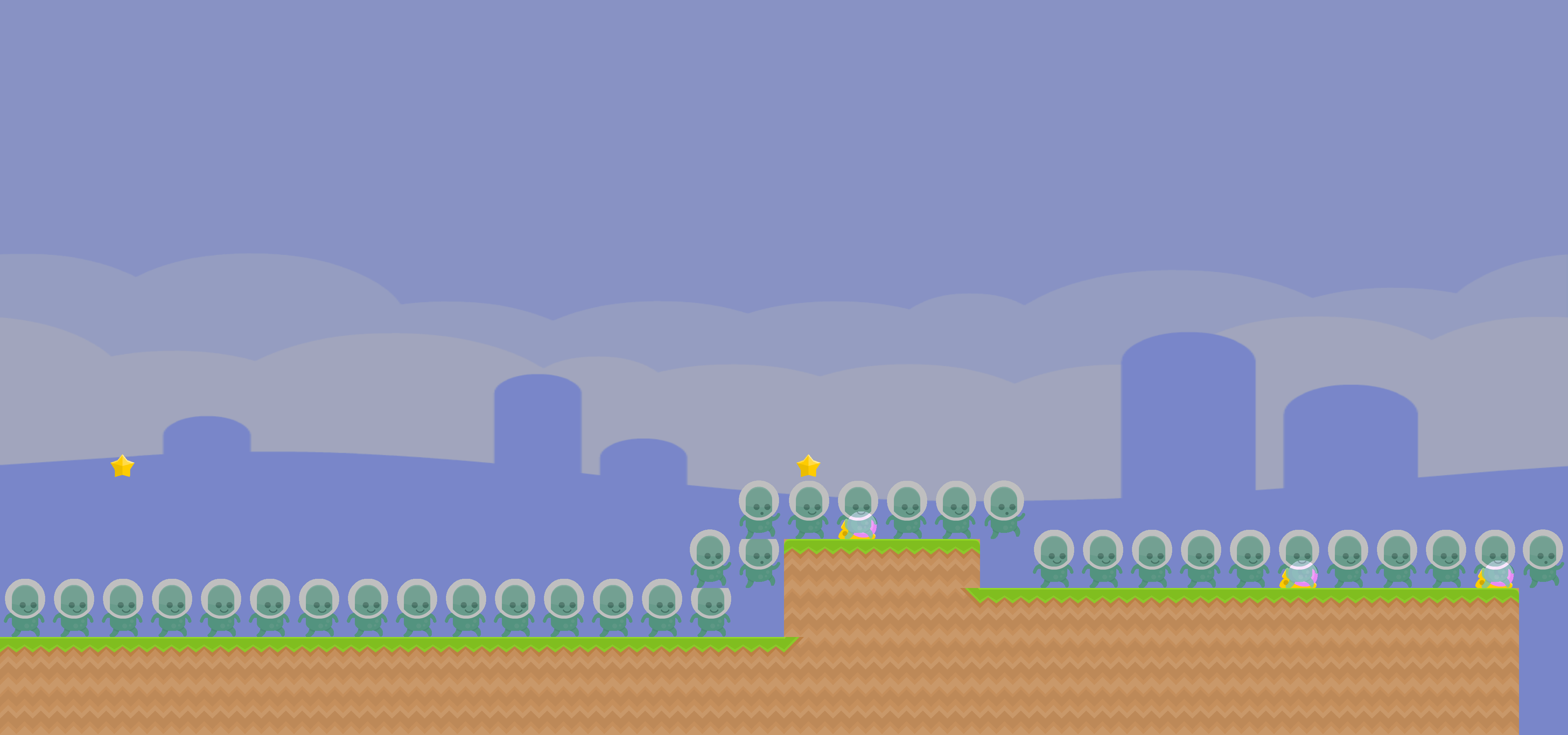}

\includegraphics[width=0.12\pdfpagewidth]{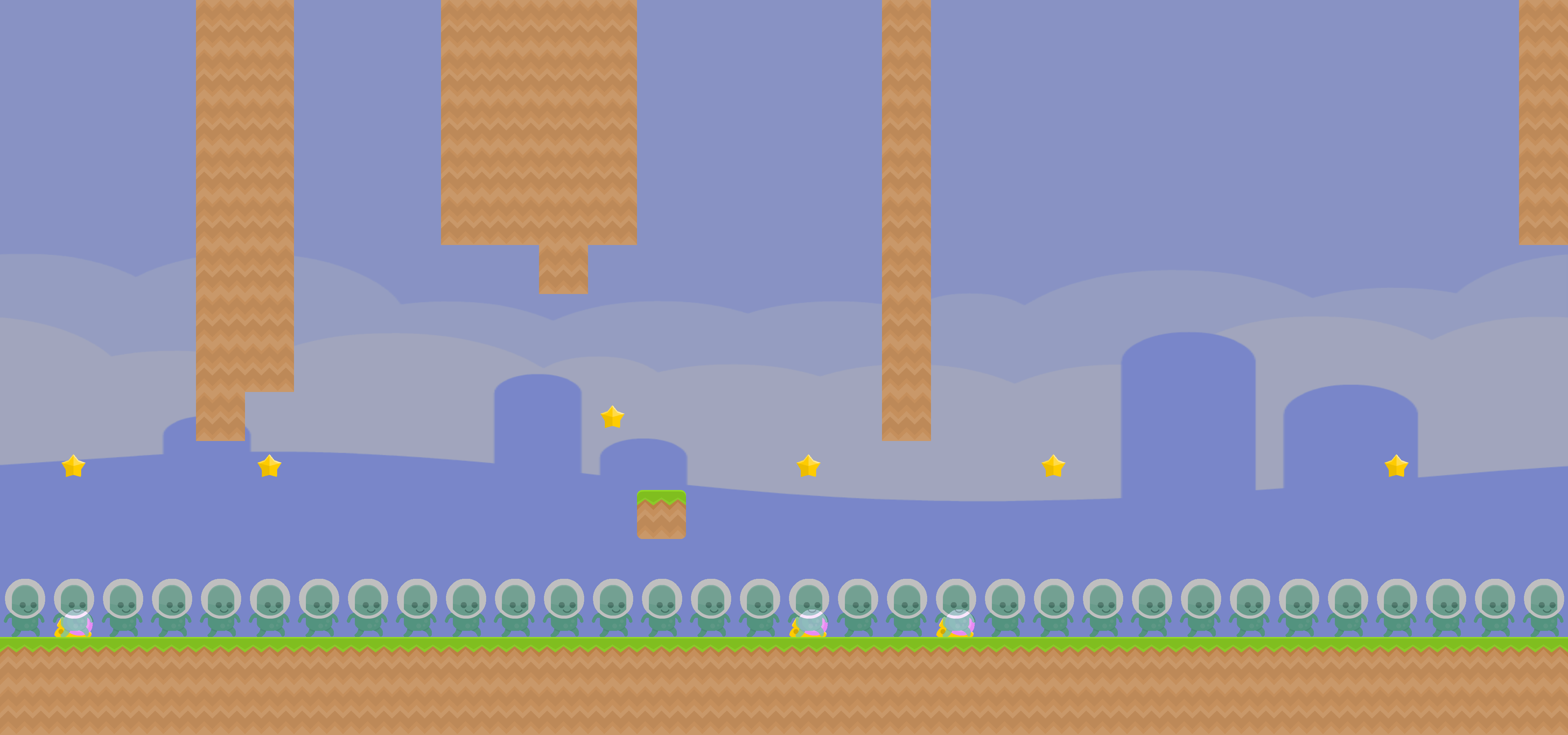}
\caption{GRU  SMB $\downarrow$ CV}
\label{fig:gru_interp_Mario_Cast}
\end{subfigure}%
\begin{subfigure}{.18\textwidth}
\centering
\includegraphics[width=0.12\pdfpagewidth]{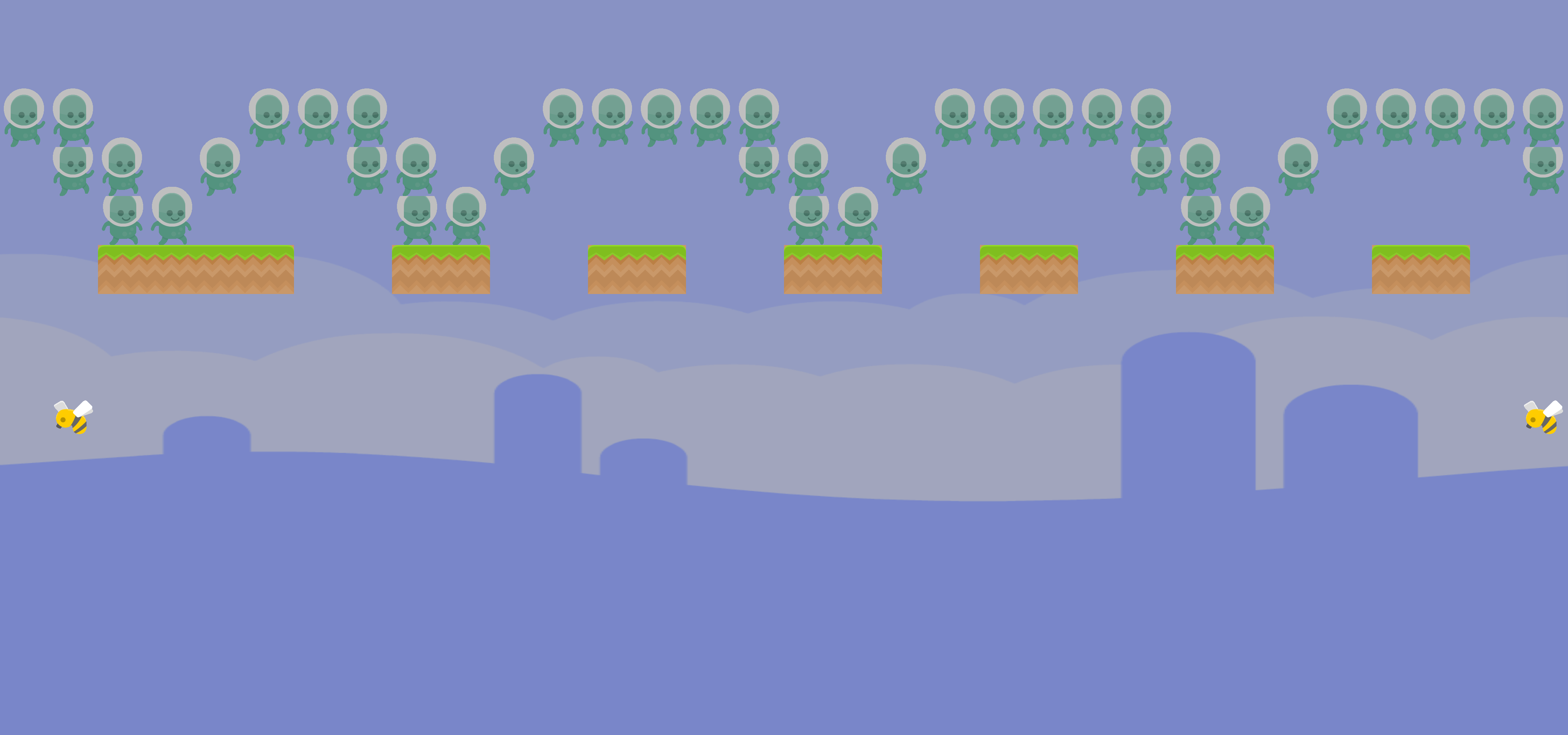}

\includegraphics[width=0.12\pdfpagewidth]{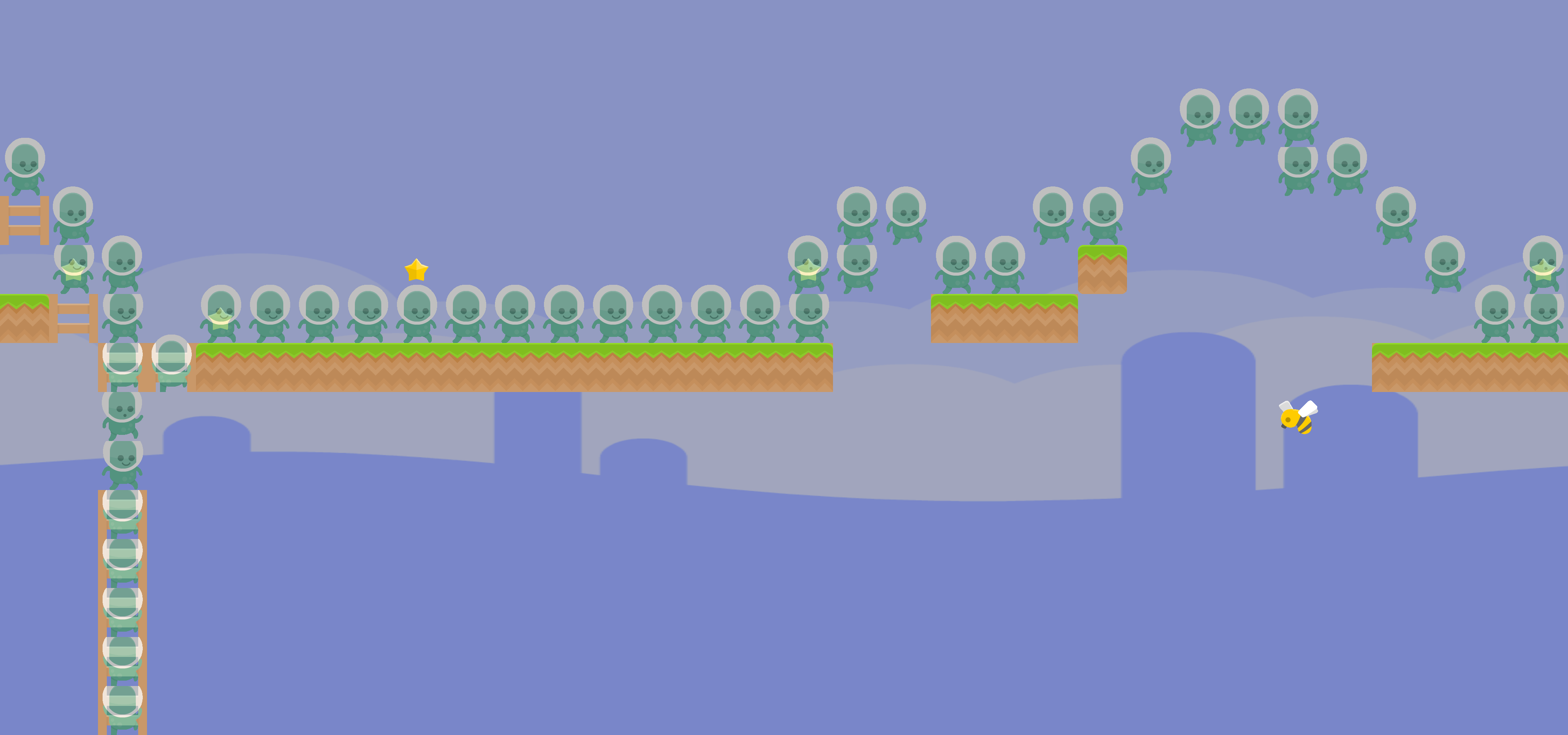}

\includegraphics[width=0.12\pdfpagewidth]{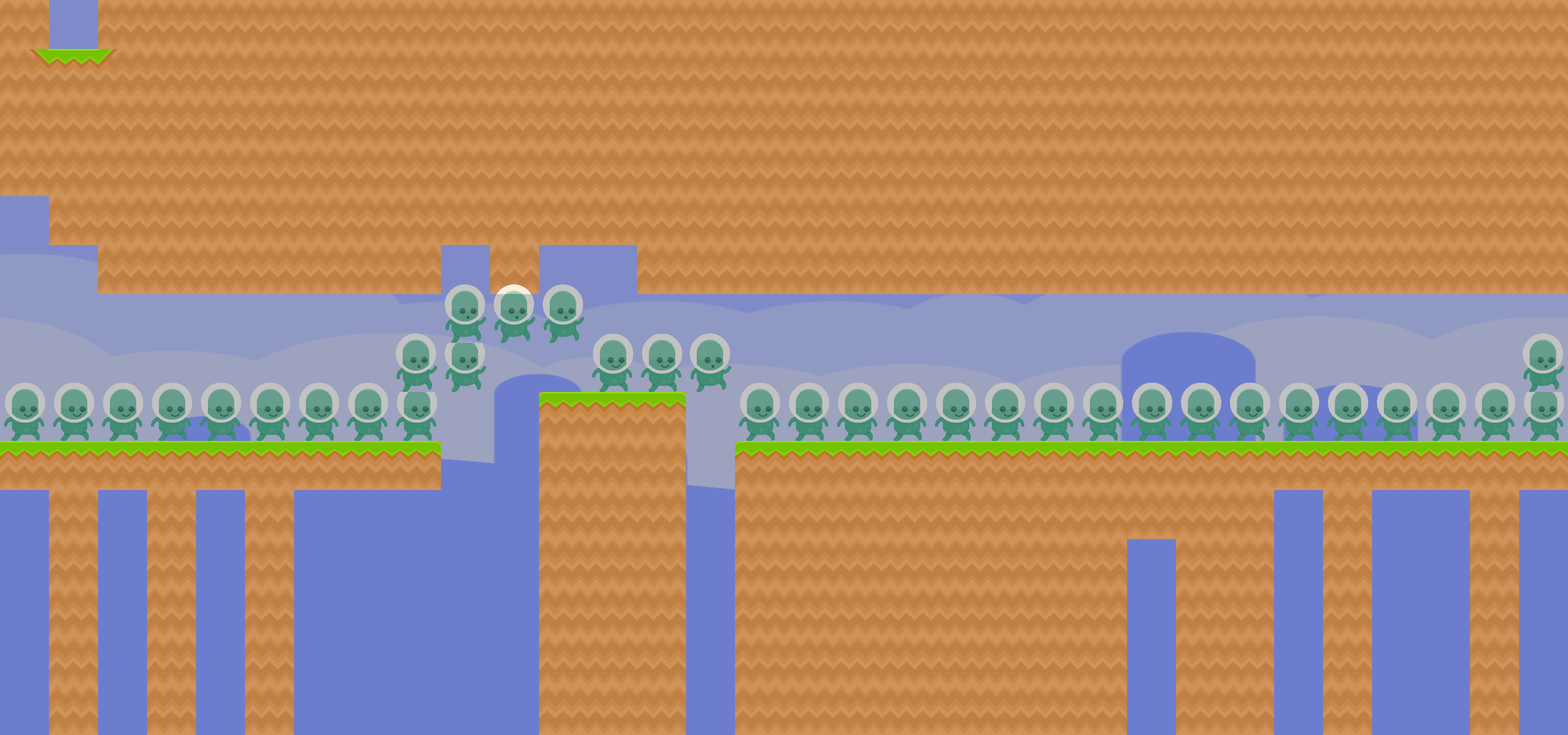}

\includegraphics[width=0.12\pdfpagewidth]{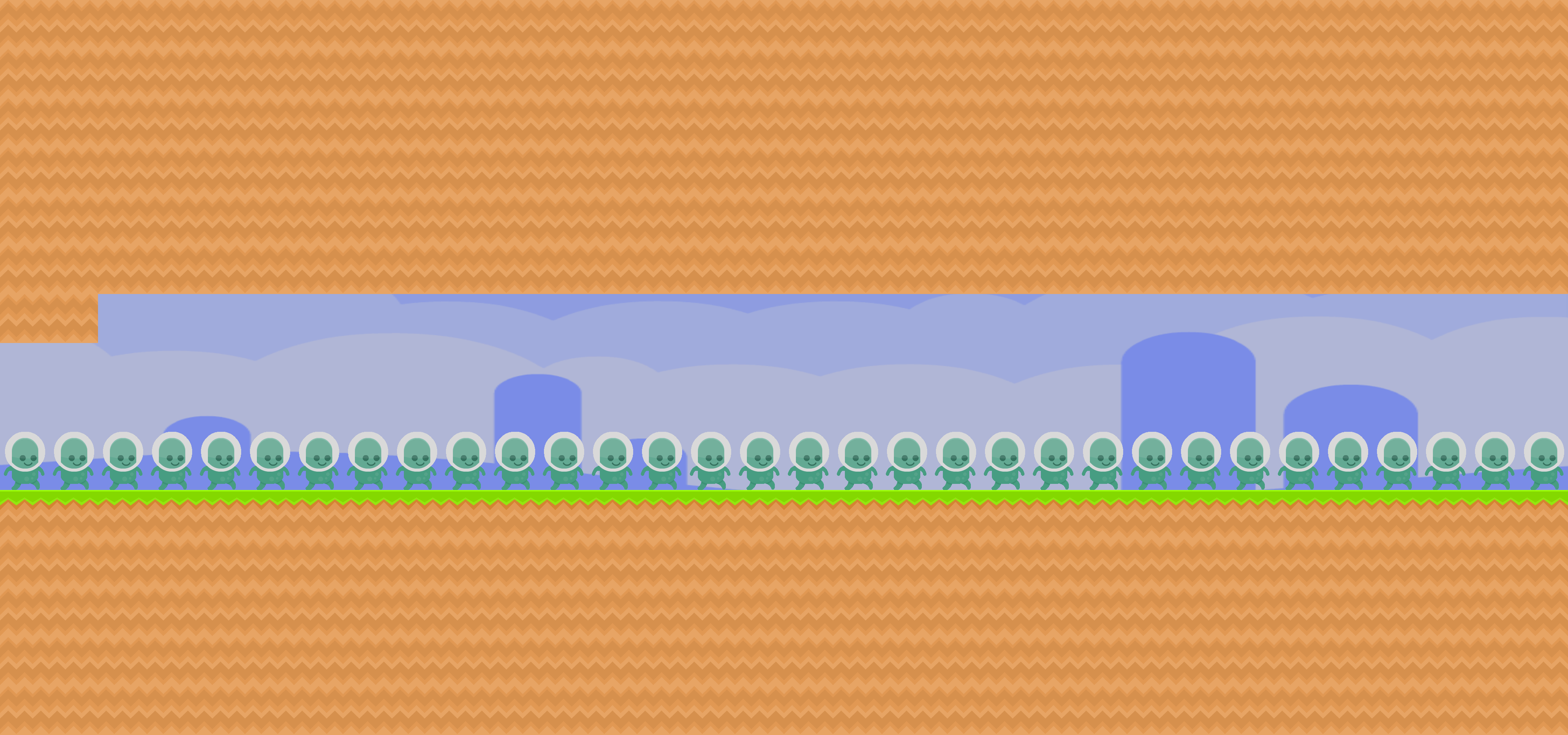}

\includegraphics[width=0.12\pdfpagewidth]{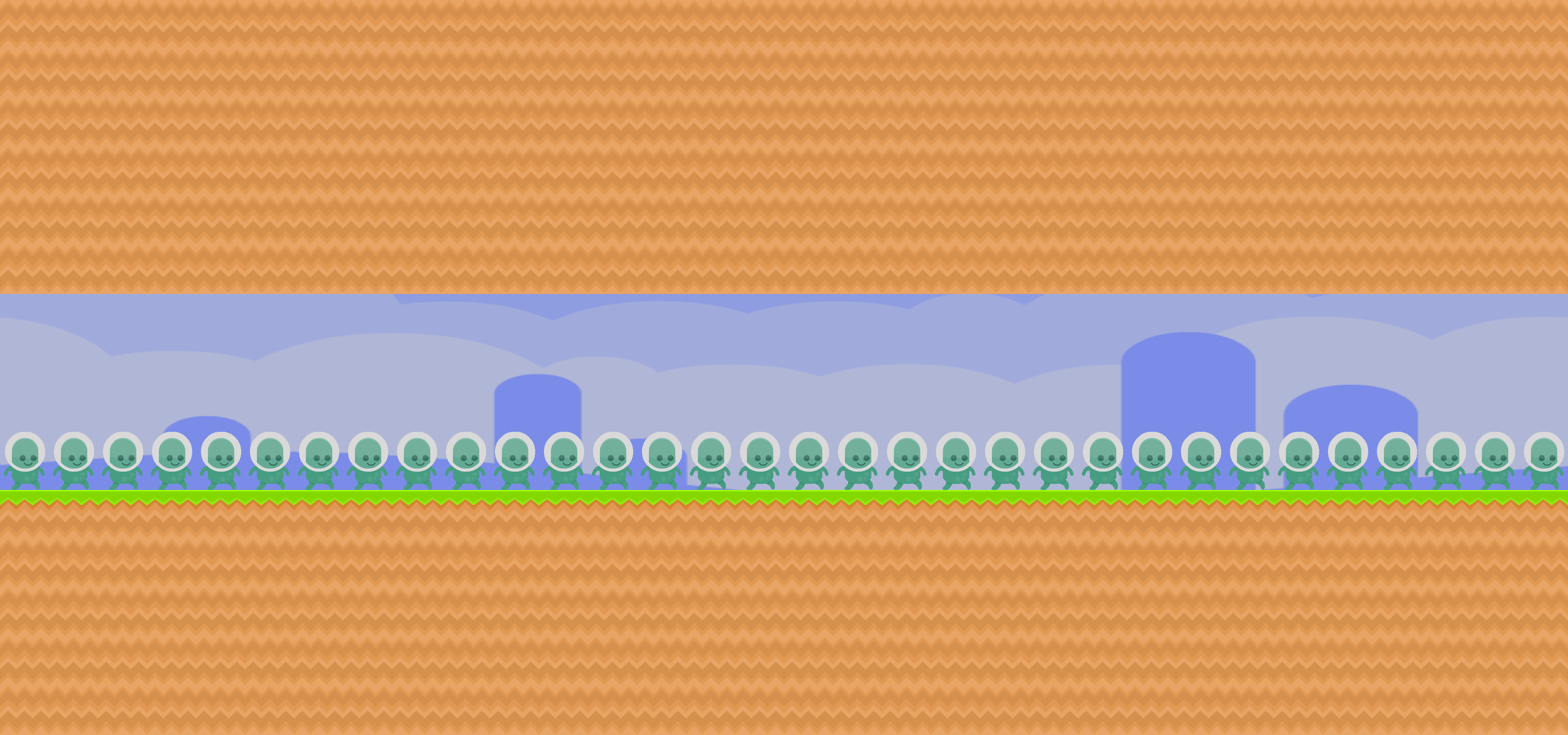}
\caption{GRU  CV $\downarrow$ MM}
\label{fig:gru_interp_Cast_MM}
\end{subfigure}
\begin{subfigure}{.18\textwidth}
\centering
\includegraphics[width=0.12\pdfpagewidth]{GRU/Castlevania_1012-txt-100_small}

\includegraphics[width=0.12\pdfpagewidth]{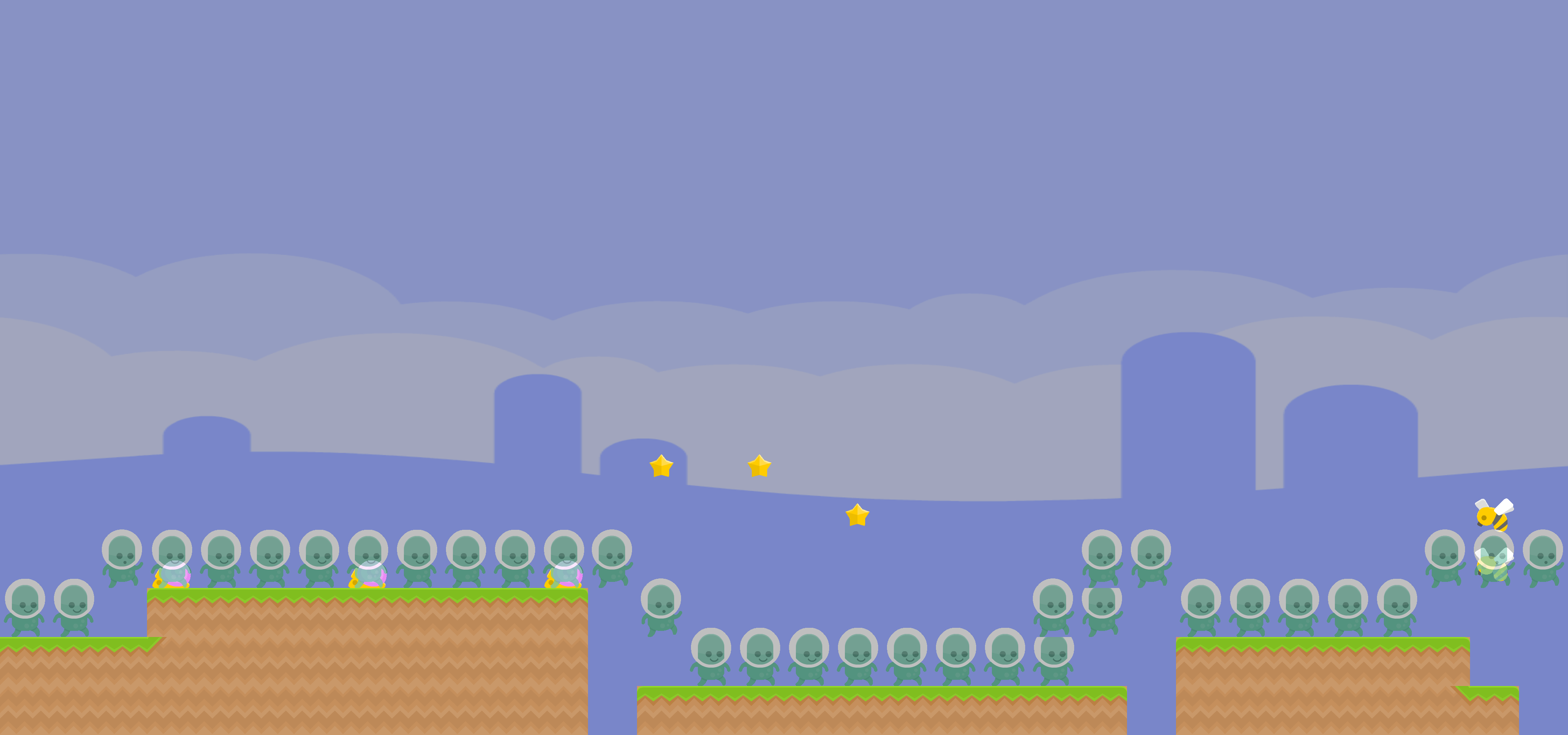}

\includegraphics[width=0.12\pdfpagewidth]{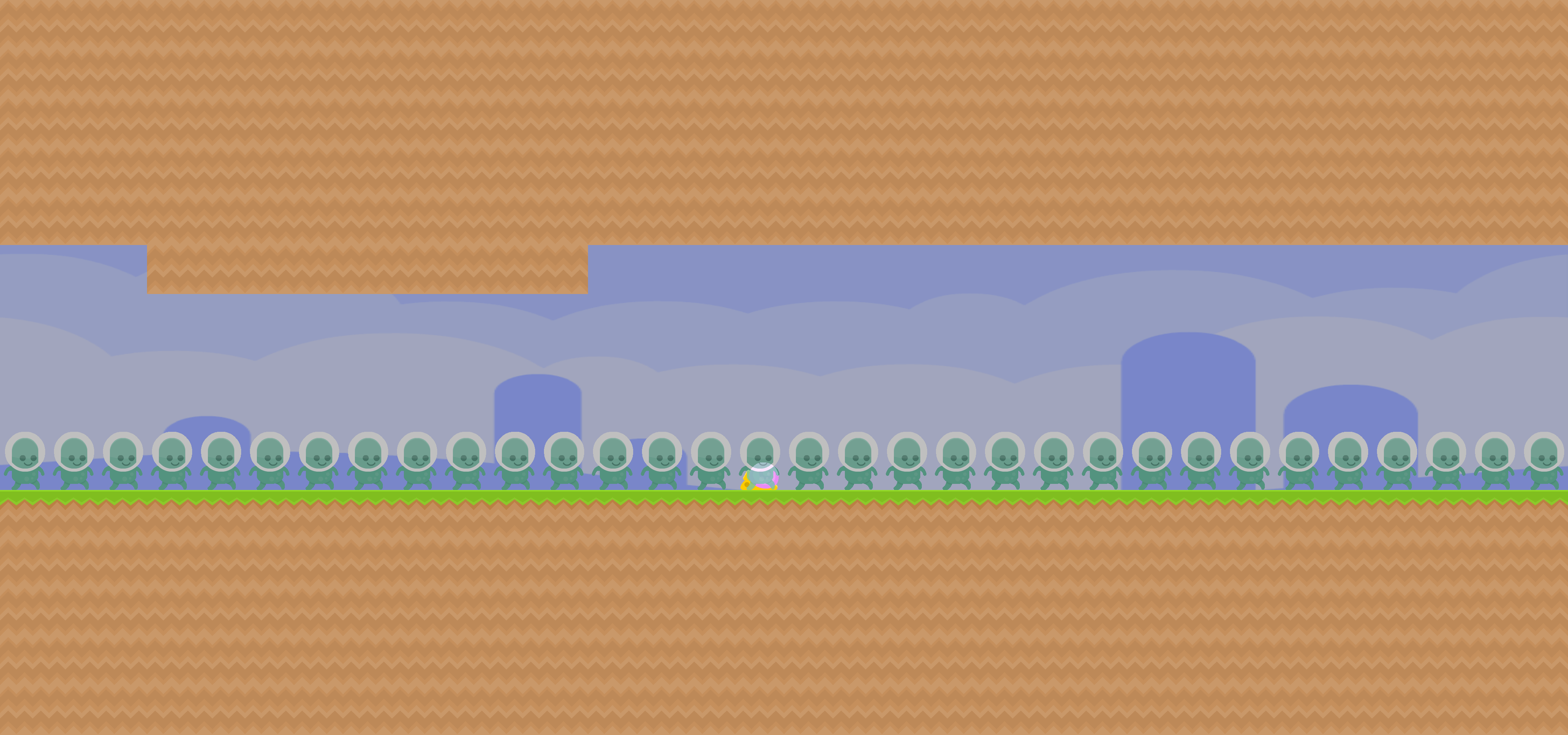}

\includegraphics[width=0.12\pdfpagewidth]{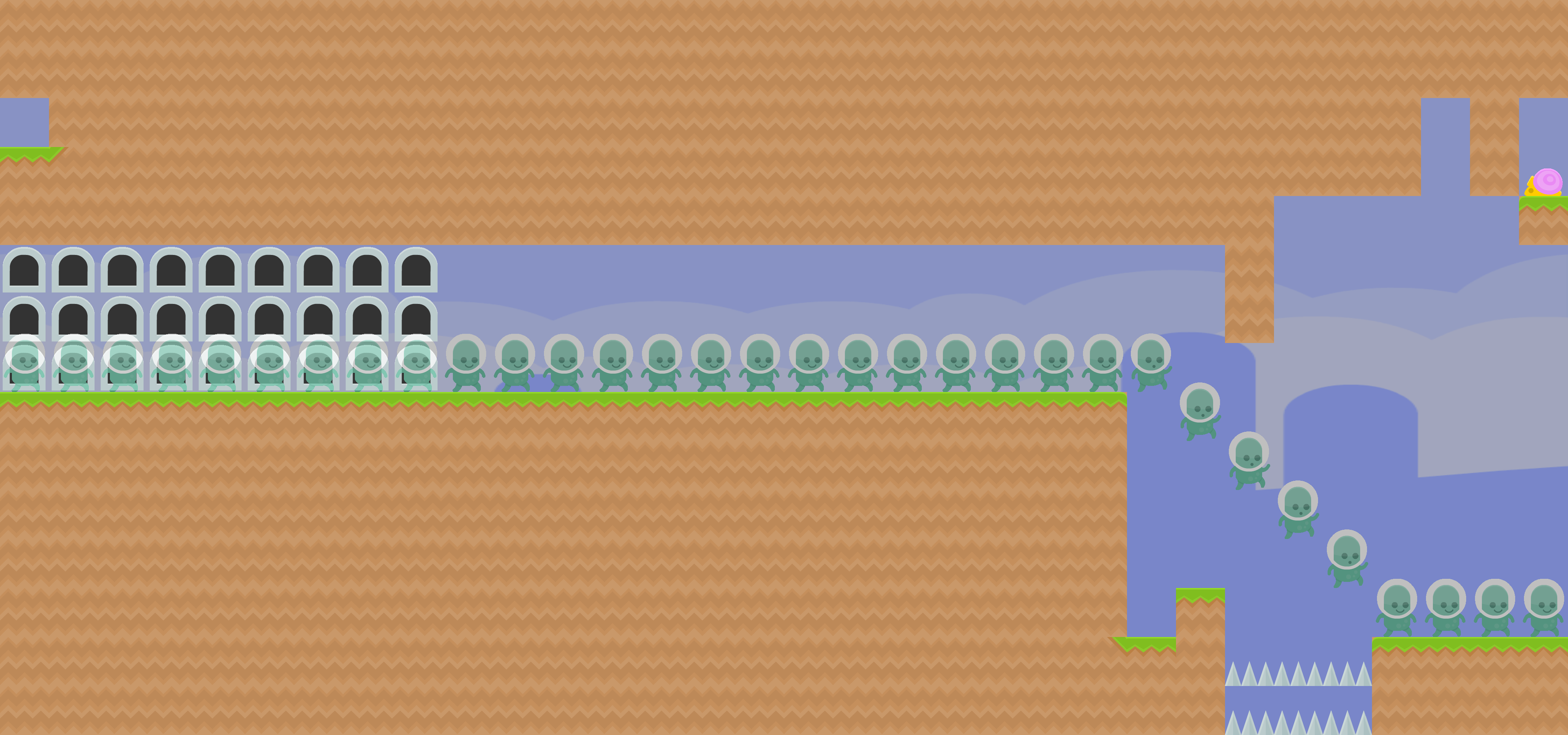}

\includegraphics[width=0.12\pdfpagewidth]{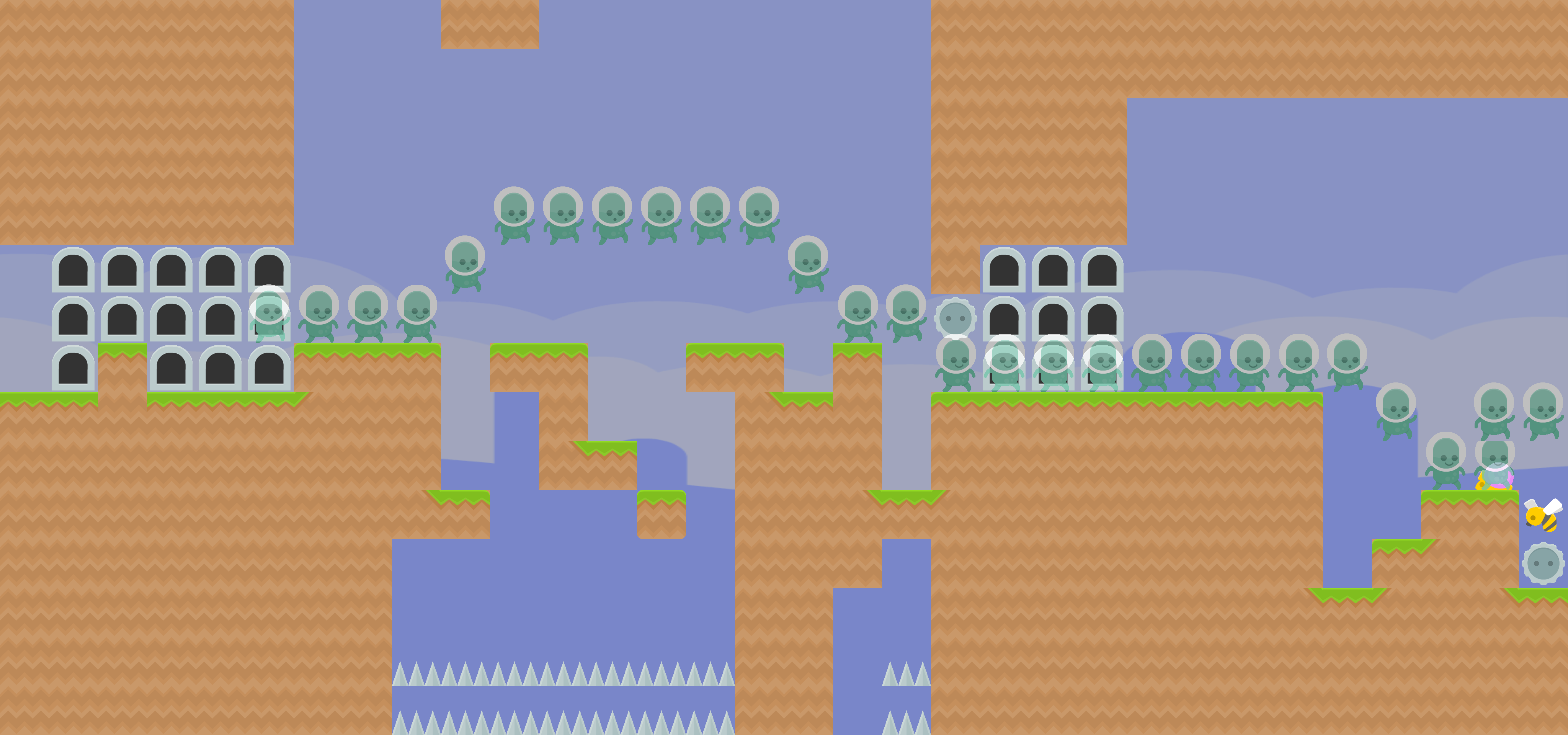}
\caption{GRU  CV $\downarrow$ Met}
\label{fig:gru_interp_Cast_Met}
\end{subfigure}%
\begin{subfigure}{.18\textwidth}
\centering
\includegraphics[width=0.12\pdfpagewidth]{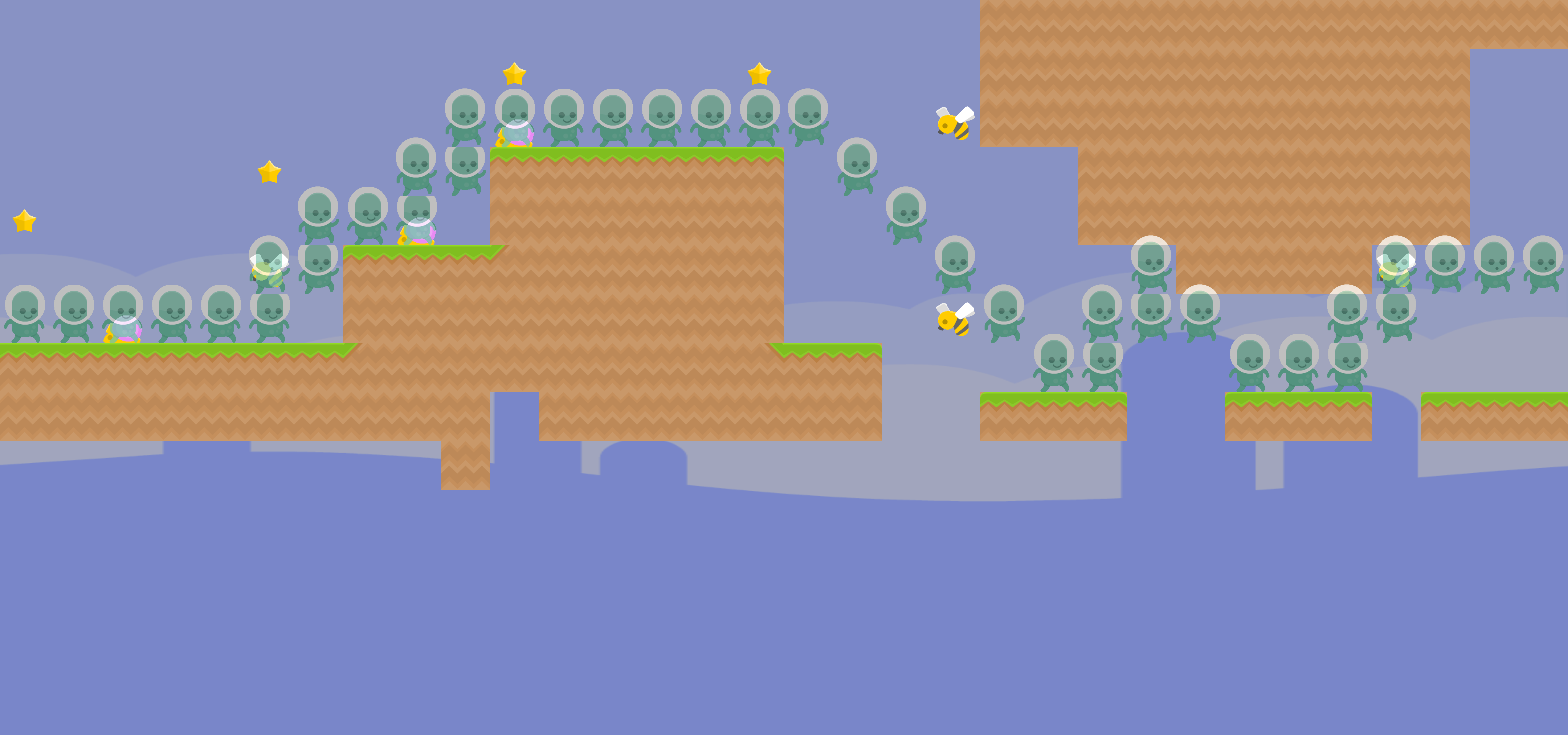}

\includegraphics[width=0.12\pdfpagewidth]{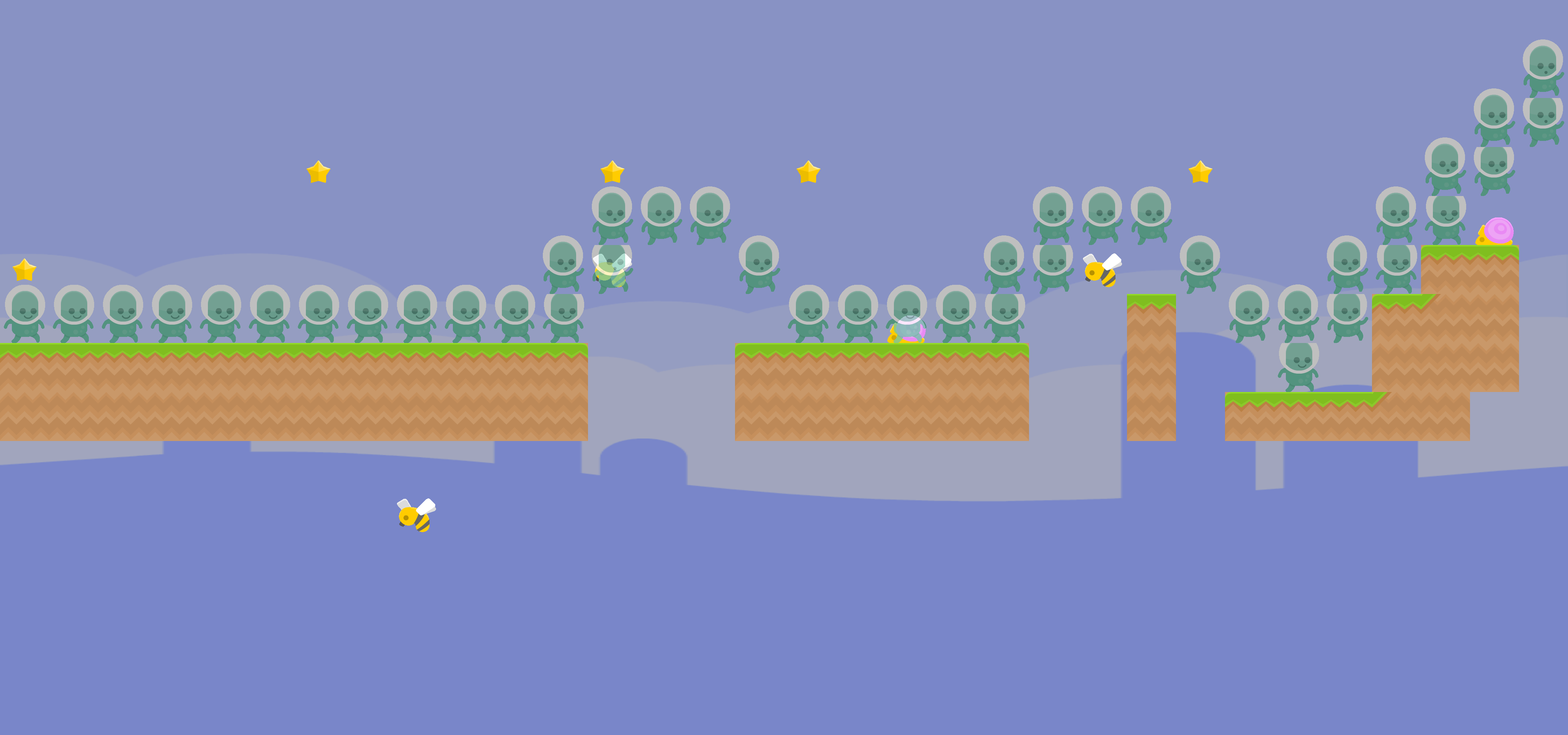}

\includegraphics[width=0.12\pdfpagewidth]{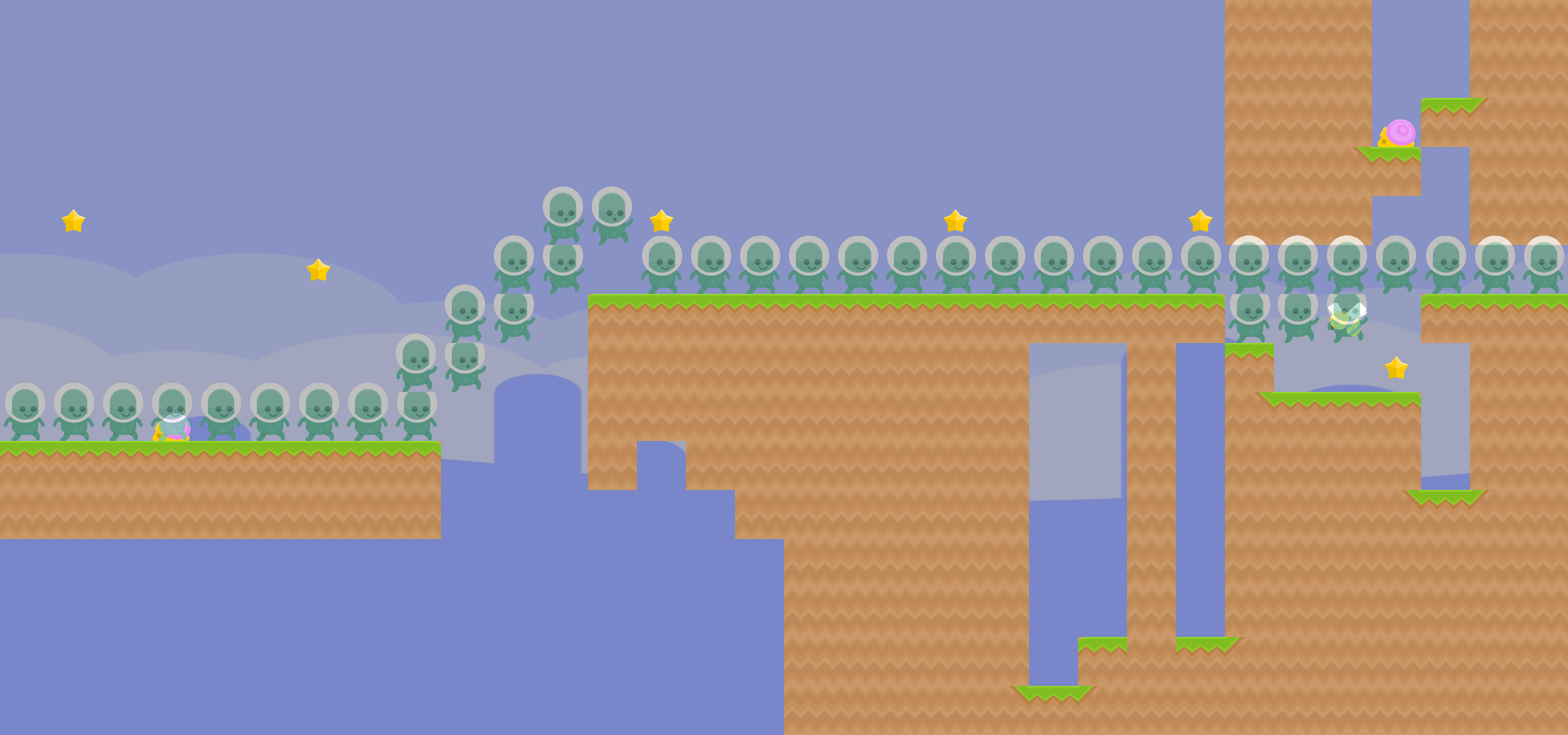}

\includegraphics[width=0.12\pdfpagewidth]{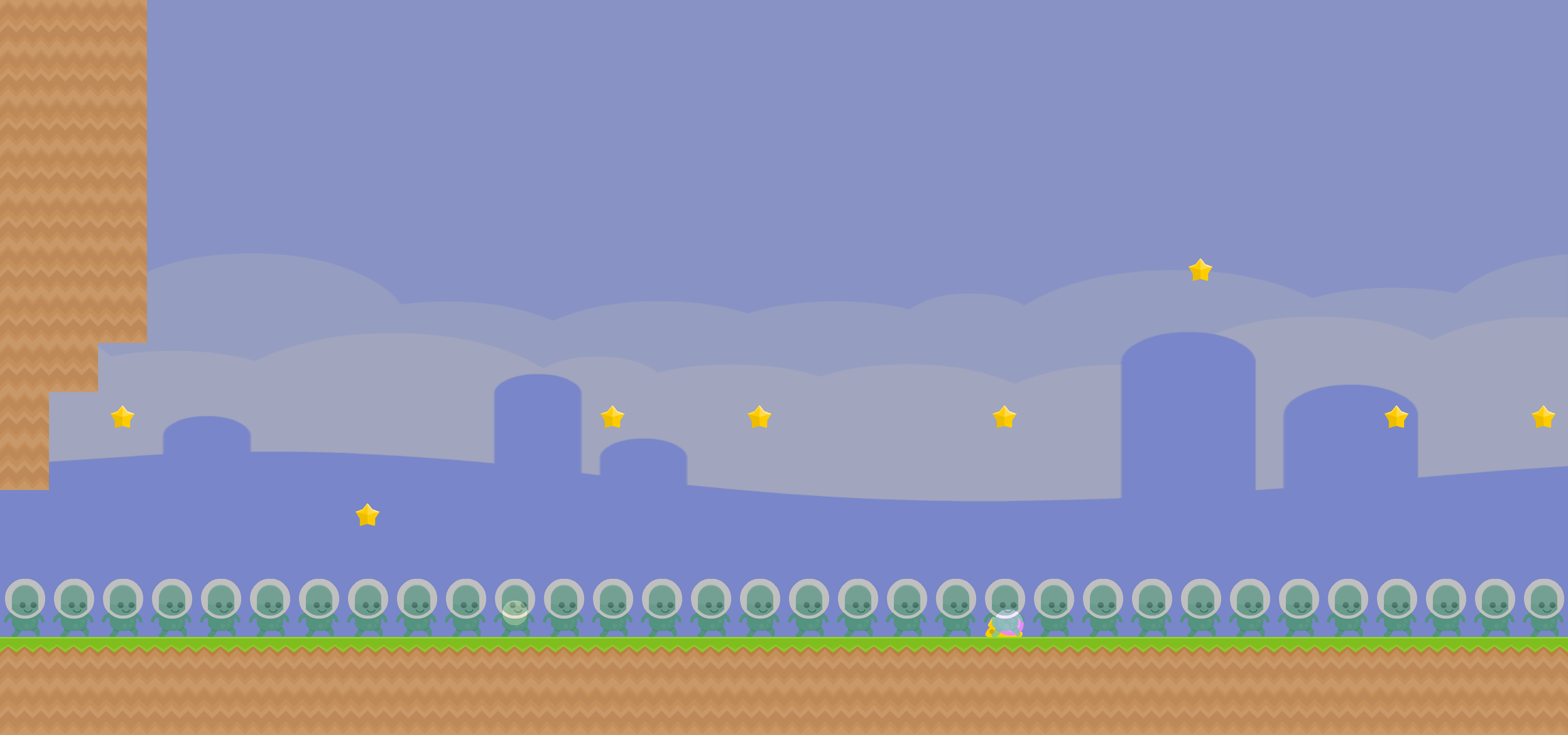}

\includegraphics[width=0.12\pdfpagewidth]{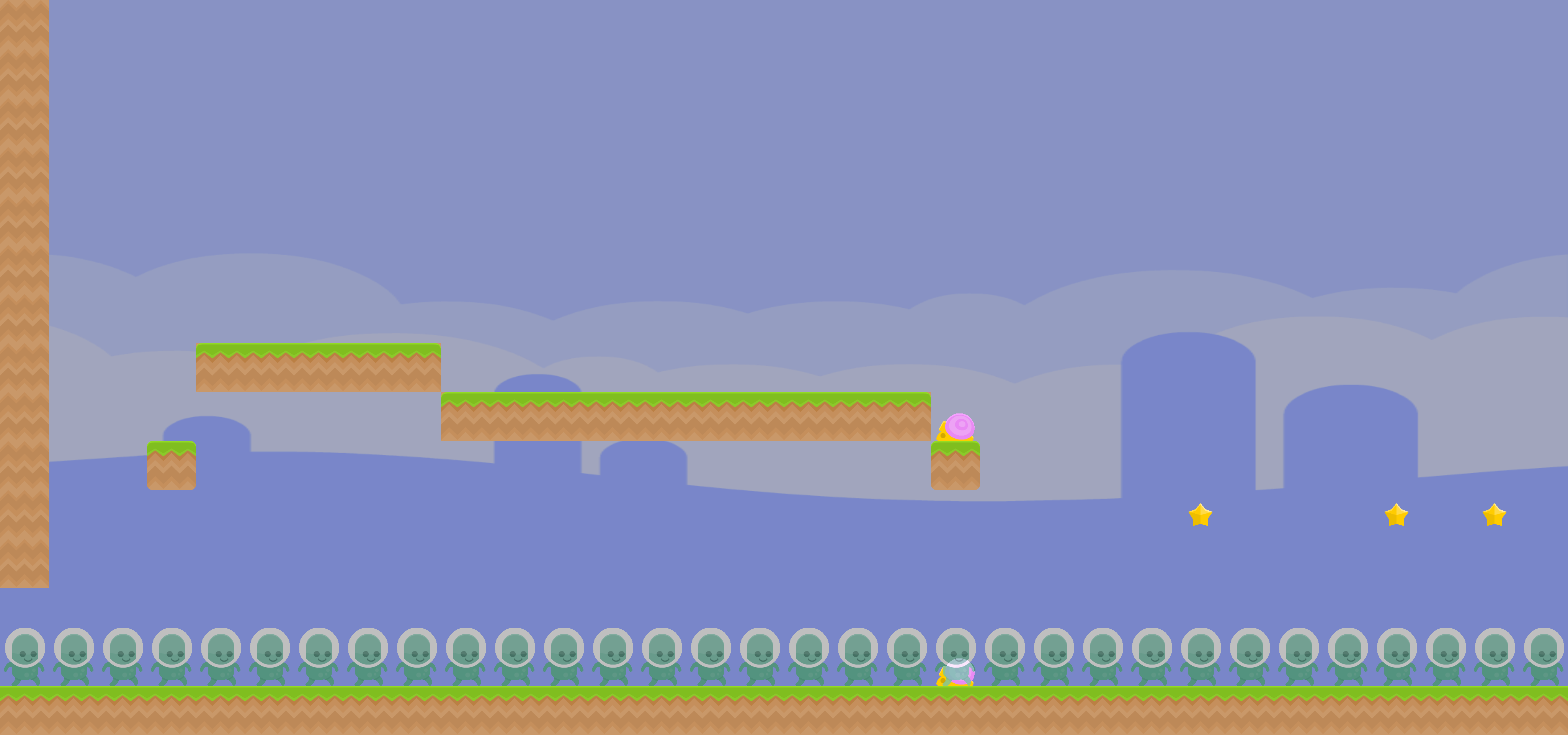}
\caption{GRU  CV $\downarrow$ NG}
\label{fig:gru_interp_Cast_NG}
\end{subfigure}%
\begin{subfigure}{.18\textwidth}
\centering
\includegraphics[width=0.12\pdfpagewidth]{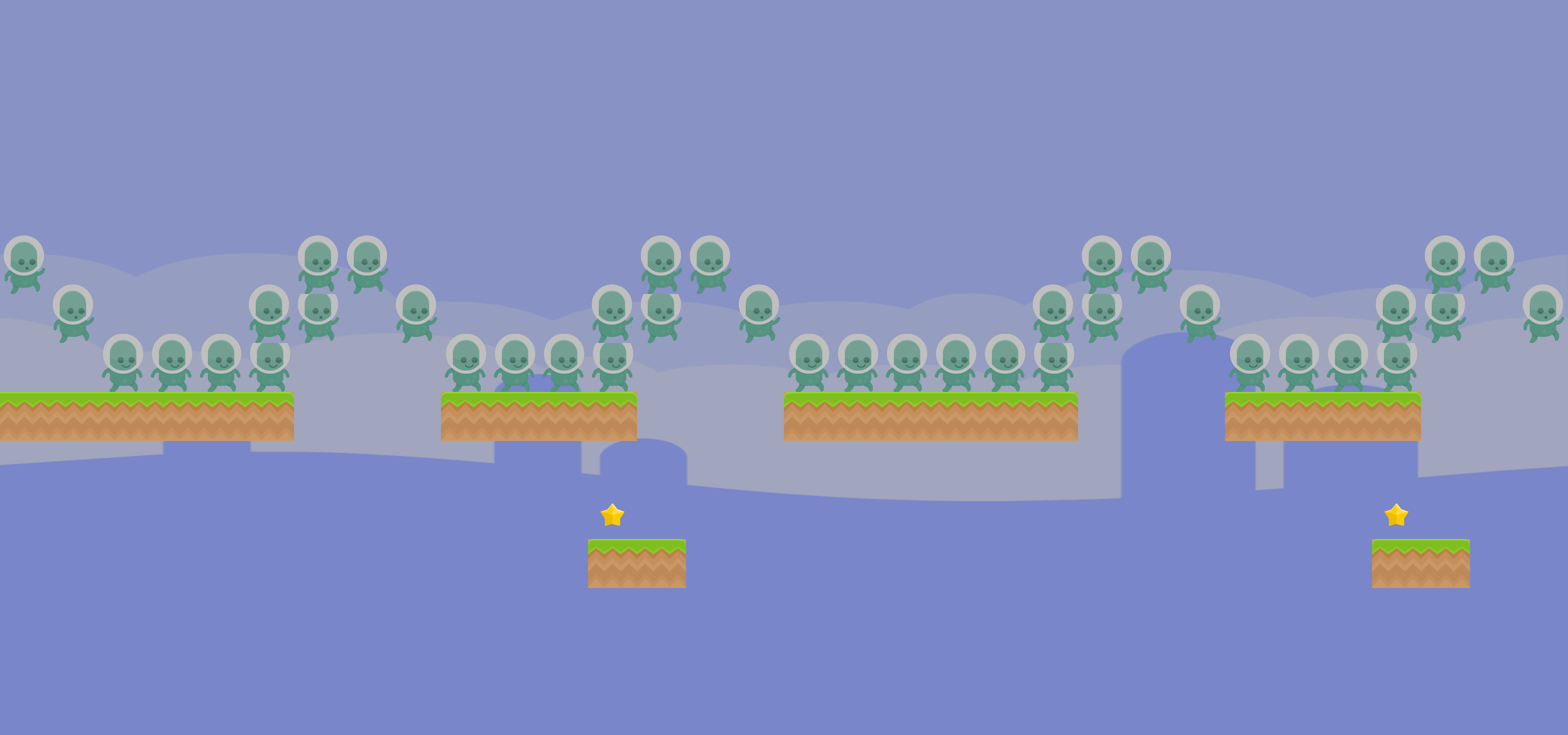}

\includegraphics[width=0.12\pdfpagewidth]{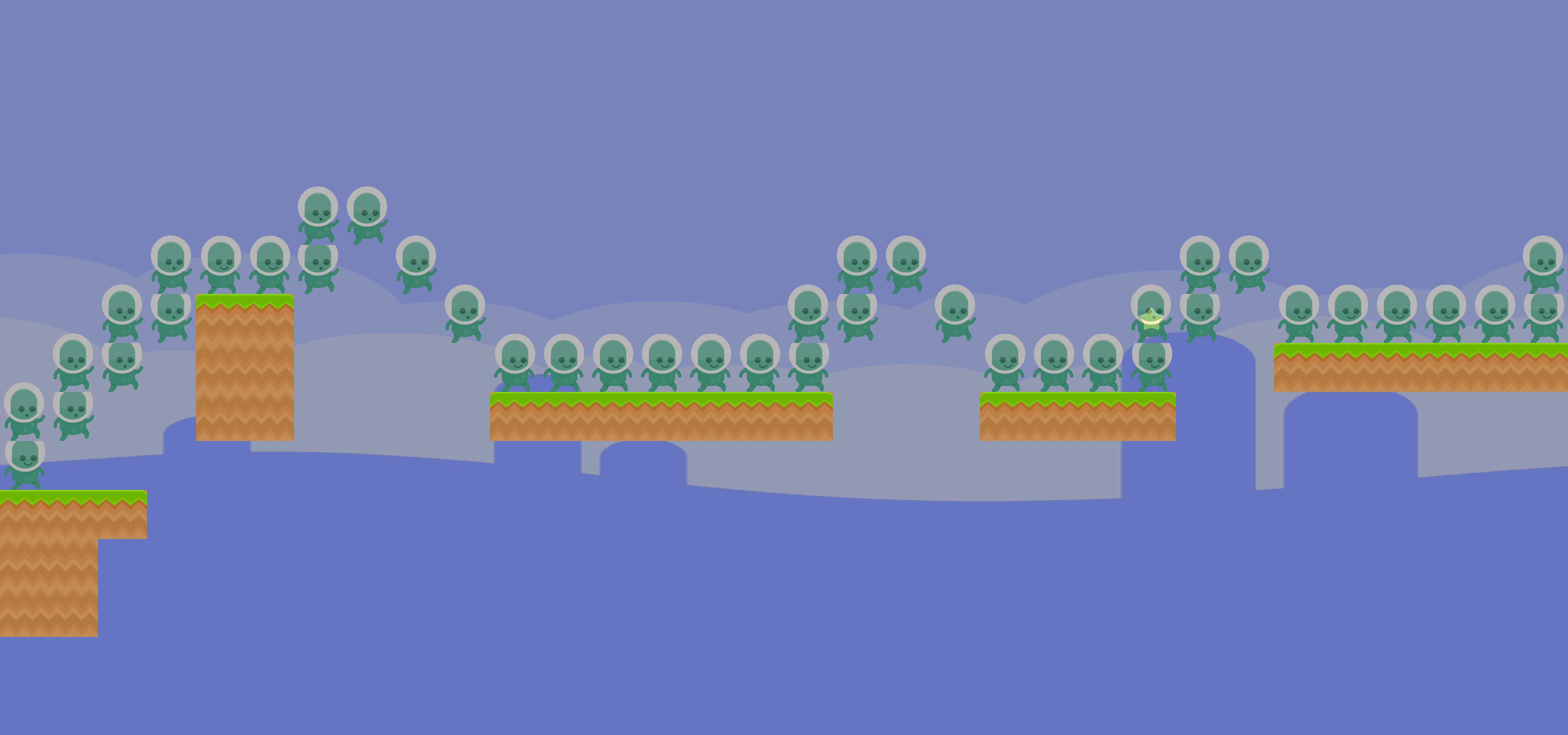}

\includegraphics[width=0.12\pdfpagewidth]{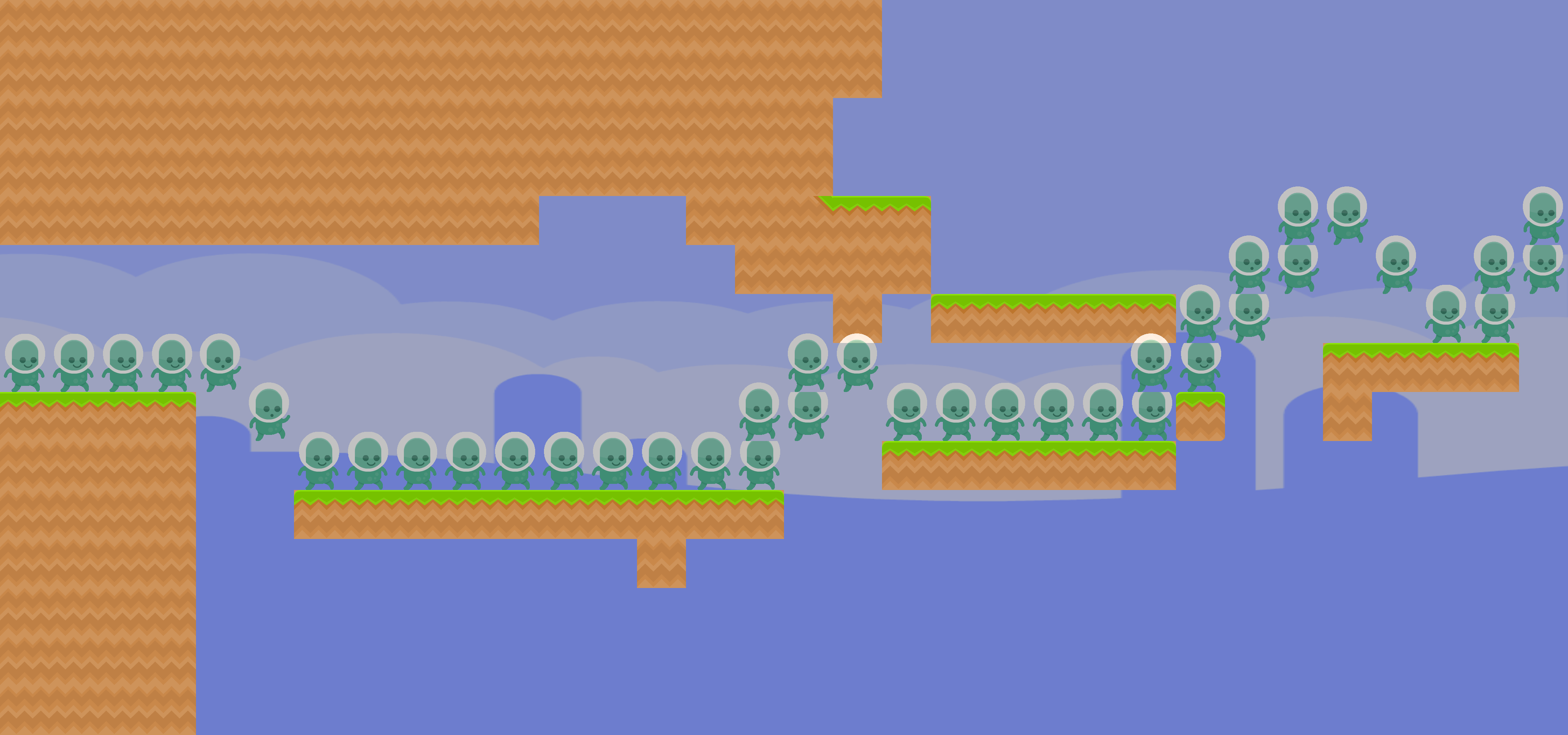}

\includegraphics[width=0.12\pdfpagewidth]{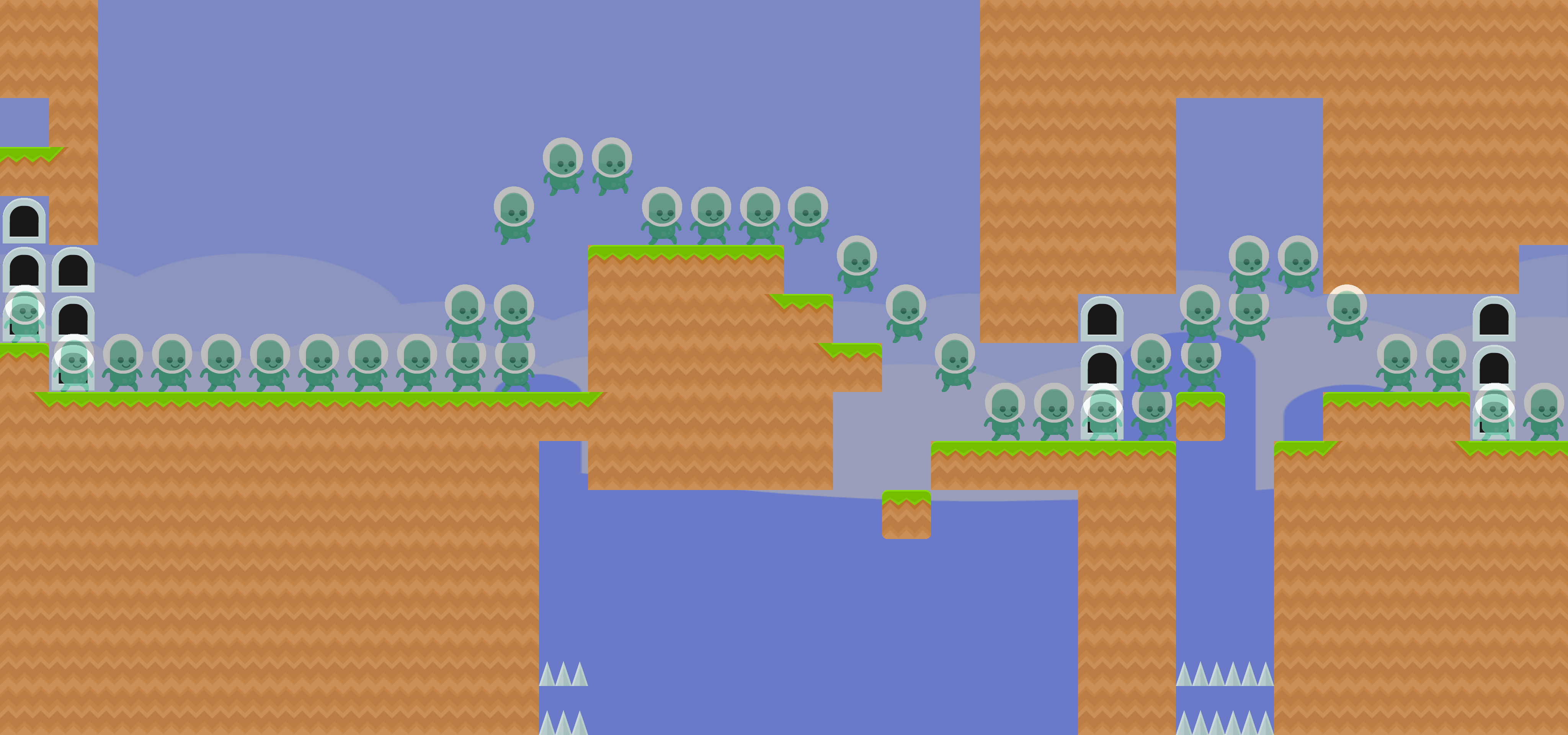}

\includegraphics[width=0.12\pdfpagewidth]{GRU/Metroid_1544-txt-100_small}
\caption{GRU  MM $\downarrow$ Met}
\label{fig:gru_interp_MM_met}
\end{subfigure}%
\begin{subfigure}{.18\textwidth}
\centering
\includegraphics[width=0.12\pdfpagewidth]{GRU/MegaMan_3315-txt-100_small}

\includegraphics[width=0.12\pdfpagewidth]{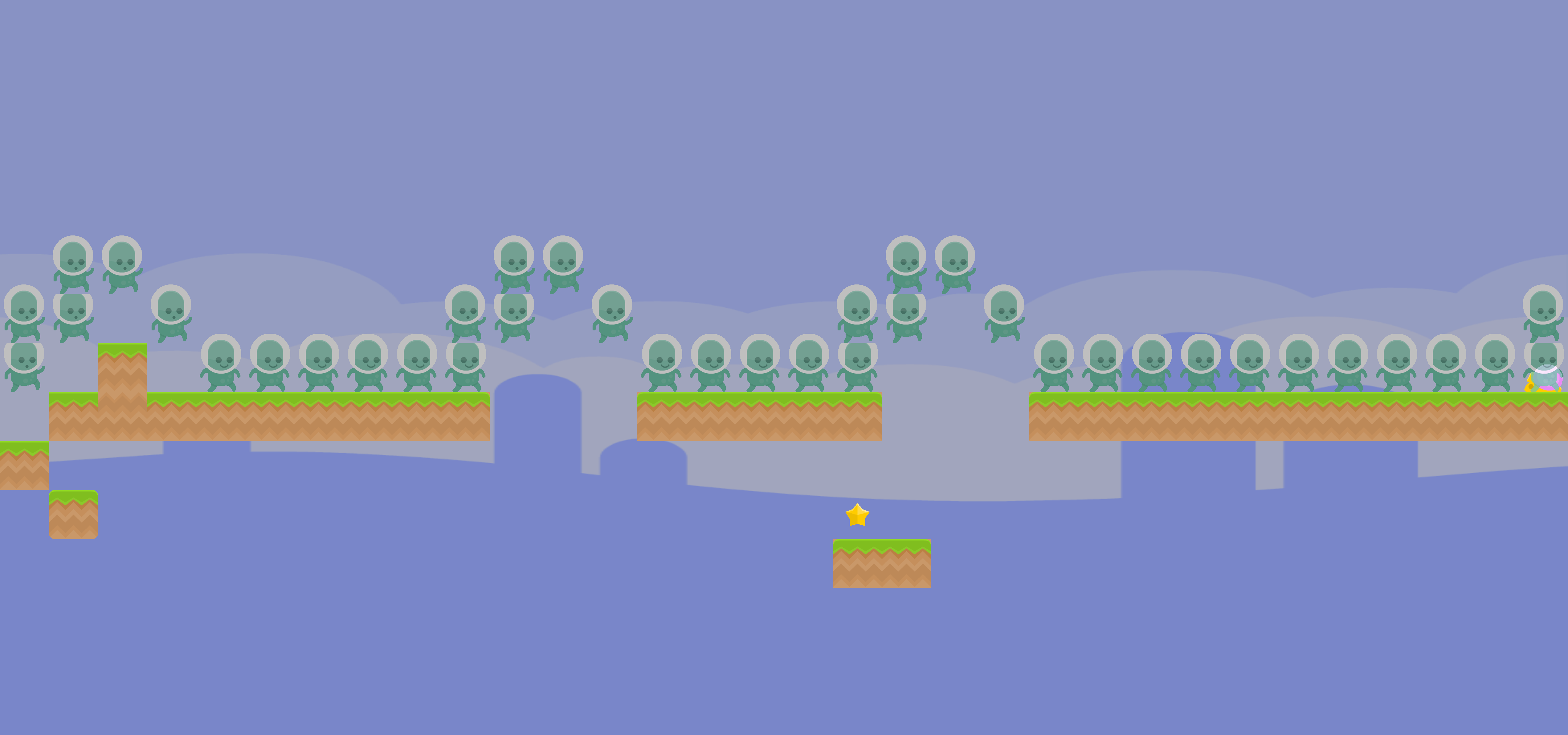}

\includegraphics[width=0.12\pdfpagewidth]{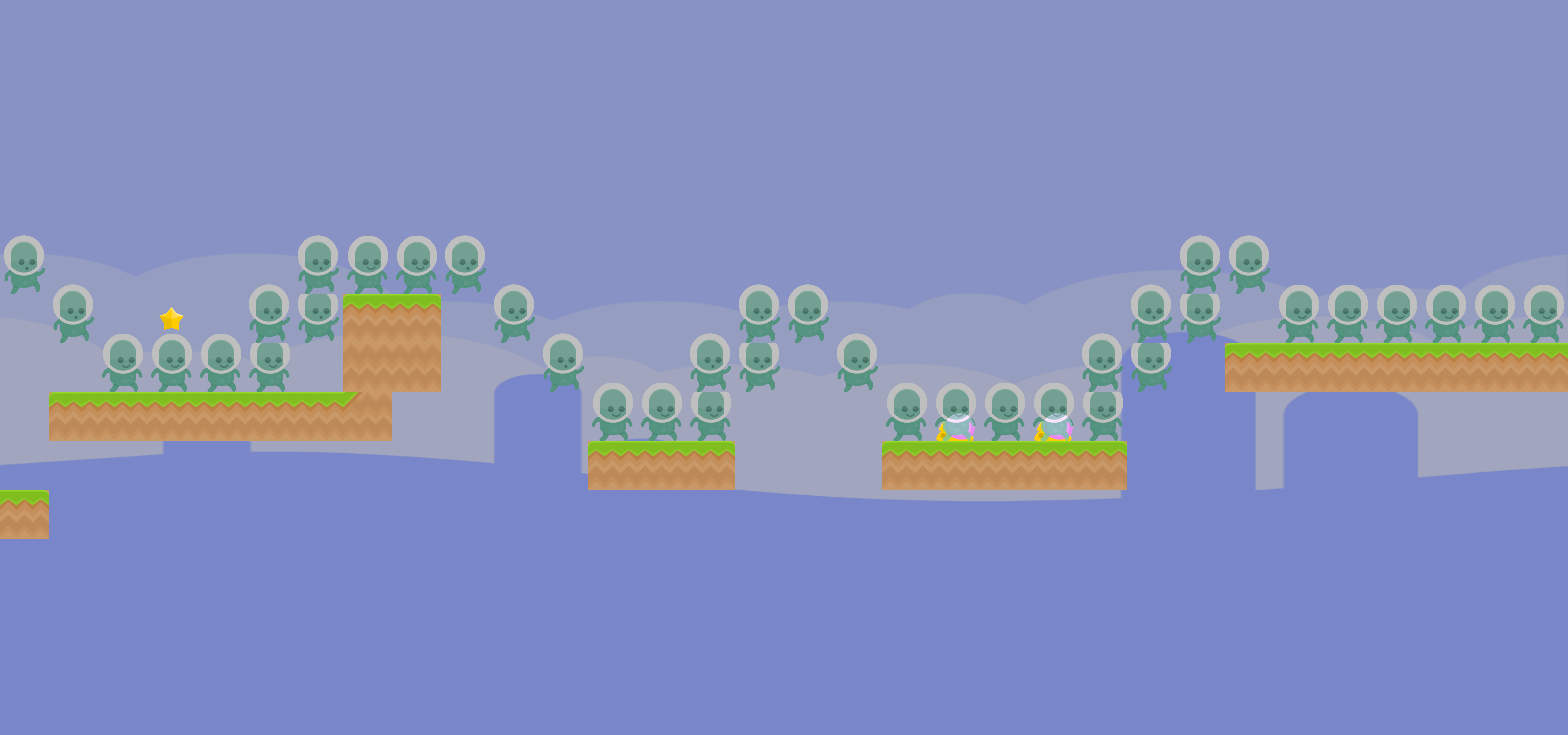}

\includegraphics[width=0.12\pdfpagewidth]{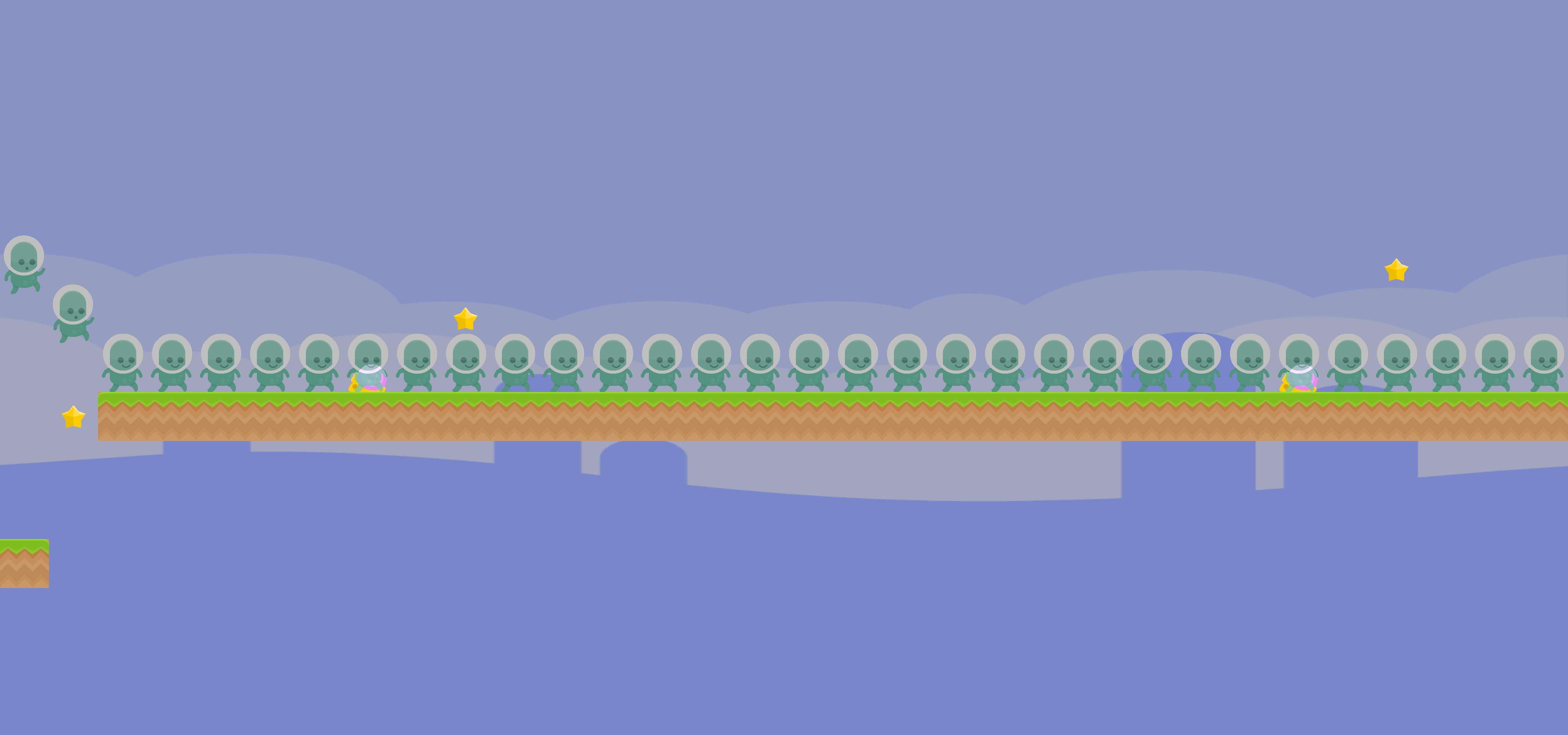}

\includegraphics[width=0.12\pdfpagewidth]{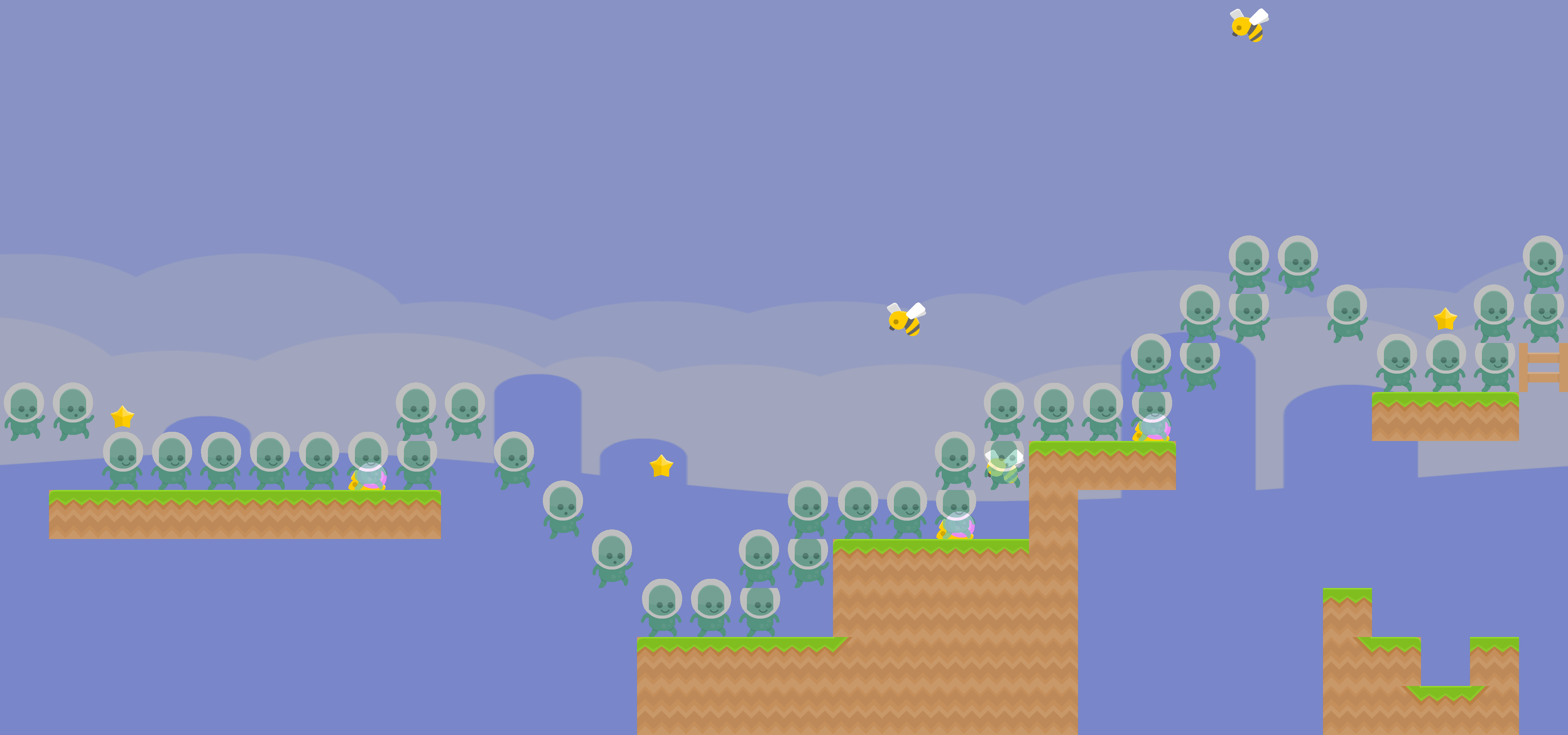}
\caption{GRU MM $\downarrow$ NG}
\label{fig:gru_interp_MM_NG}
\end{subfigure}%
\begin{subfigure}{.18\textwidth}
\centering
\includegraphics[width=0.12\pdfpagewidth]{GRU/Metroid_1544-txt-100_small}

\includegraphics[width=0.12\pdfpagewidth]{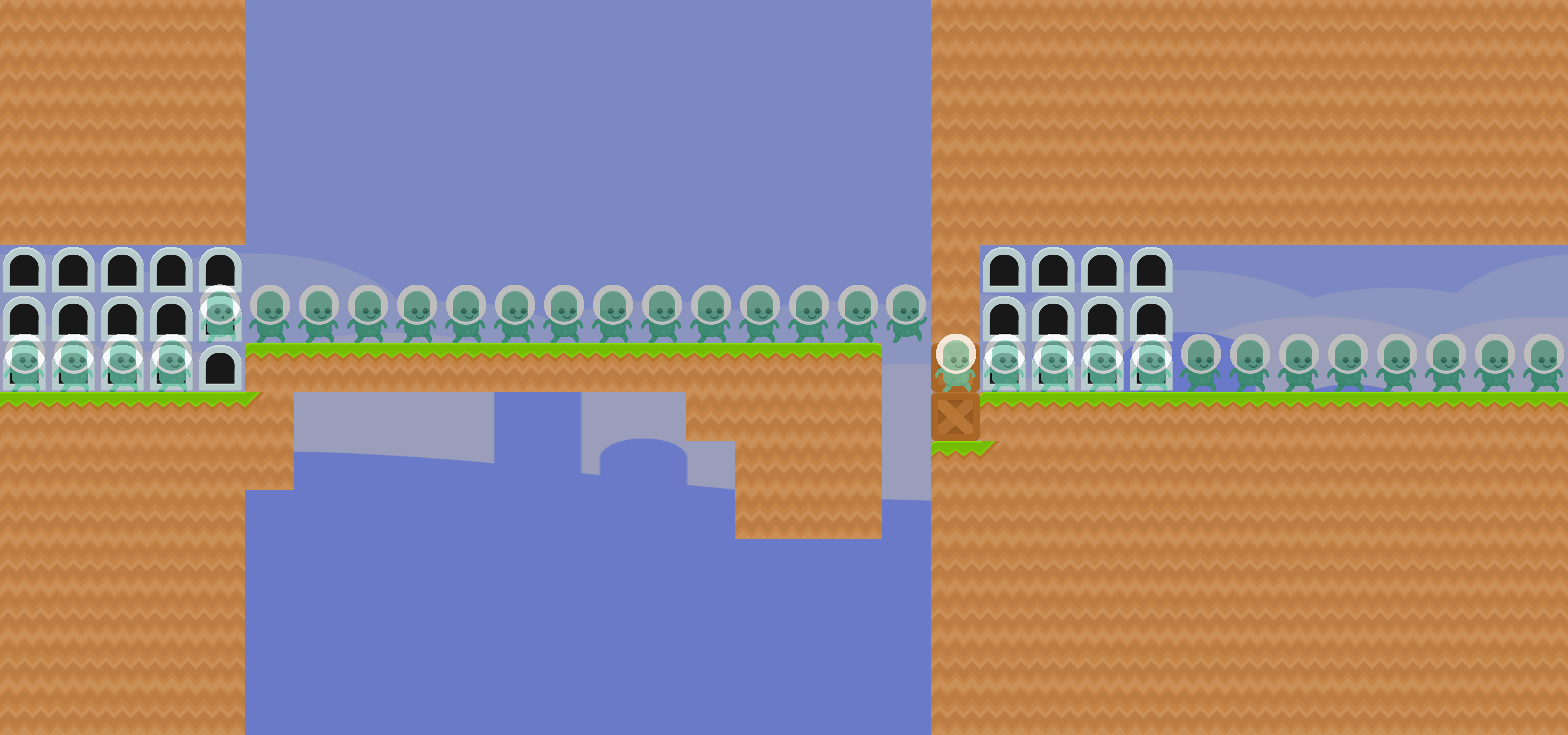}

\includegraphics[width=0.12\pdfpagewidth]{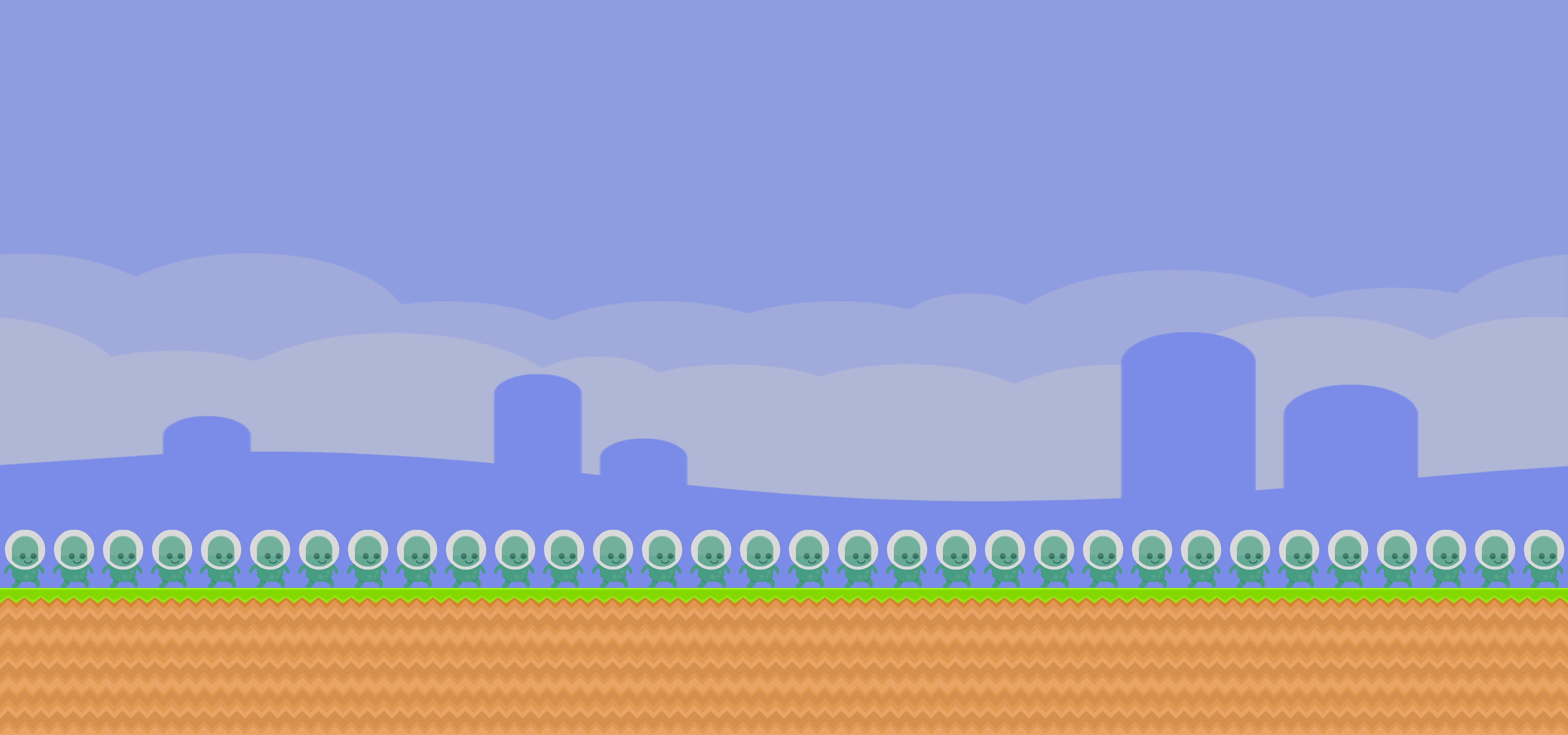}

\includegraphics[width=0.12\pdfpagewidth]{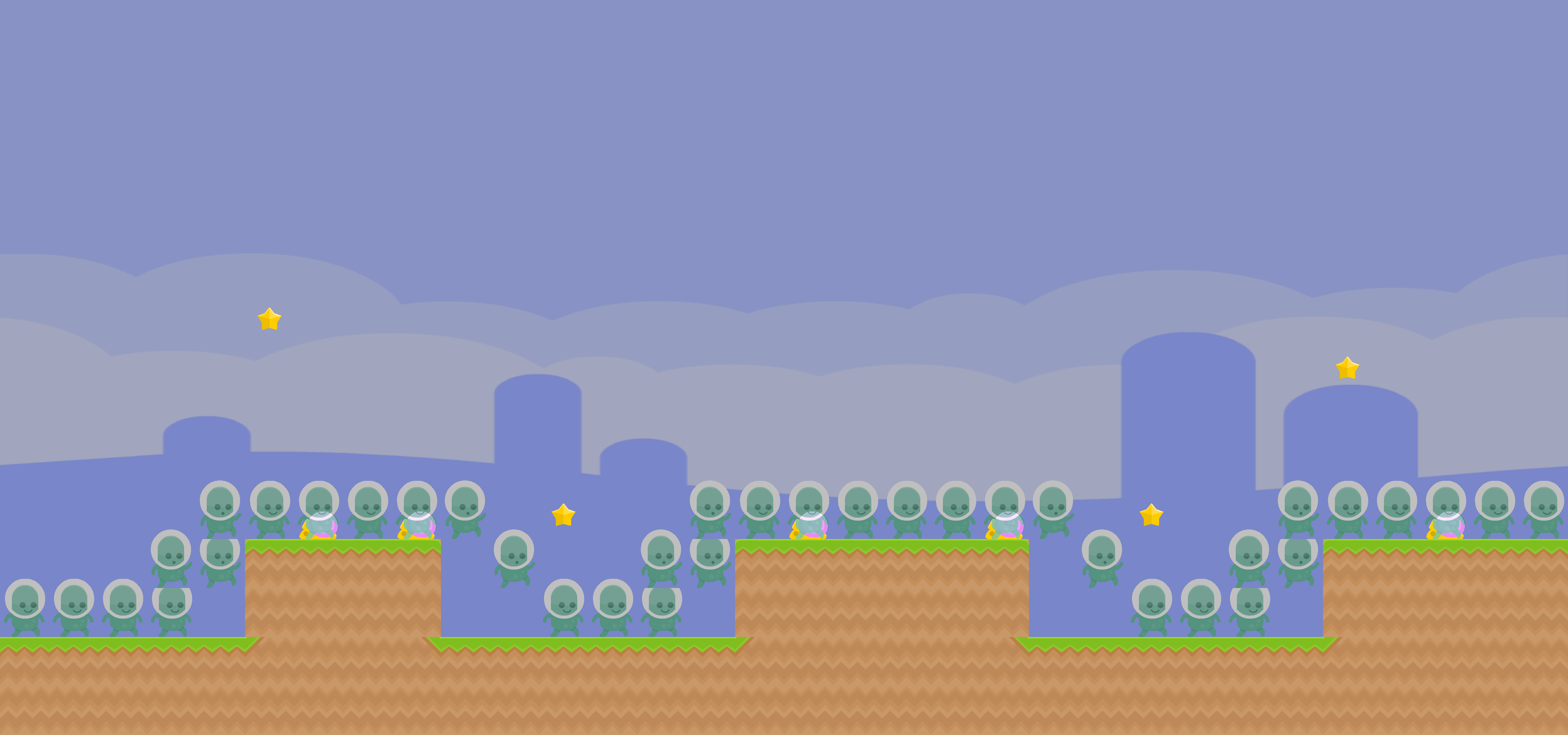}

\includegraphics[width=0.12\pdfpagewidth]{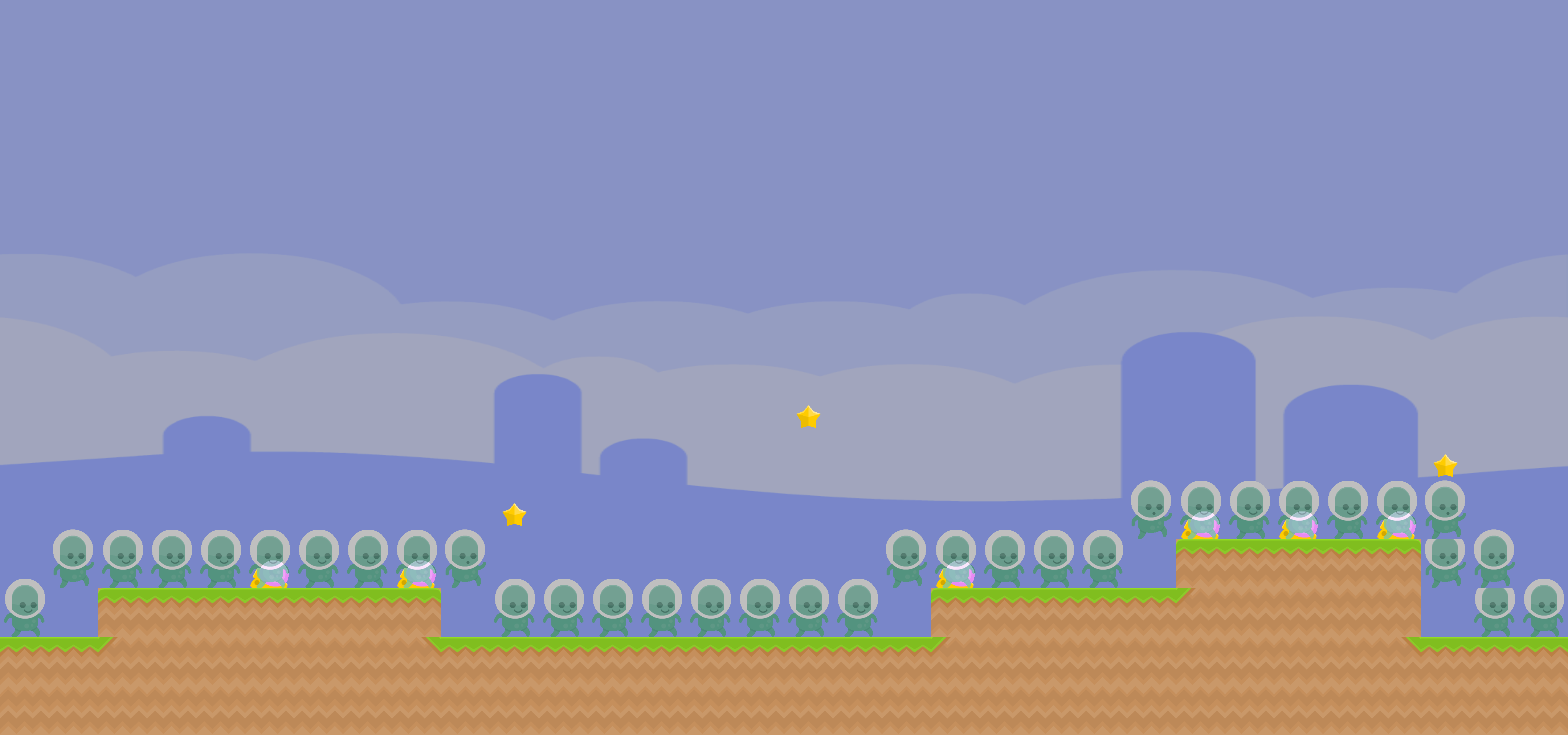}
\caption{GRU  Met $\downarrow$ NG}
\label{fig:gru_interp_Met_NG}
\end{subfigure}
\caption{Example interpolations for all pairs of games.}
\label{fig:gru_interp}
\end{figure*}

\begin{figure}[ht!]
\centering
\includegraphics[width=0.12\pdfpagewidth]{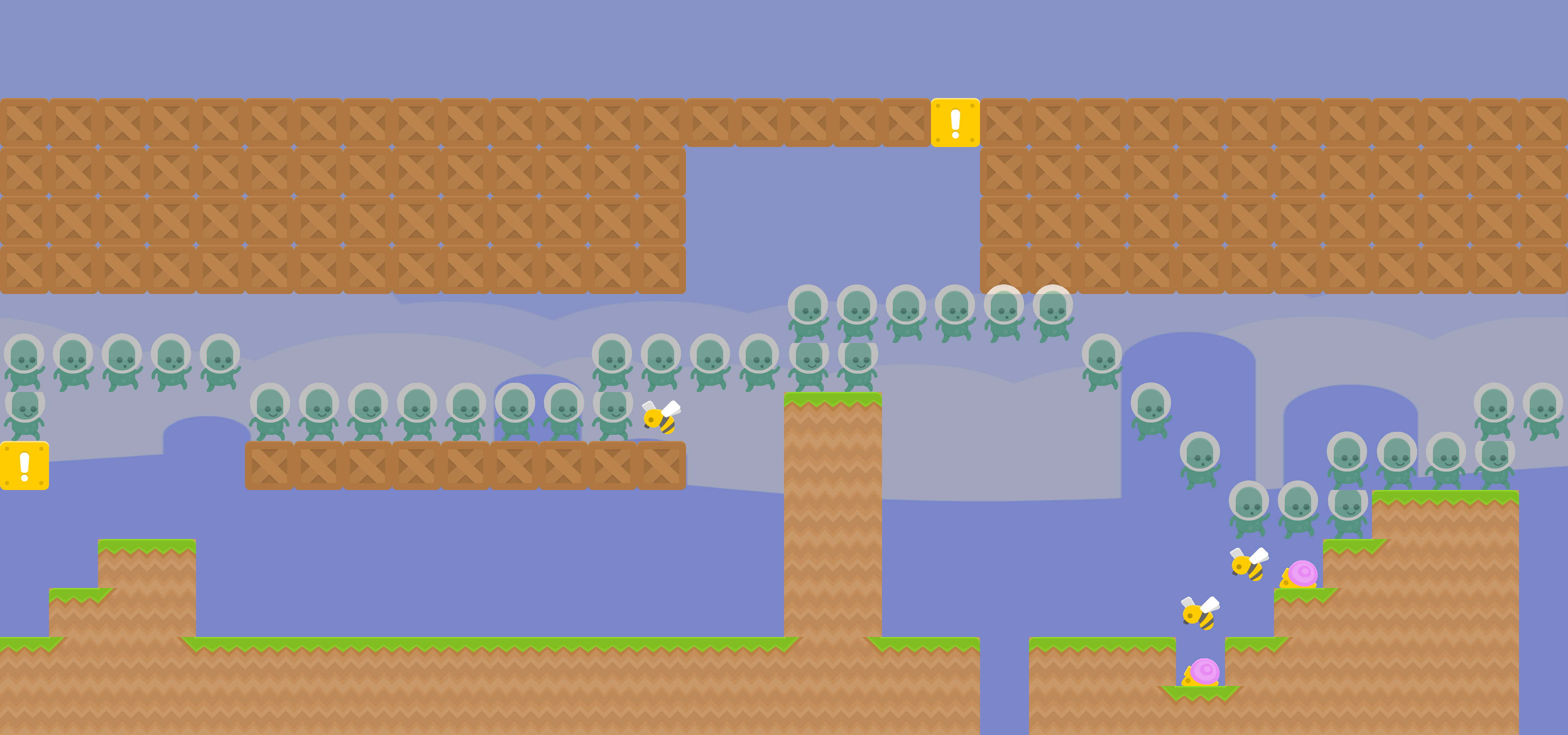}

\includegraphics[width=0.12\pdfpagewidth]{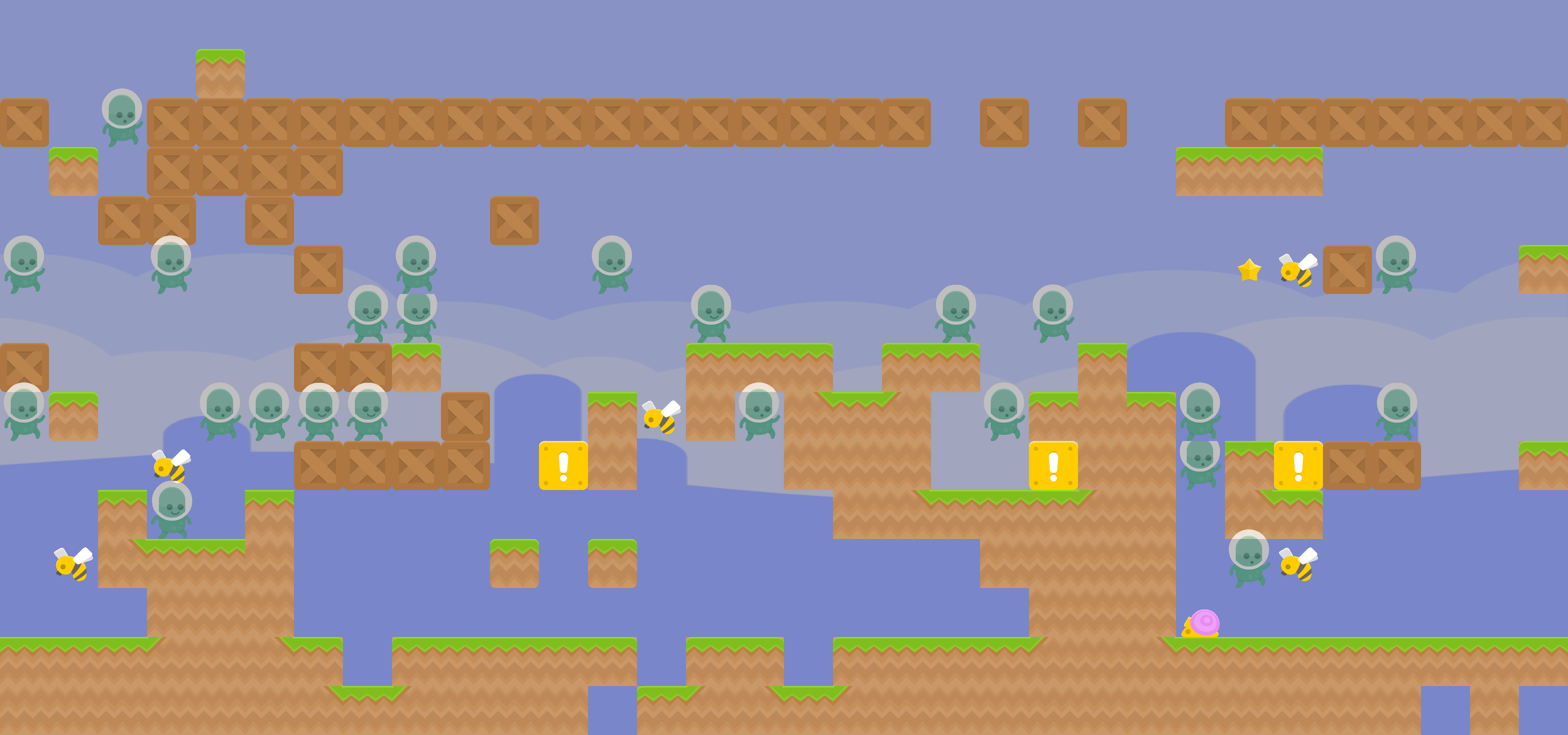}

\includegraphics[width=0.12\pdfpagewidth]{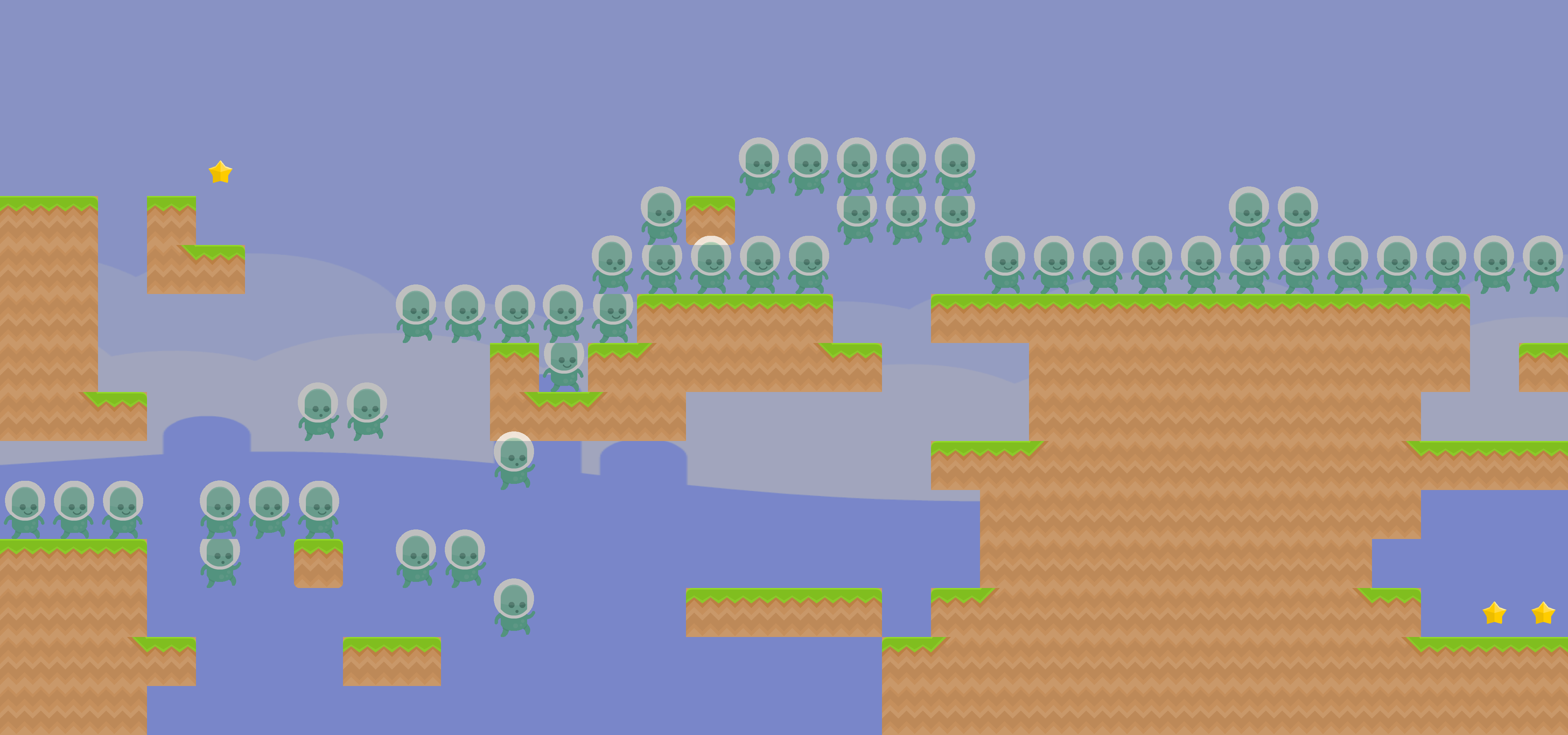}

\includegraphics[width=0.12\pdfpagewidth]{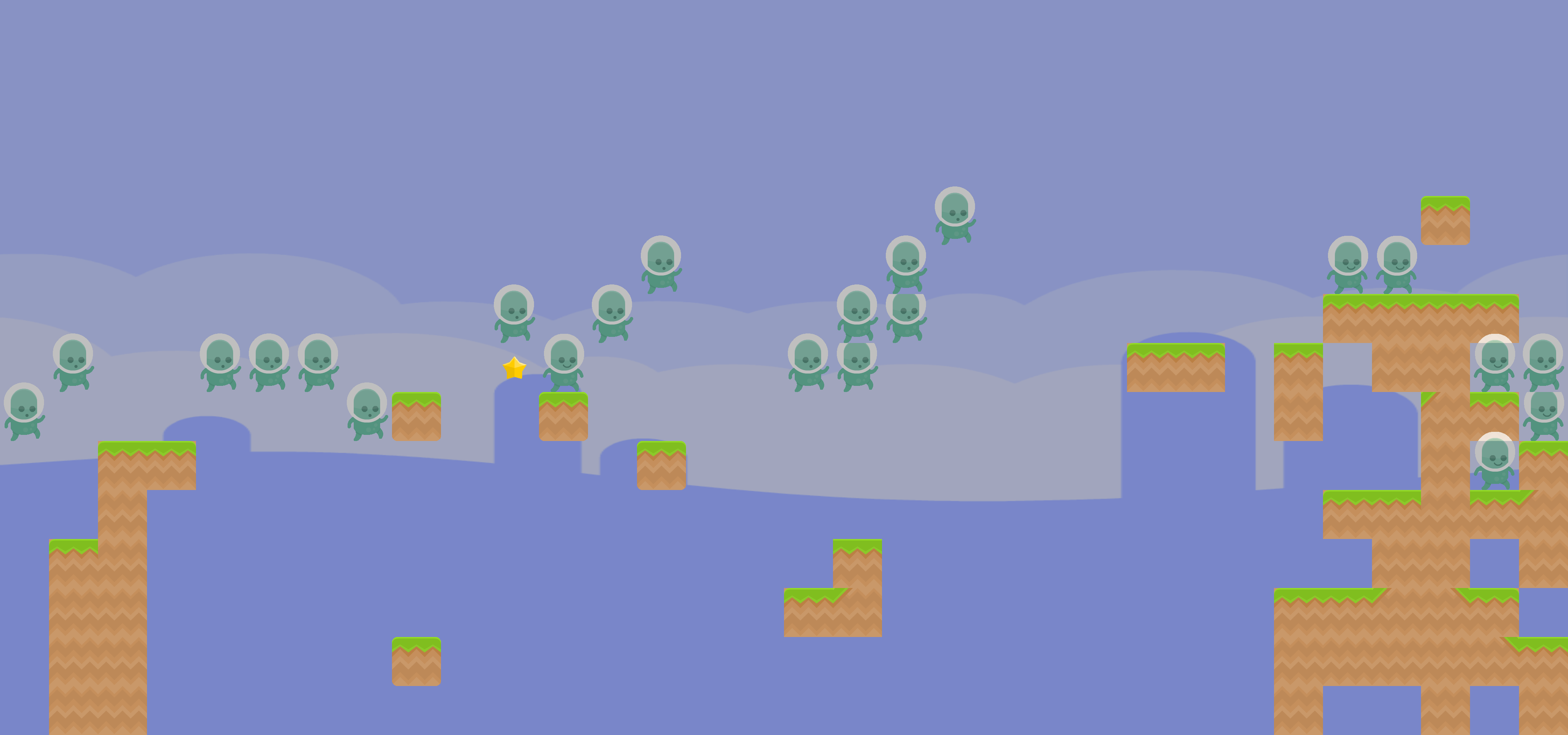}

\includegraphics[width=0.12\pdfpagewidth]{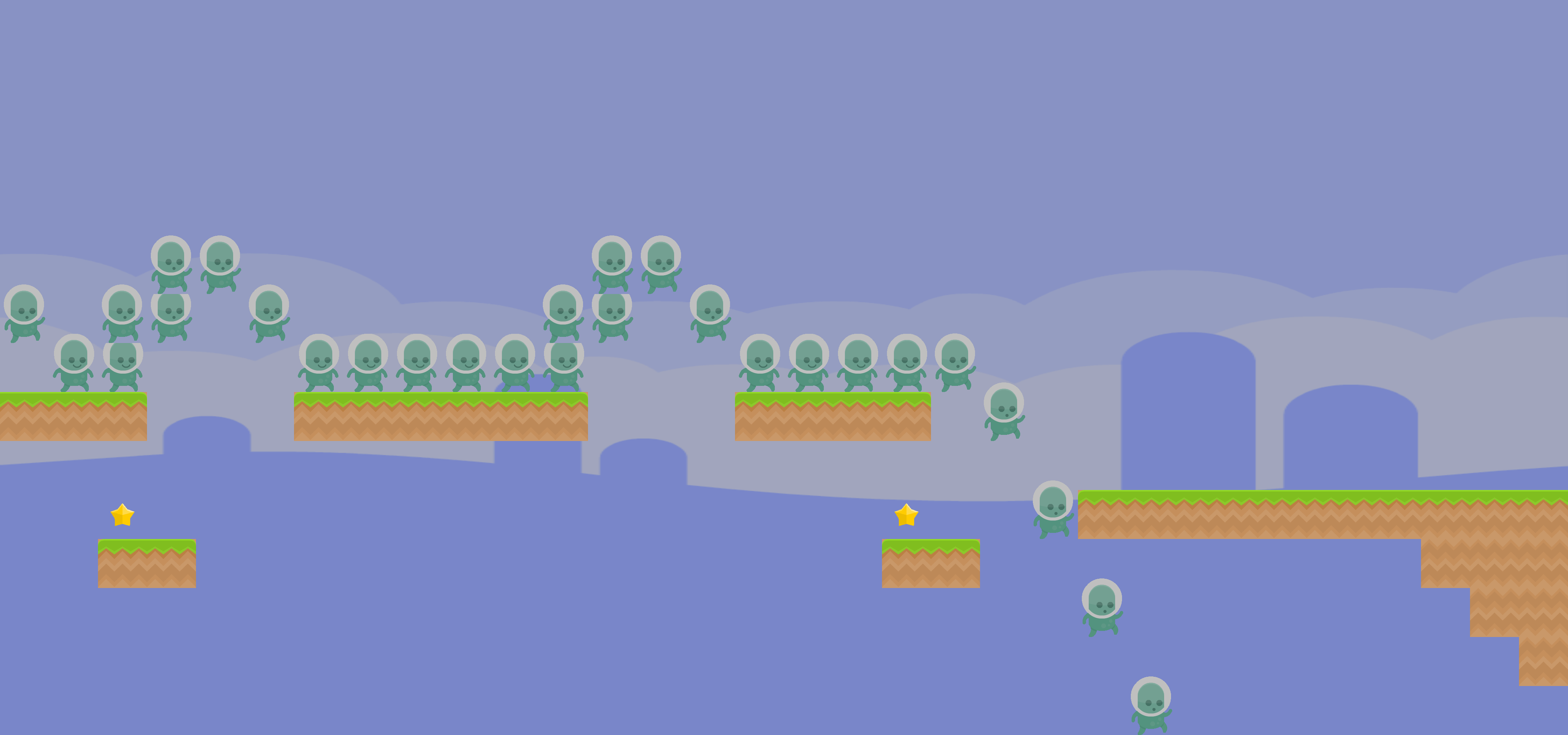}
\caption{Linear SMB $\downarrow$ MM}
\label{fig:lin_interp}
\end{figure}

Examples of randomly sampled levels generated using both models are shown in Figures \ref{fig:lin_sample} and \ref{fig:gru_sample} using a neutral sprite representation for all games (taken from Kenney\footnote{\url{https://www.kenney.nl/assets/platformer-art-deluxe}}). Both the linear and the GRU models are able to learn different ``styles'' of levels---e.g., some are cave-like while some are more open. The GRU model produces contiguous paths while the linear model's paths are noisier and less coherent. Example interpolations between all pairs of games using the GRU model are shown in Figure \ref{fig:gru_interp}. Figure \ref{fig:lin_interp} shows a single such interpolation using the linear model. The GRU model has a number of segments that it finds highly likely---e.g., a flat section with a hill (SMB $\downarrow$ Met, NG, CV) or a long cavelike corridor (CV $\downarrow$ MM, Met).  The  autoregressive generation of the GRU is possibly a reason such segments show up---it tries to produce sequences that are likely based on what has been generated so far. The linear model's output looks much closer to a straight interpolation of the two segments (tiles that appear in one and are empty in the other disappear in the interpolation) with the patterns that emerge looking noisier than the endpoints of the interpolation. The GRU model's interpolations are more coherent with none of the noise of the linear model; however, the interpolations are also less obviously interpolations.  Many interpolations make  sense: SMB $\downarrow$ MM goes from a lot of destructible blocks and little empty space to the opposite; CV $\downarrow$ MM goes from a wide-open space to a cave-like hallway with the interpolations getting more and more filled in; and MM $\downarrow$ Met goes from wide open to more constrained with doors, and the 50\% mark looks like a compromise of the two. On the other hand, the interpolations between SMB $\downarrow$ NG, SMB $\downarrow$ CV, and Met $\downarrow$ NG seem to be determined by their respective endpoints.  MM $\downarrow$ NG has very little variation in the interpolations, perhaps due to the end point similarity.

\section{Conclusion and Future Work}\label{sec:conc}
We presented a new PCGML approach that leverages path information and a new affordance vocabulary to extend existing VAE-based level generation and blending techniques to produce traversable blended levels across a greater number of game domains. In the future, our approach to generate traversable level blends by incorporating paths from multiple domains could enable learning blended physics models spanning multiple games. Physics models are useful for informing game-playing agents such as the A* agents used to encode paths in this work. Thus, learning and extracting such models across multiple game domains would complement the multi-domain blended levels presented above. 

We limited our approach to horizontal level sections in this work. Future work can explore blending vertical sections from \textit{Metroid}, \textit{Mega Man} and \textit{Ninja Gaiden} along with games like \textit{Kid Icarus}. Blending levels and physics for horizontal and vertical domains will support future work in blending games that simultaneously scroll in two directions. Finally, while prior PCGML works have looked at other game genres, techniques like blending and domain transfer have not been applied outside of platformers. Future work should test such approaches in other genres such as action-adventure games and dungeon crawlers. 


\bibliographystyle{aaai}
\bibliography{references.bib}

\end{document}